\documentclass[letterpaper, journal, twoside]{ieeetran}
\usepackage{amsmath,amsfonts}
\usepackage{array}
\usepackage[caption=false,font=normalsize]{subfig}
\usepackage{textcomp}
\usepackage{siunitx}
\usepackage{stfloats}
\usepackage{url}
\usepackage{verbatim}
\usepackage{cite}
\usepackage{graphicx}
\usepackage{cite}
\usepackage{amsmath,amssymb,amsfonts}
\usepackage{algorithm}
\usepackage{algorithmic}
\usepackage{xcolor}
\usepackage{courier} 
\usepackage{textcomp}
\usepackage{caption}
\usepackage{booktabs} 
\usepackage{multirow}
\usepackage{bm}
\usepackage{subcaption}
\def\etal{{\em et al.}}
\usepackage{siunitx}
\usepackage{color}
\usepackage[dvipsnames]{xcolor}
\definecolor{arroworange}{RGB}{255, 185, 0}
\definecolor{arrowyellow}{RGB}{255, 255, 0}
\definecolor{arrowpurple}{RGB}{212,0,212}
\usepackage[table]{xcolor}
\usepackage{colortbl}

\usepackage{threeparttable}
\usepackage{booktabs}
\usepackage{array}

\begin{document}

\title{LiDAR Teach, Radar Repeat: Robust Cross-Modal Navigation in Degenerate and Varying Environments}

\author{Renxiang Xiao$^{\dagger}$, Yichen Chen$^{\dagger}$, Yuanfan Zhang, Qianyi Shao,  Yushuai Chen,\\ Yuxuan Han, Yunjiang Lou, Liang Hu$^*$
\thanks{R. Xiao, Y. Chen, Q. Shao, Y. Chen, Y. Han, Y.Lou and L. Hu  are with School of Intelligence Science and Engineering, Harbin Institute of Technology, Shenzhen, China. Y. Zhang is with School of Computer Science and Technology, Harbin Institute of Technology, Shenzhen, China. $^{\dagger}$ Equal Contribution, $^*$ For correspondence: {\tt\footnotesize l.hu@hit.edu.cn}.}
}

\markboth{IEEE Transactions on Robotics. PREPRINT VERSION. ACCEPTED May, 2026}
{Xiao \MakeLowercase{\textit{et al.}}: LiDAR Teach, Radar Repeat: Robust Cross-Modal Navigation in Degenerate and Varying Environments} 

\maketitle

\begin{abstract}
Long-term autonomy requires robust navigation in environments subject to dynamic and static changes, as well as adverse weather conditions. Teach-and-Repeat (T$\&$R) navigation offers a reliable and cost-effective solution by avoiding the need for consistent global mapping; however, existing T$\&$R systems lack a systematic solution to tackle various environmental variations such as weather degradation, ephemeral dynamics, and structural changes. This work proposes LTR$^2$, the first cross-modal, cross-platform LiDAR-Teach-and-Radar-Repeat system that systematically addresses these challenges. LTR$^2$ leverages LiDAR during the teaching phase to capture precise structural information under normal conditions and utilizes 4D millimeter-wave radar during the repeating phase for robust operation under environmental degradations. To align sparse and noisy forward-looking 4D radar with dense and accurate omnidirectional 3D LiDAR data, we introduce a Cross-Modal Registration (CMR) network that jointly exploits Doppler-based motion priors and the physical laws governing LiDAR intensity and radar power density. Furthermore, we propose an adaptive fine-tuning strategy that incrementally updates the CMR network based on localization errors, enabling long-term adaptability to static environmental changes without ground-truth labels.

We demonstrate that the proposed CMR network achieves state-of-the-art cross-modal registration performance on the open-access dataset. Then we validate LTR$^2$ across three robot platforms over a large-scale, long-term deployment (40+ km over 6 months), including challenging conditions such as nighttime smoke. Experimental results and ablation studies demonstrate centimeter-level accuracy and strong robustness against diverse environmental disturbances, significantly outperforming existing approaches.
\end{abstract}


\section{Introduction}

\begin{figure}[t]
    \centering
        \subfloat[ Repeat successfully through smoke, unseen in teach.\label{fig:intro_smoke}]{
        \includegraphics[width=\linewidth]{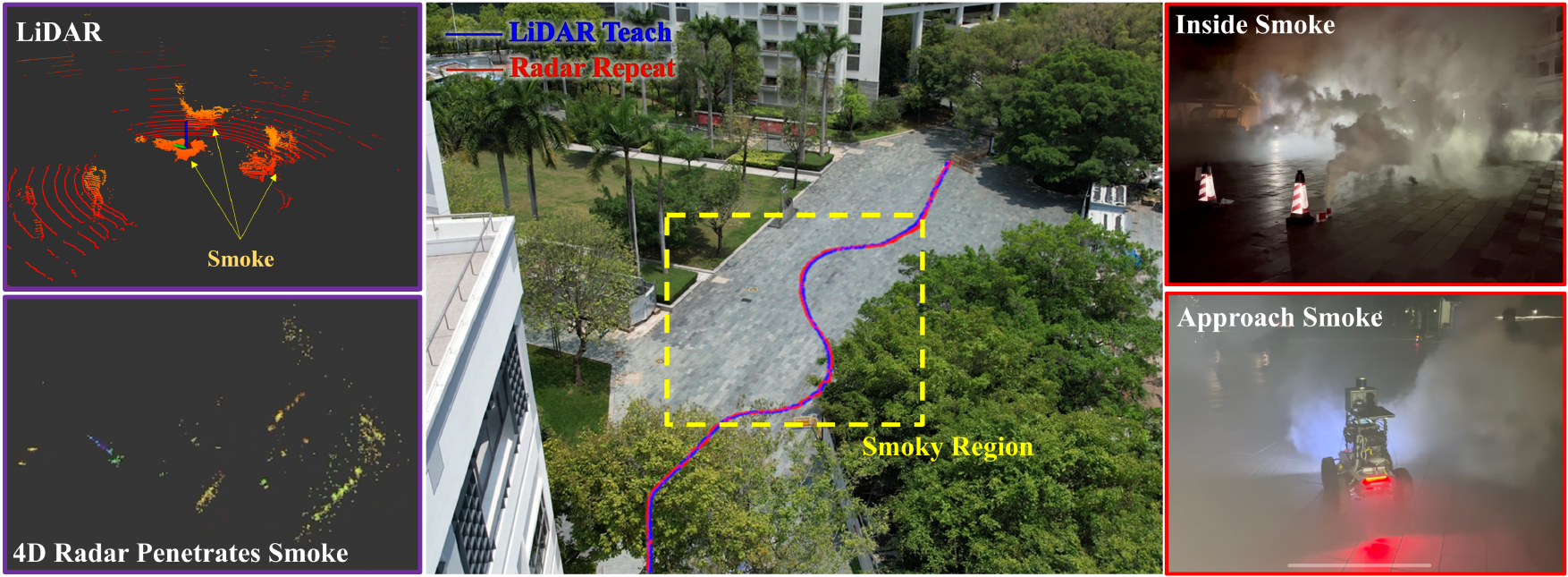}
    }\\
    \subfloat[A single T$\&$R trajectory for 4.627 km on campus.\label{fig:intro3}]{
        \includegraphics[width=\linewidth]{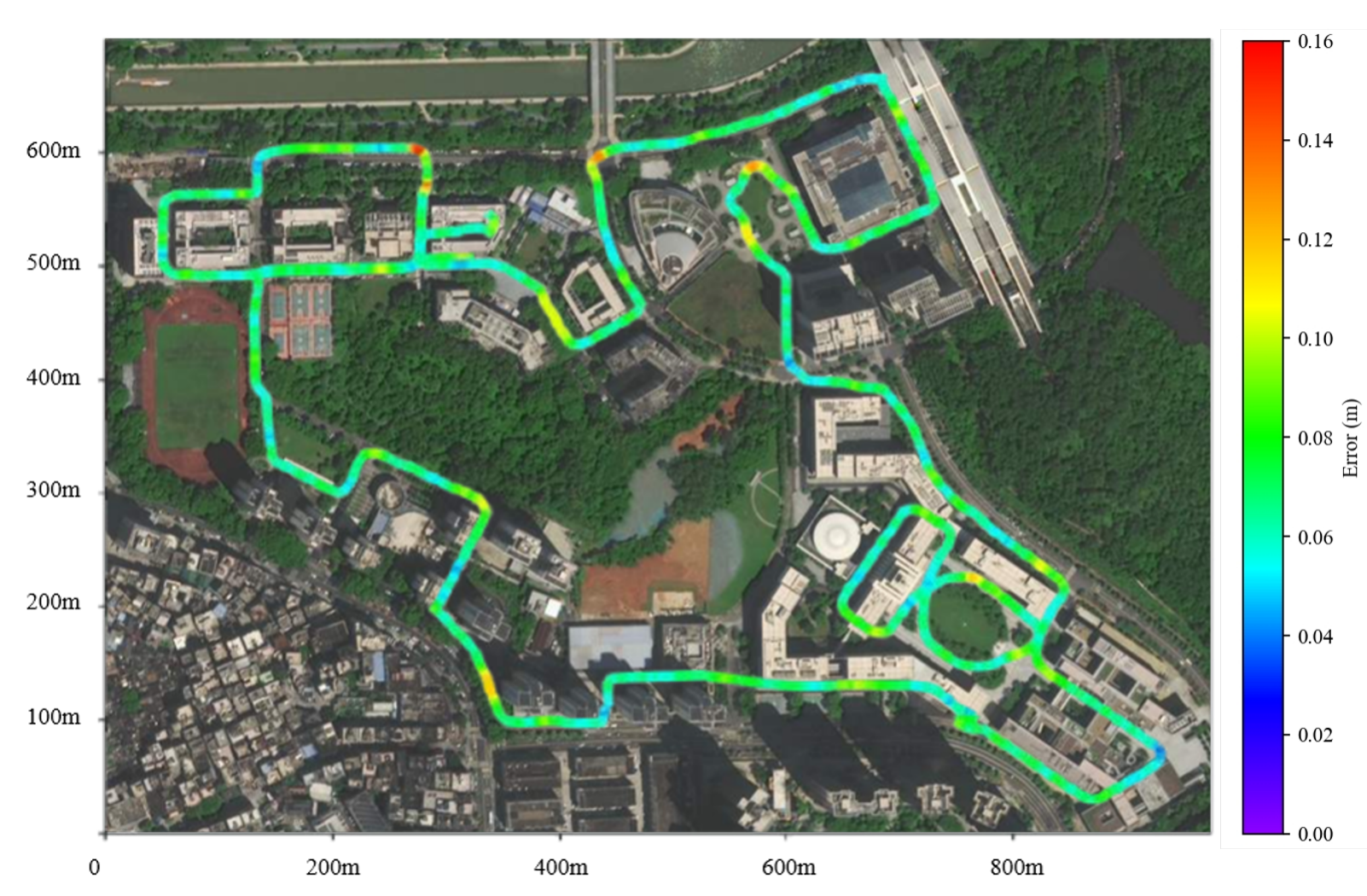}
    }
\caption{\textbf{Demonstration trajectories of the LTR$^2$ system.} (a) The repeat trajectory is covered by an unexpected smoke, which was not seen during trajectory teaching. (b) A navigation demonstration with a color-coded error map visualizing deviations between teach-and-repeat paths, achieving centimeter-level accuracy on average over multi-kilometer trajectories.}
    \label{fig:intro}
    \vspace{-1mm}
\end{figure}
Long-term autonomy is a fundamental, yet challenging research topic that has attracted sustained interest within the robotics community \cite{lowry2015visualsurveytro,churchill2013experiencenewman, barfoot2013specialIJRR}. Among various navigation paradigms, Teach-and-Repeat (T$\&$R) has emerged as a promising approach to long-term navigation that distinguishes itself by removing the costly processing of constructing and maintaining metrically consistent maps. It operates in a two-phase framework: during the teaching phase, the robot records detailed environmental information along a piloted trajectory; in the repeating phase, it utilizes this recorded data to navigate the same path autonomously. Owing to its reliability and low cost, T$\&$R has been widely adopted for navigation in diverse environments, including underground mines\cite{qiao2024mine}, underwater scenarios \cite{king2018teachunderwater}, forests \cite{Baril2021KilometerscaleAN}, planetary terrains \cite{tim2010, akemoto2023cooperative} and large-scale areas where maintaining and updating a precise map is costly \cite{mcmanus2013lightingtim}. 

Despite its advantages, T\&R faces key challenges during long-term deployment due to inevitable environmental changes. On one hand, the weather changes over time bring variations such as changes in lighting conditions, rain, snow, fog, or smoke. On the other hand, the environment may experience changes such as urban redevelopment or road construction, which we call static changes and assume to evolve slowly (we exclude scenarios like robot kidnapping, where the robot operates in a
completely different environment). These changes can significantly impact the feature matching between the teach and repeating phases, degrading navigation performance. Although some prior works have attempted to enhance the robustness of visual T$\&$R\cite{krusi2015lighting, rouvcek2022self,sun2021robust} or LiDAR T$\&$R\cite{mcmanus2013lightingtim,burnett_ral22,sprunk2013LiDART&R,cattaneo2025cmrnexttro} against appearance changes due to lighting conditions and seasons, a systematic solution addressing all different environmental changes remains absent, constituting the primary focus of our research. 

Recently, millimeter-wave (mmWave) radar has shown robust perception capabilities under adverse weather conditions, thanks to its ability to penetrate small particles and detect objects at long range, even beyond dynamic occlusions \cite{adolfsson2022lidartro,harlow2024newtro,zhang20234dradthermal,zhang2024adaptive,park20213Dego,xiaoradgs}. However, mmWave radars typically produce sparse point clouds with low spatial resolution and unstable feature extraction. To mitigate these issues, some pioneering works \cite{yin2020radaronLiDAR,yin2021rall,yin2021radar2LiDAR,ma2023rolm,ma2024fmcw,park2019radarlikehood,nayak2024ralf} have focused on aligning 3D mmWave radar (x, y, Doppler) with LiDAR in the BEV (Bird’s Eye View), leveraging complementary strengths from both modalities.

Inspired by the idea of cross-modal alignment of mmWave radar with LiDAR, we propose a \textit{cross-modal} LiDAR Teach Radar Repeat (LTR$^2$) solution that leverages cross-modal alignment to enhance its robustness to environmental changes during long-term navigation. Different from existing T$\&$R systems, it employs different modalities rather than a single modality at the teach and repeating phases: the high-precision LiDAR to capture detailed structural features in normal conditions during the teaching phase, and the 4D mmWave radar (denoted as 4D radar for short thereafter) to ensure reliable navigation across various weather conditions and ephemeral scene changes during the repeating phase, as illustrated in Fig.~\ref{fig:intro}.  Such a heterogeneous sensing configuration at different stages reflects a practical system design for long-term deployment. In many real-world scenarios, route teaching is carried out under normal conditions using well-instrumented platforms, whereas route repetition is performed continuously over extended periods. Under such a setting, using high-precision but relatively costly LiDAR during teaching, and lightweight, cost-effective radar during repetition, provides a realistic trade-off between sensing fidelity and operational robustness. This design is particularly relevant in environments such as industrial sites, underground facilities, or fire-affected areas, where visibility conditions may degrade due to dust, smoke, or dynamic disturbances.

Nonetheless, aligning 4D radar with LiDAR presents additional challenges: compared to 3D mmWave radar, 4D radar features sparser point clouds, smaller horizontal fields of view (FOV), and lower resolution.
Thus, existing LiDAR-3D radar alignment methods are inapplicable to LiDAR-4D radar registration. To fill the gap, we introduce a novel data preprocessing and learning-based cross-modal registration procedure in which both the Doppler-based motion prior and the underlying physical principles governing  LiDAR intensity and radar power density are exploited. Moreover, we propose an incremental learning based finetuning strategy that enables adaptive evaluation and updating of the Cross-Modal Registration (CMR) network to maintain high performance under significant static environmental changes over time. Together, the integration of robust 4D radar sensing, reliable cross-modal registration, and adaptive fine-tuning strategies enables the LTR$^2$ system to maintain long-term, robust navigation in the presence of weather, ephemeral, and static changes.

    
    


Our research contributions are summarized as follows:
\begin{enumerate}
\item We introduce a learning-based LiDAR-teach-4D radar-repeat system, achieving robust navigation in degenerate and
changing environments, with the potential of large-scale
deployment across diverse robotic platforms;

\item We propose a novel data preprocessing and CMR method that leverages not only geometric features but also the underlying physical laws of LiDAR point cloud intensity and radar power density,  enabling effective end-to-end alignment between 4D radar and LiDAR; 

\item  We introduce an adaptive fine-tuning strategy that enhances CMR’s adaptability to static environmental changes, using a localization-error-based metric for detection and an incremental learning-based self-supervised update mechanism;

\item We validated our proposed LTR$^2$ system through extensive outdoor experiments across three robot platforms, covering over 40+ km over 6 months, and including tests in challenging conditions such as smoky scenarios at night. Comparative analysis and ablation studies demonstrate the superiority of our proposed method.
\end{enumerate}

The remainder of this paper is organized as follows: Sec. \ref{Related work} reviews the related work. Sec. \ref{Method} introduces the overall system workflow of LTR$^2$ and the details of its subsystems. Sec. \ref{Cross-Modal localization} analyzes the data alignment preprocessing and the proposed cross-modal localization network. Sec. \ref{Experiment} presents the experiment results and analysis. Finally, Sec. \ref{Conclusion} concludes the paper.

\section{Related Work}\label{Related work}
\subsection{Teach-and-Repeat Navigation}
T$\&$R paradigms were initially proposed as an alternative to GPS-dependent navigation, enabling long-term autonomous operation without the need for explicit mapping \cite{lowry2015visualsurveytro,kunze2018long-termrobotautonomysurvey}. Early visual T$\&$R methods predominantly relied on feature extraction and matching, where features were filtered and stored based on repeated experiences \cite{tim2010, churchill2012practicenewman}. Recently, deep neural networks have been incorporated into visual T$\&$R to improve the system's adaptiveness to changes in lighting or structure~\cite{Krajník2018,liu2023self,rouvcek2024predictive}. Nonetheless, it remains difficult to obtain reliable image matching in the extreme environments of poor lighting or low textures. On the other hand, LiDAR~\cite{deschenes2021LiDAR,krusi2015lighting,sprunk2013LiDART&R,Baril2021KilometerscaleAN} or LiDAR-vision fusion\cite{mcmanus2013lightingtim} T$\&$R methods are robust to illumination changes, but they struggle under adverse weather due to perception degradation and significant structural changes due to their reliance on geometric features. Therefore, maintaining reliable navigation against evolving point clouds over long periods remains an open challenge. Very recently, Qiao et al. \cite{qiao2024mine} has explored a radar-based T$\&$R solution, but its effectiveness is limited by the lack of elevation information from the spinning radar.  

Despite advancements in camera-, LiDAR-, and radar-based T\&R systems, enhancing their robustness and reliability across diverse environmental conditions remains a significant challenge. To date, there still lacks an effective solution to systematically tackle the wide range of environmental changes commonly encountered in long-term deployment. A promising direction is to harness the complementary merits of different sensing modalities thereby increasing robustness to weather variants and transient dynamics, while leveraging learning-based methods to fine-tune the system for gradual, static environmental changes. Motivated by it,  we propose a cross-modal T$\&$R system that aligns the 4D radar with  LiDAR through a CMR network.

\subsection{Radar on LiDAR Localization}
The complementary characteristics of radar and LiDAR motivate extensive research into cross-modal data alignment and localization \cite{harlow2024newtro}. Yin~et al. \cite{yin2020radaronLiDAR,yin2021rall,yin2021radar2LiDAR} introduce deep learning-based methods to align heterogeneous sensors. Specifically,  Yin~\cite{yin2020radaronLiDAR} uses a generative adversarial network to denoise radar data and synthesize LiDAR-like representations, and Yin~\cite{yin2021rall} employs a U-Net architecture to learn cross-modal shared feature embeddings under pose supervision. Building on these, the subsequent study~\cite{yin2021radar2LiDAR} further trains the joint position recognition of radar and LiDAR data to achieve robust cross-modal pose tracking. Other methods such as Cen et al.\ \cite{cen2018precisesignal} propose a noise-removal scheme and handcrafted descriptors for radar signals, while Ma et al. \cite{ma2024fmcw,ma2023rolm} use polar coordinates and Cartesian projections to select the most suitable LiDAR frames for ICP-based alignment. In addition, Park~et al.\cite{park2019radarlikehood} introduce a corresponding matrix into the generalized likelihood ratio test framework to evaluate and optimize the radar-LiDAR shape matching, and Nayak~et al. \cite{nayak2024ralf} propose pixel-wise alignment using encoders on BEV projections.

Most existing methods focus on aligning geometric features derived from traditional 3D mmWave radar data to LiDAR, and cannot be directly applied to the forward-looking 4D radar due to their fundamental differences. As summarized in Tab. \ref{tab:sensor_comparison}, compared to the 3D radar and LiDAR, the  4D radar produces a much sparser point cloud with a narrower horizontal FOV and lower resolutions, posing substantial challenges to reliable feature matching across LiDAR and 4D radar modalities.

\begin{table}[t]
  \centering
  \caption{Sensor Comparison.}
  \label{tab:sensor_comparison}
  \begin{threeparttable}
    \resizebox{\linewidth}{!}{%
      \begin{tabular}{>{\raggedright\arraybackslash}p{4.0cm}|ccc}
        \toprule
        Sensor & 4D Radar & Spinning Radar & LiDAR \\ \midrule
        Type   & Oculii Eagle G7 & Navtech RAS6 & RoboSense Helios 32 \\ \midrule
        Data Type &
          \begin{tabular}[c]{@{}c@{}}x,\,y,\,z,\\ Doppler, power\end{tabular} &
          \begin{tabular}[c]{@{}c@{}}x,\,y,\\ power, Doppler$^{*}$\end{tabular} &
          \begin{tabular}[c]{@{}c@{}}x,\,y,\,z,\\ intensity\end{tabular} \\ \midrule
        Range Resolution (m)      & 0.86 & 0.043 & 0.01 \\
        Azimuth Resolution (°)    & 0.5  & 0.9   & 0.1 \\
        Elevation Resolution (°)  & 1.0  & /     & 0.5 \\ \midrule
        Detection Range (m)       & 400  & 330   & 150 \\
        Azimuth FOV (°)           & 120  & 360   & 360 \\
        Elevation FOV (°)         & 30   & /     & 70 \\ \midrule
        Frequency (Hz)            & 14   & 4     & 10 \\ 
        Points per Second         & $\approx 4.16 \times 10^{3}$ & $\approx 8 \times 10^{3}$ & $\approx 6.0 \times 10^{4}$ \\ 
        \bottomrule
      \end{tabular}
    } 
    \vspace{0.5em}
    \begin{tablenotes}
      \footnotesize
      \parbox{0.72\linewidth}{
      \item[/] Indicates the absence of elevation data in spinning radar measurements.
      \item[$^{*}$] Doppler velocity is not directly available for the RAS6 spinning radar but can be estimated from successive azimuth scans~\cite{25raltim}.}
    \end{tablenotes}
  \end{threeparttable}
\end{table}

\label{tab:sensor-compare}

\subsection{Navigation against Environmental Changes}
Environmental changes are commonly encountered in long-term navigation, and four kinds of methods are proposed to mitigate their impact on navigation performance.  First, experience-based frameworks \cite{churchill2012practicenewman,maddern20171newman,milford2013vision} maintain multiple historical observations rather than a unified global map, allowing systems to select relevant memories during operation adaptively. Second, another research direction models environmental dynamics to predict future or unobserved states. Wang et al. \cite{wang2024longterm} propose a spatiotemporal map prediction framework that captures the temporal dynamics of obstacle occupancy. Similarly, Krajník et al.\cite{krajnik2017fremen} introduce the frequency map enhancement model to describe periodic environmental variations, while Neubert et al.\cite{neubert2015superpixel} propose SP-ACP to learn seasonal appearance shifts by constructing a dictionary of visual words across different conditions. Third, lifelong map updating techniques \cite{gil2025ephemerality,yang2024lifelong,woo2024no, Halodová,vodisch2022continual} focus on continuously updating environmental representation to maintain the consistency and usability of point cloud maps over extended periods. Finally, lifelong localization aims to maintain accurate pose estimation under environmental variations. Rouček et al. \cite{rouvcek2024predictive} propose a self-supervised data selection framework to improve localization accuracy in visual teach-and-repeat, while Clark et al. \cite{clark2017vidloc} introduce continual network training and fine-tuning to implicitly embed visual variations into model update.  Hausler et al. \cite{hausler2021patchnet} and Radwan et al. \cite{radwan2018vlocnet++} propose incremental updates and deep semantic embedding strategies to improve long-term localization robustness.

Within the context of T$\&$R paradigms, reliable localization is prioritized over maintaining full map consistency. However, the visual-based, lifelong localization is vulnerable to moving objects such as pedestrians and vehicles, especially in urban areas. In contrast, our method leverages the over-the-horizon perception of millimeter-wave radar, allowing it to perceive static structure behind dynamic objects and thus providing more stable perception under dynamic environments.

\begin{figure*}[t]
    \centering
    \includegraphics[width=0.9\linewidth]{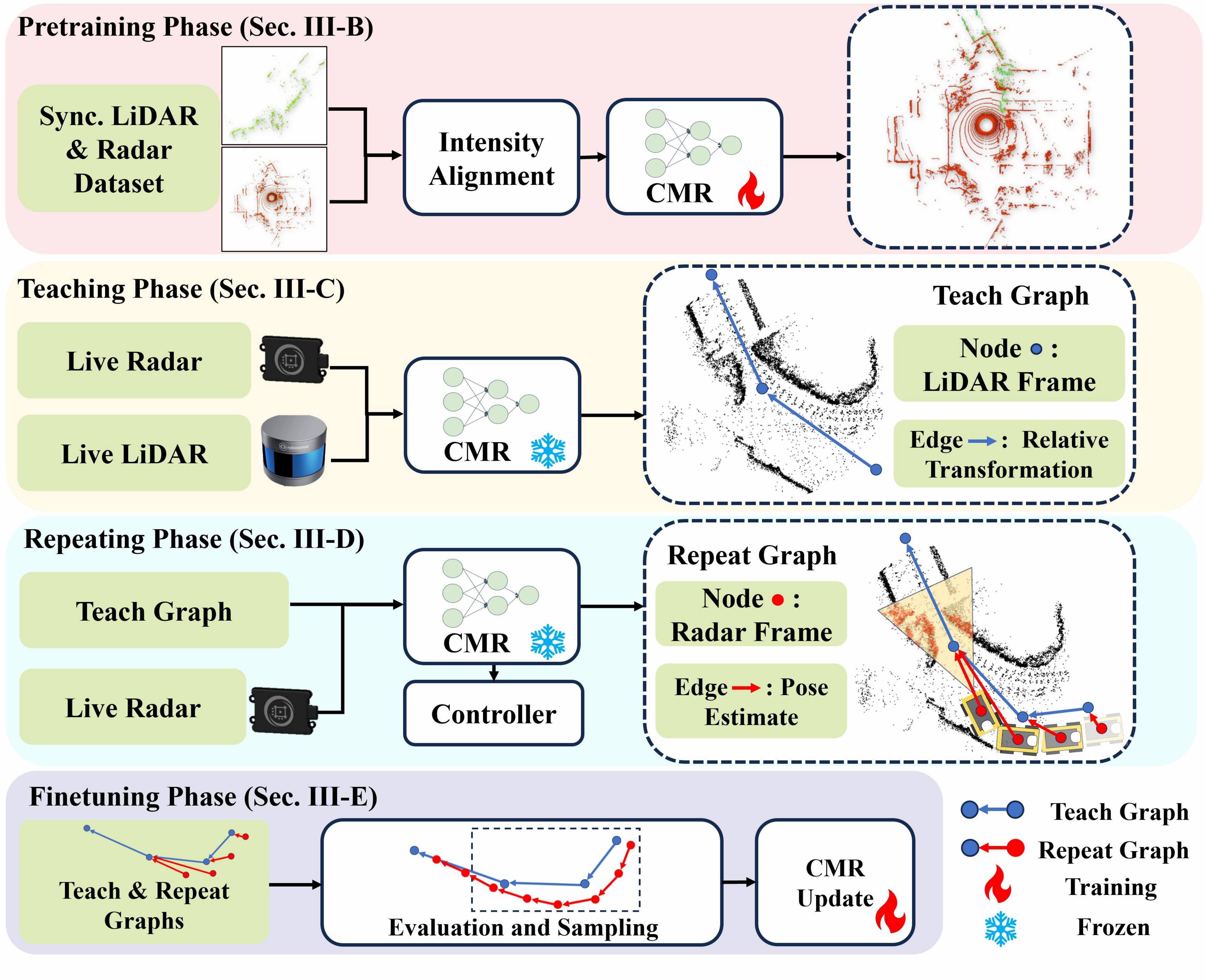}
    \caption{\textbf{System overview of the LTR$^2$ pipeline.} The system operates in four phases. In the pre-training phase, a Cross-Modal Registration (CMR) network is trained using synchronized LiDAR and 4D Radar frames. Then the trained CMR network is frozen and used in the teaching and repeating phases. In the teaching phase, the teach graph is constructed, whose nodes contain LiDAR-frame observations and edges store relative transformations. In the repeating phase, the live 4D radar frame is registered to LiDAR frames stored in the nodes of the teach graph via the CMR network, based on which the repeat graph is constructed with pose estimates stored in the edges to guide trajectory tracking. In the finetuning phase, the teach and repeat graphs are jointly evaluated to select candidate samples for updating the CMR network parameters. Snowflake and flame icons indicate frozen and training CMR network, respectively.}
    \label{fig:T&Rflowchart}
\end{figure*}

\section{LTR$^2$ System}\label{Method}
\subsection{System Overview}
As illustrated in Fig. \ref{fig:T&Rflowchart}, the LTR$^{2}$ navigation system consists of four key phases: pre-training, teaching, repeating, and fine-tuning. A CMR network plays the core role in connecting all four phases. This section focuses on the overall workflow and key phases of LTR$^2$  systems, while the network architecture and training details of CMR are deferred to the next Sec.~\ref{Cross-Modal localization}.

In the pre-training phase, the CMR network is trained using synchronized LiDAR and 4D radar data (with x, y, z, and Doppler information) collected in diverse environments. In the subsequent teaching phase, LiDAR and 4D radar measurements are recorded along the teaching path. These measurements are processed by the pretrained CMR to iteratively extract reliable teach nodes based on cross-modal consistency. During the repeating phase, the robot searches the target node for each 4D radar frame. Then a pure-pursuit controller enables the robot to follow the target nodes accurately and smoothly. After each trajectory repetition, the cross-modal localization errors are evaluated along the route to detect potential static environmental changes, and the radar-LiDAR pairs collected over the regions with environmental changes are accumulated in the sample dataset. If the changes are significant enough to trigger fine-tuning,  the CMR network will be incrementally updated using accumulated samples to maintain reliable localization over time. 

\subsection{Pre-training Phase}\label{Pre-training}
The CMR network is trained in this phase such that it can output the relative pose estimate of a 4D radar frame with respect to another LiDAR frame. The CMR network takes a LiDAR frame $\mathcal{L}_n$, a radar frame $\mathcal{R}_m$, and a prior of the relative pose $\mathbf{T}_m^{\prime n}$ as inputs,  and output $\mathbf{T}_m^n$ which corrects the coarse prediction $\mathbf{T}_m^{\prime n}$ as below:
\begin{equation}\label{eq:cmr_forward}
\mathbf{T}_m^n = \mathrm{CMR}(\mathbf{T}_m^{\prime n}, \mathcal{L}_n, \mathcal{R}_m).
\end{equation}

The CMR network is trained using the recorded LiDAR and 4D radar data when the robot is manually guided along designed paths. Each LiDAR frame is temporally paired with $N$ 4D radar frames using timestamp synchronization and known extrinsic calibration parameters. These paired samples are used to train the CMR network. To improve generalization, we add random perturbations in rotation and translation over the ego velocity of the 4D radar during CMR network training. Such data augmentation simulates varying sensor viewpoints and dynamic operating conditions, allowing the CMR network to learn robust feature associations between LiDAR and 4D radar data under real-world variations. Once trained, the CMR network is then deployed in the teaching and repeating phases and is further refined during the fine-tuning phases.

\begin{algorithm}[t]
\caption{Two-Stage Node Selection}
\label{alg:AdaptiveNodeSelection}
\footnotesize
\begin{algorithmic}[1]

\STATE \textbf{Function} \texttt{NodeSelection}()

\STATE // The First Stage: 

\STATE $\mathcal{V}_{t} \gets \emptyset$
// Node set $\mathcal{V}_{t} $
\STATE $\mathcal{V}_{t}  \gets \mathcal{V}_{t}  \cup \{(t_0, \mathbf{p}_0, L_0)\}$
// Add start node
\STATE $\mathcal{V}_{t}  \gets \mathcal{V}_{t}  \cup \{(t_{N-1}, \mathbf{p}_{N-1}, L_{N-1})\}$
// Add end node

\WHILE{$j < N-2$}  
    \WHILE{$i < N-2$} 
        \STATE $\texttt{loss} \gets \texttt{CMR}(T', L_i, R_j)$ // By \eqref{eq:cmr_forward}
        \IF{$\texttt{loss} > \tau$}
            \STATE $\mathcal{V}_{t}  \gets \mathcal{V}_{t}  \cup \{(t_{i-1}, \mathbf{p}_{i-1}, L_{i-1})\}$
            \STATE $j \gets i-1$ 
            \STATE \textbf{break}
        \ELSE
            \STATE $i \gets i + 1$
        \ENDIF
    \ENDWHILE
\ENDWHILE

\STATE // The Second Stage: 
\FOR{$ k = 1$ to $|\mathcal{V}_{t} | - 1$}
     \STATE // Boundary nodes for Hermite interpolation: $w_{k-1}, w_k$
     \STATE $w_{k-1} = (t_{k-1},\mathbf{p}_{k-1}),\ w_{k} = (t_k,\mathbf{p}_k)$
     \STATE // $\mathsf{v}_{i}$: expected velocity
     \STATE $\mathbf{s}_k[t_{k-1},\, t_k] \gets \texttt{CubicHermiteInterpolation}(w_{k-1},w_{k},v_{i})$
     \STATE $\varepsilon_k \gets \texttt{Eval}(\mathbf{s}_k[t_{k-1},\, t_k],\ \texttt{TeachOdom}[t_{k-1},\, t_k])$ // By \eqref{eq: total error}
     \STATE $\texttt{GlobalError} \gets \texttt{GlobalError} + \varepsilon_k$
\ENDFOR
 \IF{$\texttt{GlobalError} > \varepsilon$}  
    \FOR{$ k = 1$ to $|\mathcal{V}_{t} | - 1$}
         \IF{$\varepsilon_k > \varepsilon_{\text{seg}}$}
             \STATE $c_k \gets \texttt{CalculateCurvature}() $ // By \eqref{eq:3Dcurvature}
             \STATE $\mathcal{V}_{t}  \gets \mathcal{V}_{t}  \cup \texttt{argmax}(c_k)$
         \ENDIF
    \ENDFOR
\ENDIF
\STATE  $ \mathcal{V}_t \gets \{ (p_i, L_i) | (t_i, p_i, L_i) \in \mathcal{V}_t \}
$
\RETURN $\mathcal{V}_{t}$
\end{algorithmic}

\end{algorithm}

\subsection{Teaching Phase}
In the teaching phase, a teach graph is constructed to support robust trajectory repeat rather than global map optimization, since global consistency is not required for successful teach-and-repeat navigation \cite{krawciw2025local}. The teach graph is defined as $\mathcal{G} = (\mathcal{V}_{t}, \mathcal{E}_{t})$, where each node $v_{t}^{i}\in \mathcal{V}_{t}$ stores its associated teach LiDAR frame $\mathcal{L}_i$, and each directed edge $e_i \in \mathcal{E}$ connected from a teach node $v_{t}^{i-1}$ to its successive node $v_{t}^{i}$ and stores the relative transformation  $\mathbf{T}_{i-1}^{i}$ from the LiDAR frame $\mathcal{L}_{i-1}$ to $\mathcal{L}_{i}$,  which can be calculated using open-source LiDAR odometry algorithm such as LIO-SAM\cite{shan2020lio}.
Given synchronized LiDAR and 4D radar streams collected during the teach path, nodes are incrementally selected to construct the teach graph via an iterative two-stage selection mechanism. In such a way, we aim to construct a teach graph that has a minimal number of nodes, but still provides reliable cross-modal registration for teach-and-repeat navigation.

\begin{figure}[t]
    \centering
    \subfloat[Initial node selection]{
        \includegraphics[width=\linewidth]{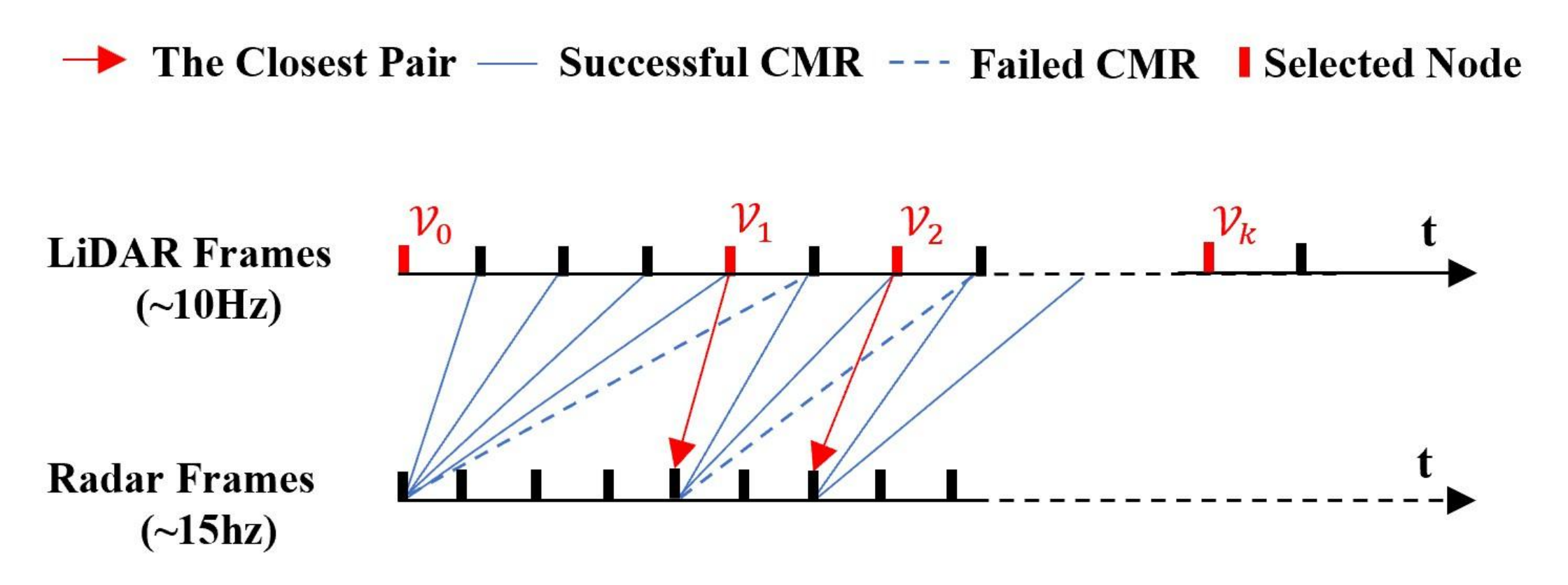}
    }\
    \subfloat[Iterative node addition]{
        \includegraphics[width=\linewidth]{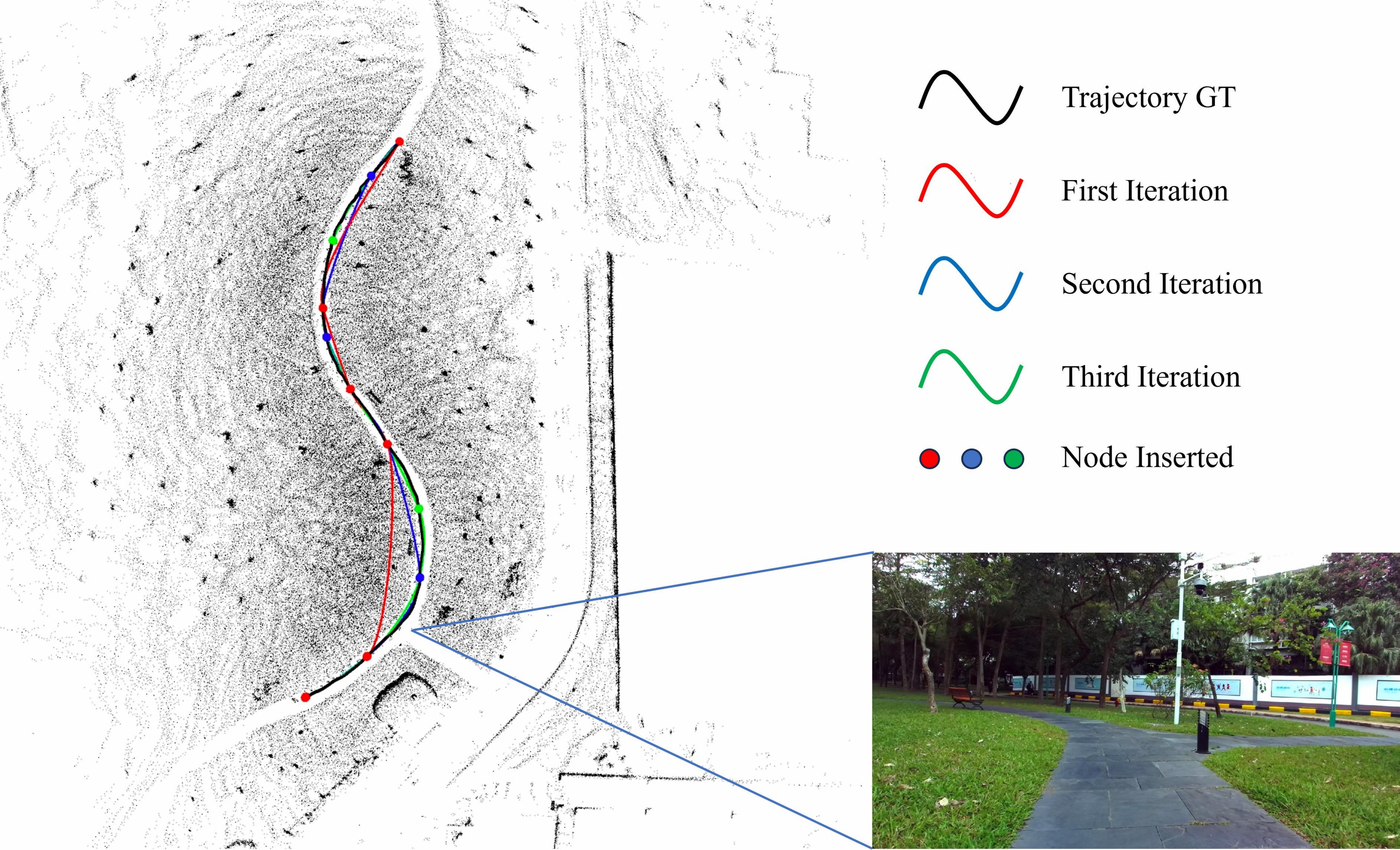}
    }
    \caption{\textbf{Iterative node selection.} (a) Initial node selection based on successful cross-modal registration between radar and LiDAR frames and LiDAR co-collected in the teaching phase. (b) Iterative node addition, where successive cubic Hermite curves refine the trajectory and new nodes are inserted in regions requiring higher spatial fidelity. The magnified view indicates the need for high-fidelity path interpolation  in the narrow scenario.}
    \label{fig:node_search}
\end{figure}

\textbf{Initial Node Selection:} As described in lines 1-17 of Algorithm \ref{alg:AdaptiveNodeSelection}, sparse nodes are selected in the first stage based on the CMR  quality. As shown in Fig.~\ref{fig:node_search}~(a), the nodes associated with the first and last LiDAR frames are firstly fixed as the start and end nodes. The CMR network is then used to register subsequent LiDAR frames to the first radar frame. Once a LiDAR frame and the first radar frame fail to register, the most recent LiDAR frame is added as a new node $v_1$. The radar frame closest to $v_{1}$ is then selected as the new reference radar frame, which is used to select the next node $v_2$ by repeating the previous procedure. Iteratively, it yields the initial compact set of sparse teach nodes $\{v_0, v_{1}, \dots,v_{k}\}$ together with the stored LiDAR frames at these nodes.

\textbf{Iterative Node Addition: } In the second stage, new nodes are added such that the trajectory error between the interpolated trajectory over nodes and the teach path is minimized. Given a sequence of initial node set $\mathcal{V}_{t}$, where each node provides a timestamp and pose $(\mathbf{t}_i,\mathbf{p}_i)$, that forms a strictly increasing sequence, a sliding-window cubic Hermite interpolation \cite{barfoot2024state} is employed to generate a continuous trajectory $\mathbf{s}(t)$ between consecutive nodes, as illustrated in Fig.~\ref{fig:node_search}~(b). 

For each time interval $[t_i,t_{i+1}]$, the interpolant is constrained to match the node positions and their directional derivatives:
\begin{equation}
\begin{aligned}
\mathbf{s}(t_i) &= \mathbf{P}_i, 
&\quad \dot{\mathbf{s}}(t_i) &= v_i\,\mathbf{d}_i,\\
\mathbf{s}(t_{i+1}) &= \mathbf{P}_{i+1},
&\quad \dot{\mathbf{s}}(t_{i+1}) &= v_i\,\mathbf{d}_{i+1}.
\end{aligned}
\end{equation}
where  $p_i=\begin{bmatrix}
 \mathbf{R}_i& \mathbf{P}_i\\  0&  1 
\end{bmatrix}$,  $\mathbf{P}_i\in\mathbb{R}^3$ is the position, and $\mathbf{R}_i\in SO(3)$ is  the rotation matrix of orientation, $\mathbf{d}_i$ is a unit direction vector extracted from $\mathbf{R}_i$, and $v_i$ is a user-defined speed. It ensures that the interpolated curve remains continuous in position and orientation at each node. 

Given the interpolated trajectory, an error evaluation is then conducted. Both positional error $e_{\text{pos}}(t_i)$ and rotational errors $e_{\text{rot}}(t_i)$ are calculated as below:
\begin{equation}\label{evaluation}
e_{\text{pos}}(t_i) = \|\mathbf{s}(t_i) - \mathbf{p}_{\textbf{orig}}(t_i)\|,
\end{equation}
where $\mathbf{e}(t_i)$ and $\mathbf{p}_{\text{orig}}(t)$ represent the position of the fitted interpolated trajectory and the teaching path at time $t_i$.
For orientation, let $\mathbf{R}_{\textbf{intp}}(t_i)$ and $\mathbf{R}_{\text{orig}}(t_i)$ be the orientation associated with the interpolated trajectory and the original teaching trajectory at $t_i$, respectively. The orientation error is calculated in Lie algebra space as:
\begin{equation}
e_{\text{rot}}(t_i) = \|\log\left(\mathbf{R}_{\textbf{intp}}(t_i)^\top \mathbf{R}_{\textbf{orig}}(t_i)\right)^\vee\|,
\end{equation}
where $\log(\cdot)$ maps a rotation matrix to its corresponding skew-symmetric form in $\mathfrak{so}(3)$, and $(\cdot)^\vee$ converts it to a rotation vector in $\mathbb{R}^3$.

Finally, the total error is defined as:
\begin{equation}\label{eq: total error}
E(t_i) = w_{\text{pos}} e_{\text{pos}}(t_i) + w_{\text{rot}} e_{\text{rot}}(t_i),
\end{equation}
where the weighting parameters $w_{\text{pos}}, w_{\text{rot}}>0$ are chosen based on application priorities. 


When the interpolation error of a trajectory segment exceeds the predefined threshold,  additional points are inserted with the largest curvature. The relative poses among LiDAR frames provided by LiDAR odometry are used to calculate the 3D curvature. Specifically, for every three consecutive frames, the 3D curvature $c_i$ at $p_i$ is calculated as below:
\begin{equation}
c_i = \frac{ \,\bigl\|(\mathbf{p}_{i} - \mathbf{p}_{i-1}) \times (\mathbf{p}_{i+1} - \mathbf{p}_{i})\bigr\|}{\|\mathbf{p}_{i} - \mathbf{p}_{i-1}\| \;\|\mathbf{p}_{i+1} - \mathbf{p}_{i}\| \;\|\mathbf{p}_{i+1} - \mathbf{p}_{i-1}\|},
\label{eq:3Dcurvature}
\end{equation}
where \(\mathbf{p}_{i} = (x_{i}, y_{i}, z_{i})\) denotes position of the $i$-th LiDAR frame, and \(\|\cdot\|\) denotes the Euclidean norm.

The trajectory interpolation and error evaluation process repeat iteratively until the overall trajectory error meets a user-defined threshold,  which is described in lines 19-36 of Algorithm \ref{alg:AdaptiveNodeSelection}. Once the nodes are determined, so are the edges, which means the teach graph is completed.

\begin{figure*}
    \centering
    \includegraphics[width=\linewidth]{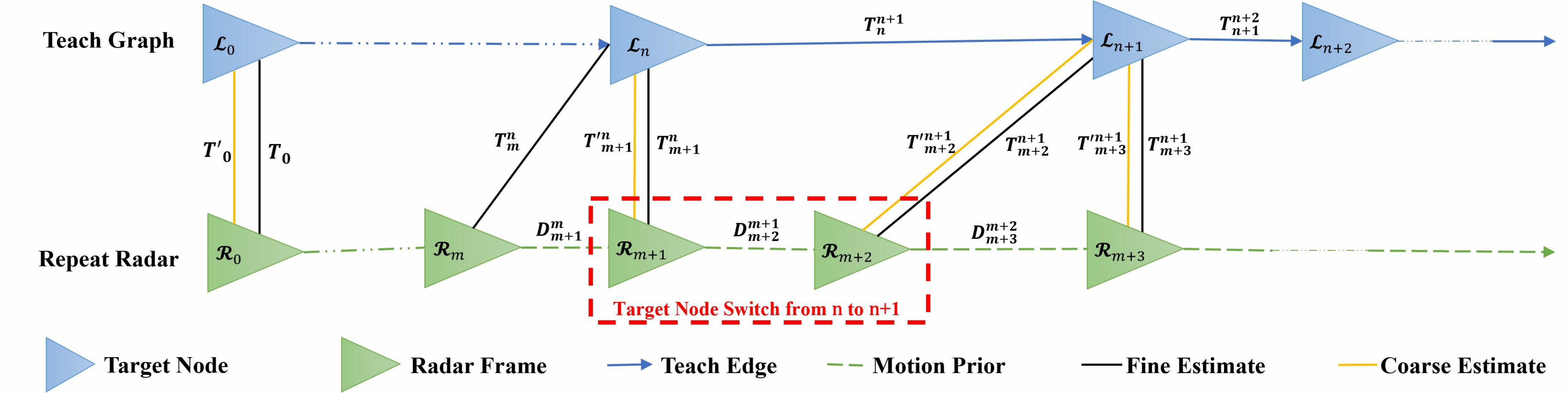}
    \caption{\textbf{Target Node selection during the repeating phase.} Green triangles represent radar frames during repeat. Blue triangles denote LiDAR-based nodes obtained during the teaching phase via LiDAR odometry. Target node change indicates a switch to the subsequent teach node for the latter radar frame. The black line denotes the fine estimate after CMR between the radar frame and the LiDAR frame in the target node, while the yellow line represents the coarse estimate before CMR.}
    \label{fig:pose_prior}
\end{figure*}

\subsection{Repeating Phase} \label{Doppler prior}
In the repeating phase, the robot searches for the target node from the teach graph and then repeats the teach path by tracking the target nodes. By repeating the path, a repeat graph is constructed incrementally to connect the target node to the repeat node associated with the radar frame. The repeat graph is defined as $\mathcal{G} = (\mathcal{V}_{t}, \mathcal{V}_{r}, \mathcal{E}_{r})$, where each node $v_{r}^{j}\in \mathcal{V}_{r}$ stores the radar frame $\mathcal{R}_j$ collected at the node, and each directed edge $e_j \in \mathcal{E}$ connected from the repeat node $v_{r}^{j}$ to its target node $v_{t}^{i}$ and  stores the relative transformation  $\mathbf{T}_{j}^{i}$ from  the radar frame $\mathcal{R}_{j}$ to the LiDAR frame $\mathcal{L}_{i}$, which is calculated by CMR network.

The target node selection is initialized via CMR-based place recognition, where the first 4D radar frame is registered against all teach nodes, and the teach node yielding the minimal transformation is selected as the first target node. Once the first target node is established, it proceeds to match each incoming 4D radar frame to the appropriate node in the teach graph. The entire target node searching process consists of two steps: (i) target node switch and (ii) cross-modal registration.

\textbf{Target Node Switch:}
Given that the teach node $v^n_t$ is selected as the target node for the repeat node $v^{m+1}_r$, the relative pose of the radar frame $\mathcal{R}_{m+1}$ to the teach LiDAR frame $\mathcal{L}_{n}$ is obtained as $\mathbf{T}_{m+1}^{n}$. If $\|\mathbf{T}_{m+1}^{n}\|\geq \delta$, then the incoming radar frame $\mathcal{R}_{m+2}$ will be kept aligned to the same LiDAR frame $\mathcal{L}_n$, where $\|\cdot\|$ denotes the norm of a matrix; Otherwise, it will switch to the LiDAR frame in the subsequent teach node $v^{n+1}_t$. 

As illustrated in Fig.~\ref{fig:pose_prior}, a target node switch is triggered when it goes from the repeat node $v^{m+1}_r$ to a new node $v^{m+2}_r$.  Before cross-modal registration happens, a coarse estimate $\mathbf{T}_{m+2}^{\prime n+1}$ of the relative pose from  the radar frame $\mathcal{R}_{m+2}$ to the LiDAR frame $\mathcal{L}_{n+1}$ is calculated as below:
\begin{equation}\label{eq:regR_m2}
\mathbf{T}_{m+2}^{\prime n+1} = \mathbf{T}_n^{n+1}\mathbf{T}_{m+1}^{n}(\mathbf{D}_{m+1}^{m+2})^{-1}.
\end{equation}
where $\mathbf{T}_n^{n+1}$ is the relative transformation stored in edges of the teach graph, and $D_{m+1}^{m+2}$ is the Doppler-based motion prior of the transformation between the radar frames $\mathcal{R}_{m+1}$ and $\mathcal{R}_{m+2}$, which is computed as detailed below.

Based on a constant velocity model between the two consecutive radar frames and the ego-velocity estimation method using Doppler information proposed in \cite{doer2020doppler_prior}, we have:

\begin{equation}
   \mathbf{v}_r = \left(\mathbf{H}^T \mathbf{H}\right)^{-1} \mathbf{H}^T \mathbf{v}_{\text{Doppler}}.
\end{equation}
\begin{equation}\label{eq:dopper_transform}
    D_{m+1}^{m+2} = \frac{\mathbf{v}_r}{f},
\end{equation}
where \(\mathbf{v}_{\text{Doppler}}\) is the Doppler velocity of each inlier point obtained via the RANSAC algorithm in the radar frame $\mathcal{R}_{m+1}$, $v_r$ is the ego-velocity of the 4D radar sensor, $f$ is the frequency of 4D radar, and \(\mathbf{H} = \begin{bmatrix} \hat{\mathbf{r}}_1^\top & \hat{\mathbf{r}}_2^\top & \cdots & \hat{\mathbf{r}}_N^\top \end{bmatrix}^\top\) is constructed by stacking the transposed normalized direction vectors \(\hat{\mathbf{r}}_i\) of each radar point. Here, \(\mathbf{H}\) captures the geometric relationship between the radar points' positions and the sensor's movement.

If the node switch does not occur, a coarse relative pose estimate is obtained by directly augmenting the latest pose estimate with the Doppler-based motion prior. As illustrated in Fig.~\ref{fig:pose_prior}, the coarse relative pose estimate for radar frame $R_{m+3}$ to LiDAR frame $L_{n+1}$ is obtained by augmenting $\mathbf{T}_{m+2}^{n+1}$ with the Doppler-based motion prior $\mathbf{D}_{m+2}^{m+3}$:
\begin{equation}\label{eq:regR_m3}
\mathbf{T}_{m+3}^{\prime n+1} = \mathbf{T}_{m+2}^{n+1} (\mathbf{D}_{m+2}^{m+3})^{-1}.
\end{equation}

\textbf{Registration via CMR:}
Given the coarse relative pose estimate $\mathbf{T}_{m+2}^{\prime {n+1}}$ and the LiDAR frame $\mathcal{L}_{n+1}$ and radar frame $\mathcal{R}_{m+2}$, the relative pose between $\mathcal{L}_{n+1}$  and $\mathcal{R}_{m+2}$ is estimated using the CMR network:
\begin{equation}\label{eq:regR_m2_continue}
\mathbf{T}_{m+2}^{n+1} = \mathrm{CMR}\Bigl(\mathbf{T}_{m+2}^{\prime n+1},L_{n+1}, R_{m+2}\Bigr),
\end{equation}
Following these steps, the CMR network outputs the fine relative pose estimate during the repeat phase.

\textbf{Trajectory Tracking:}
After the target node is selected, an arc-based path that connects the robot's pose to the target node is constructed. To simplify control design, we adopt the pure pursuit geometric guidance strategy \cite{purepursuit}, which maintains a fixed ratio between linear and angular velocities. 

Given the arc, a PD controller computes linear and angular velocities for trajectory tracking:
\begin{equation}\label{arc}
Arc = \frac{v_{i}}{\omega} = \frac{\|\mathbf{t}\|}{2\sin(\theta)},
\end{equation} 
where $v_{i}$ is the expected velocity, $\mathbf{t} = (t_x,t_y,t_z)^\top$ is the translation vector extracted from the CMR output $T_m^n$, $ \|\mathbf{t}\|$ is its Euclidean norm, and $\theta = \operatorname{atan2}(t_y,\, t_x)$ is the yaw-only bearing angle following the standard pure pursuit geometry.


\begin{algorithm}[t]
\caption{Adaptive Incremental CMR Fine-Tuning}
\label{alg:AdaptiveCMRFineTuning}
\footnotesize
\begin{algorithmic}[1]

\STATE {//Evaluation and Sampling}

\FOR{{each} radar frame $R$ {in the repeating phase}}
    \STATE {Obtain LiDAR node $L$}
    \STATE {$T_{l}^{r} \gets \mathrm{CMR}(T',L, R)$} // $T'$ recorded by \eqref{eq:regR_m2} and \eqref{eq:regR_m3}
    \STATE {$T_{base}^{r} \gets T_{base}^{l}T_{l}^{r}$          // Global transformation}
    \STATE {$\epsilon \gets $\texttt{NearestNeighbour($T_{base}^{r}$, TeachOdom)}}
    \STATE $(\mu_i,\sigma_i) \gets \texttt{SlidingWindowsStatistics()}$
    \IF{\texttt{$a\sigma_i-\mu_i-b>0$}} 
    \STATE // $a,b$ is empirically set from Fig.~\ref{fig:ATE_classify}.
        \STATE {Initialize $\mathcal{I}_{ft}$}
        \STATE $\mathcal{I}_{{neg}}, \mathcal{I}_{{pos}} \gets$ \texttt{SampleSelection()}
        \STATE $\mathcal{I}_{ft} \gets \mathcal{I}_{{neg}} \cup \mathcal{I}_{{pos}}$
    \ENDIF
\ENDFOR
\STATE {// Self-supervised fine-tune}
\IF{$\max_{i \in \mathcal{I}_{\text{neg}}}\mu_i > \tau_{\mu} \;\text{or}\; \sum_{i \in \mathcal{I}_{\text{neg}}}\ell_i > L_{\min}$}
    \FOR{$L^i \in \mathcal{I}_{ft}$}
        \FORALL{$R$ {match to $L^i$}}
            \STATE ${R}_{1}^i$, ${R}_{2}^i \gets$ \texttt{RandomSampling($R$)}
            \STATE {Compute loss $\mathcal{L}_{ft}$ by \eqref{ft5}}
            \STATE {Update CMR network parameters}
        \ENDFOR
    \ENDFOR
\ENDIF
\end{algorithmic}
\end{algorithm}

\subsection{Finetuning Phase}\label{ft}
This subsection presents a strategy to adaptively fine-tune the CMR network in response to static environmental changes. We first reveal that static changes are separable from other kinds of ephemeral changes. Based on this insight, a fine-tuning strategy that consists of evaluation and sampling, and self-supervised fine-tuning is proposed. Once a trajectory repeat is completed, the trajectory is evaluated to select and store samples in a library. If the finetuning criterion is met, all the samples accumulated in the library are used to fine-tune the CMR network in a self-supervised manner. The overall procedure is summarized in Algorithm \ref{alg:AdaptiveCMRFineTuning}.

\begin{figure}[t]
    \centering
    {  \includegraphics[width=\linewidth]{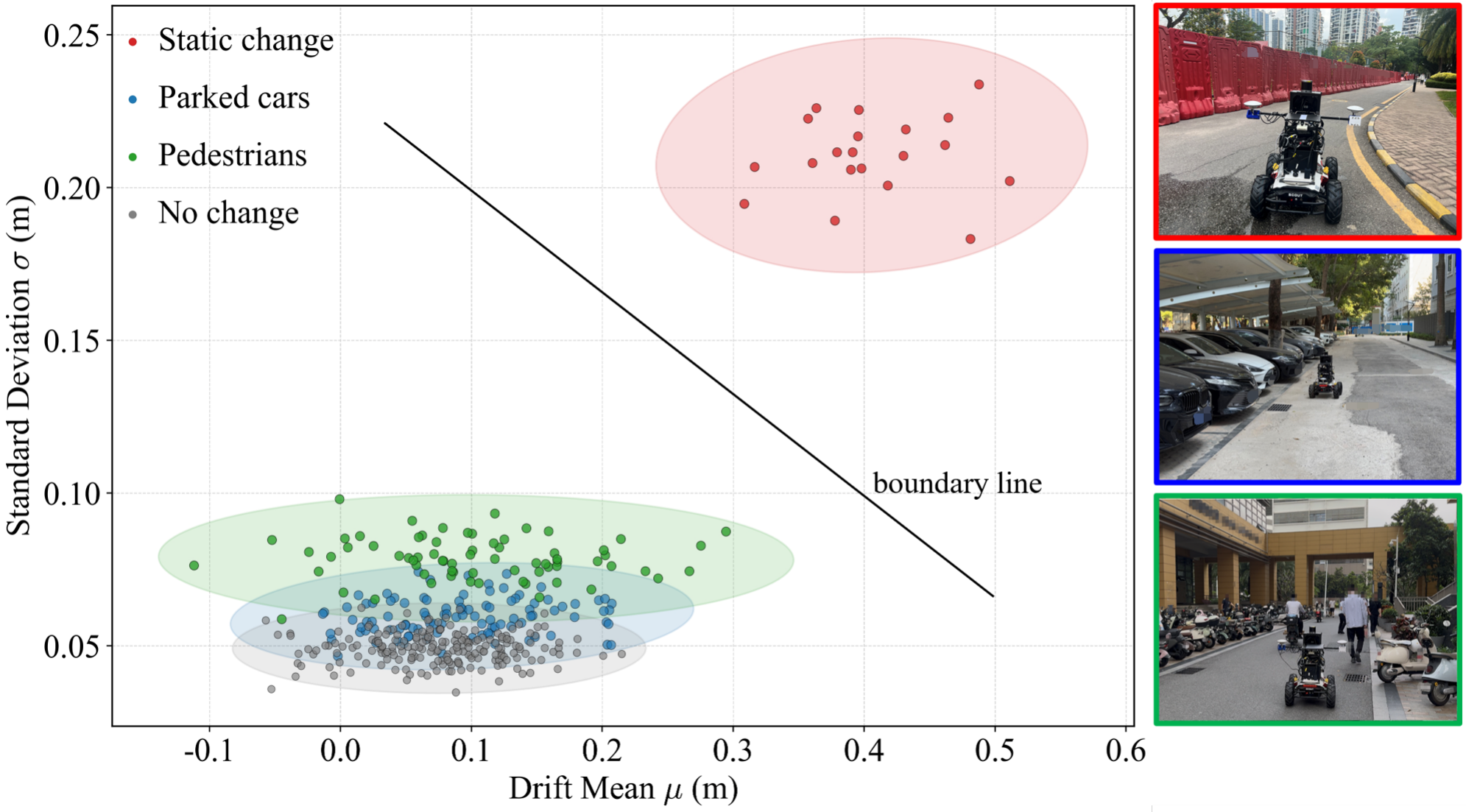}
    }\caption{\textbf{Gaussian distribution of ATE under different environment changes.} The black line is the linear decision boundary that isolates the error distribution due to static changes (red)  from the other three clusters. Representative scene images for every class are shown on the right.}\label{fig:ATE_classify}
    \end{figure}

\textbf{Separability of Environmental Changes:} Experiments are conducted to analyze how different types of environmental changes impact the navigation performance of LTR$^2$ system. As shown in Fig. \ref{fig:ATE_classify}, different types of environmental changes induce different drift distributions. In particular, static environmental changes result in significantly higher average trajectory error (ATE) than ephemeral changes such as parked cars and pedestrians, suggesting their separability. This critical observation underpins our finetuning strategy, which specifically targets static changes to increase the reliability of LTR$^2$ in long-term navigation.

\begin{figure}[t]
    \centering  \includegraphics[width=\linewidth]{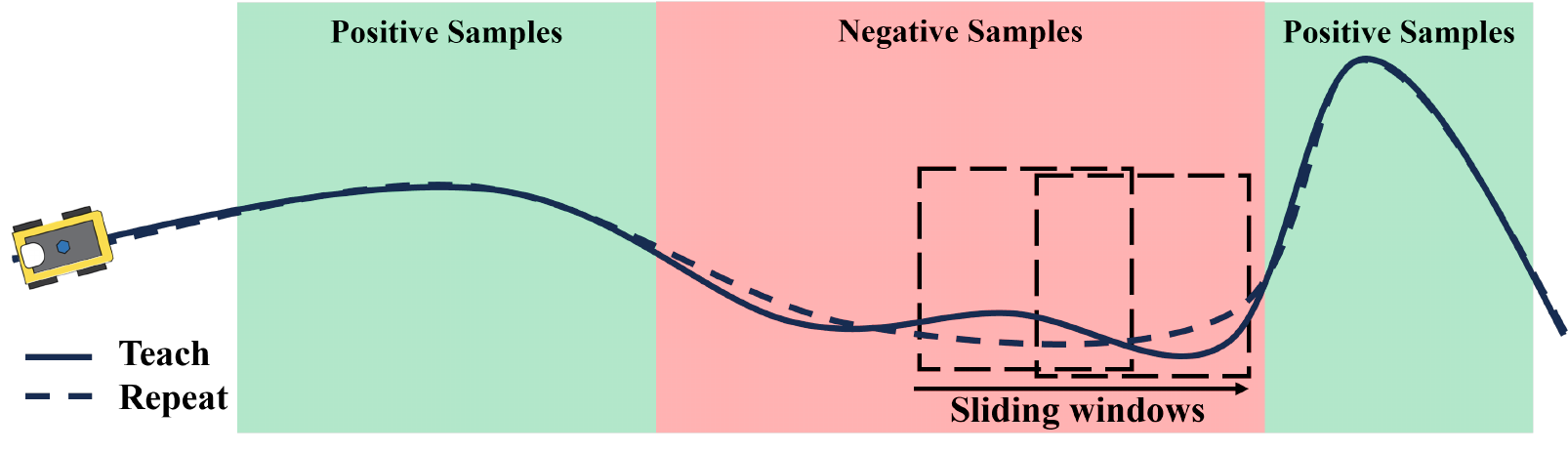}
        \caption{\textbf{Positive and negative samples collected in sliding windows.} Positive samples are segments where the repeat trajectory remains consistent with the teach trajectory, while negative samples correspond to windows exhibiting abnormal drift. The sliding-window evaluation partitions the trajectory into these two sample types for subsequent fine-tuning.}
        \label{fig:samples}
\end{figure}

\textbf{Evaluation and Sampling:}
After each trajectory repeat, the local transformation $T_{l}^{r}$ calculated by the CMR registration is transformed to the base coordinate system $T_{base}^{r}$: 
\begin{equation}\label{eq:pose of radar on teach LiDAR}
    \mathbf{T}_{base}^{r}= \mathbf{T}_{base}^{l}\mathbf{T}_{l}^{r},
\end{equation}
where $\mathbf{T}_{base}^{l}$ is the teach odometry in the base coordinate. Then, each radar is matched to its nearest neighboring node in the teaching trajectory, and the Euclidean distance is used as the drift metric. To identify segments exhibiting abnormal drift, a sliding window-based evaluation is performed. For each window, we compute the mean $\mu_i$ and standard deviation $\sigma_i$ of localization drift. As illustrated in Fig. \ref{fig:samples}, negative samples are selected from windows that satisfy the following linear classifier criteria:
\begin{equation}\label{ft3}
    \mathcal{I}_{\textrm{neg}} = \left\{ (L, R) \in w_i \mid  a\sigma_i-\mu_i-b>0\right\},
\end{equation}
where $R$ is the radar frame and $L$ is the tracked LiDAR frame of $R$, $a$ and $b$ are the hyperparameters related to the boundary line, empirically set from Fig.~\ref{fig:ATE_classify}. 

Overall, the resulting sample dataset $\mathcal{I}_{\textrm{ft}}$ is updated by:
\begin{equation}\label{ft4}
\mathcal{I}_{\textrm{ft}} =  \mathcal{I}_{\textrm{ft}} \cup \mathcal{I}_{\textrm{neg}} \cup \mathcal{I}_{\textrm{pos}},
\end{equation}
where a proper ratio $r$ between positive $\mathcal{I}_{\textrm{pos}}$ and negative samples $\mathcal{I}_{\textrm{neg}}$ is kept to prevent catastrophic forgetting of the scene in memory.

\textbf{Fine-tuning decision rule:}
Fine-tuning of the CMR network is triggered only when static drift is sufficiently significant. Formally fine-tuning is performed if 
\[
\max_{i \in \mathcal{I}_{\text{neg}}}\mu_i > \tau_{\mu} \;\textrm{or} \sum_{i \in \mathcal{I}_{\text{neg}}}\ell_i > L_{\min},\] where \(\tau_{\mu}\) is the threshold of maximum drift among negative samples and \(L_{\min}\) is the minimum total trajectory length affected by \(\mathcal{I}_{\text{neg}}\). The thresholds are set empirically to avoid unnecessary fine-tuning.

\textbf{Self-supervised Finetuning:} The CMR network update is trained in a self-supervised manner that enforces geometric and structural consistency across radar frames, eliminating the need for ground-truth annotations. From the sample dataset, we randomly sample radar frame pairs  ${R^1, R^2}$ that are associated with the same LiDAR reference frame.  They are then transformed into the LiDAR coordinate system to establish geometric correspondence. 

A geometric consistency loss $\mathcal{L}_{\text{geo}}$ is introduced to directly measure the Euclidean distance between corresponding points in the two projected radar frames, which is defined as:
\begin{equation} 
\mathcal{L}_{\textrm{geo}} = \frac{1}{N} \sum_{i=1}^{N} \left\| \mathbf{T}_i^{1} \cdot \mathbf{r}^{1} - \mathbf{T}_i^{2} \cdot \mathbf{r}^{2} \right\|_2^2, \label{eq:geo_loss} \end{equation}
where $\mathbf{r}_i^{1}, \mathbf{r}_i^{2}\in\mathbb{R}^{4}$ are a pair of corresponding radar points obtained by nearest-neighbor matching after projection into the LiDAR coordinate frame, expressed in homogeneous coordinates. $\mathbf{T}_{i}^{1}$ and $\mathbf{T}_{i}^{2}$ are the transformations of radar points to the LiDAR coordinate frame, and $N$ is the total number of matched point pairs.

The projected radar frames are further converted into pseudo-image representations to calculate the structural consistency loss $\mathcal{L}_{\textrm{SSIM}}$.

Both geometric consistency and structural consistency are integrated into a unified loss function for incremental fine-tuning:
\begin{equation}\label{ft5}
\mathcal{L}_{{ft}} = \lambda \cdot \mathcal{L}_{\textrm{SSIM}} + (1 - \lambda) \cdot \mathcal{L}_{\textrm{geo}} ,
\end{equation}
where $\lambda$ is a weighting coefficient balancing the structural and geometric terms.

\section{End-to-end Cross-Modal localization}\label{Cross-Modal localization}
In this section, we present the framework of cross-modal registration between 4D radar and LiDAR from two perspectives: data preprocessing and CMR network design. 

\subsection{Data Preprocessing}
\begin{figure*}[htbp]
    \centering
    \subfloat[Visualization of Intensity Aligned Radar in Cylindrical Projection]{
        \includegraphics[width=\linewidth]{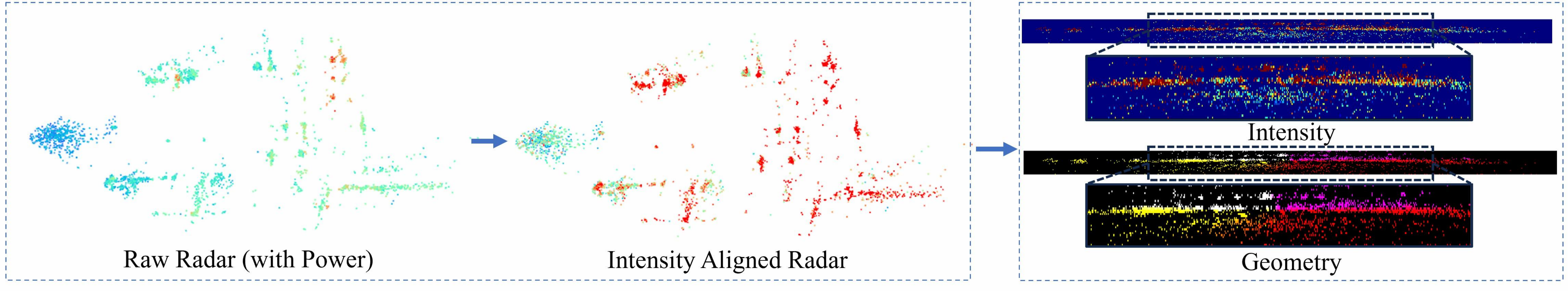}
    }\
    \subfloat[Visualization of FOV Cropped LiDAR in Cylindrical Projection]{
        \includegraphics[width=\linewidth]{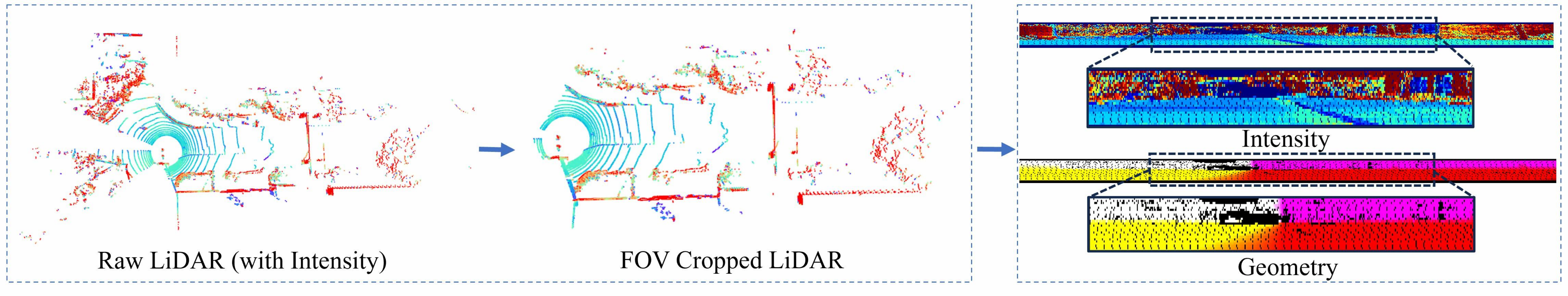}
    }
    \caption{\textbf{Data preprocessing for 4D radar and LiDAR.} (a) Raw radar with power is aligned with the intensity of LiDAR and followed by two separate cylindrical projections in the intensity and geometry. (b) Raw LiDAR is cropped using the 4D radar FOV mask and then transformed into the cylindrical projection.}
    \label{Power-Intensity_alignment}
\end{figure*}

In data preprocessing, we align the radar and LiDAR data and then transform them into a unified representation, facilitating cross-modal registration. A mask is first constructed using the coarse estimate of relative pose by \eqref{eq:regR_m2} to align the omnidirectional LiDAR FOV with the forward-looking 4D radar. Then we align the raw data in two dimensions: one in point cloud geometry, the other in normalized radar power density versus LiDAR intensity. Finally, the aligned data is transformed into the cylindrical projection in both dimensions. The entire process is depicted in Fig.~\ref{Power-Intensity_alignment}.

\textbf{LiDAR FOV Cropping: }Given the coarse estimate of relative pose by \eqref{eq:regR_m2}, we further enlarge the radar’s masking region to account for inaccuracies in the pose estimate. Specifically, we modestly broaden the radar’s FOV around the predicted pose and then project such an expanded area onto the raw LiDAR data, ensuring that sufficient LiDAR points remain available even if the coarse pose estimate deviates from the true pose.

\begin{figure}[t]
    \centering
    \includegraphics[width=\linewidth]{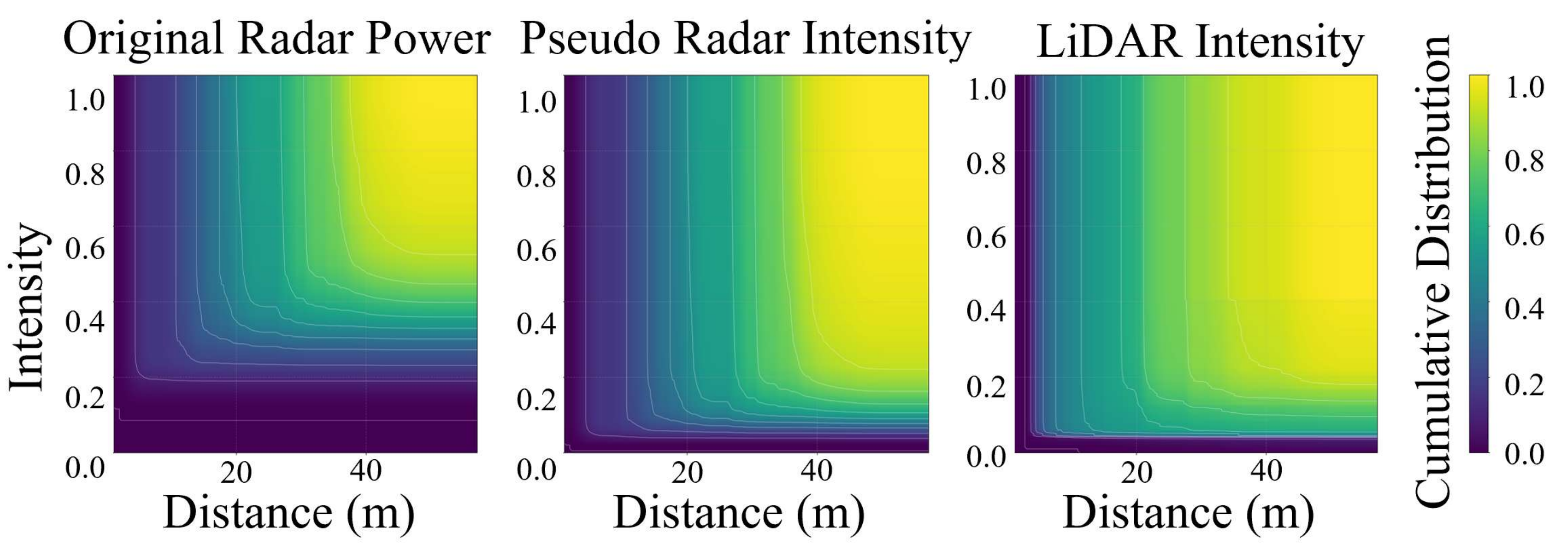}
    \caption{\textbf{Cumulative distribution heatmaps of intensity versus distance} for the original radar (left), the pseudo radar aligned to LiDAR intensity via \eqref{eq:radar2} (middle), and the LiDAR signals (right). With the intensity alignment, the discrepancy in intensity distributions between pseudo radar and LiDAR is significantly less than that between the original radar and LiDAR.}
    \label{fig:preprocess_result}
    \vspace{-3mm}
\end{figure}

\textbf{Radar Intensity Alignment:} In this subsection, we analyze the relationship between the power density of 4D radar and the reflected intensity of LiDAR, and then introduce a method to transform the radar power density to LiDAR intensity-like pseudo radar intensity. The received power density $P_\textrm{radar}$ of a 4D radar is given by \cite{skolnik1970radar}:
\begin{equation}\label{eq:radar}
    P_\textrm{radar} = \frac{P_t G_t G_r A^2 \cos^2\theta}{(4\pi)^2 \lambda^2 d^4 \eta},
\end{equation}
where $P_t$ is the transmitted power, $G_t$ and $G_r$ are the gains of the transmitting and receiving antennas, respectively,  $d$ is the range between the radar and the target, and $\eta$ represents the system loss factor, $A$ is the effective amplitude of the target, connecting the target reflectance characteristics, $\theta$ is the observation angle, and $\lambda$ is the wavelength of the signal. 

The intensity of LiDAR point cloud is modelled by \cite{fujii2005laser}:
\begin{equation}\label{eq:lidar}
    P_{\text{LiDAR}} = \frac{P_t A_r \rho(\theta)}{d^2} e^{-2\alpha d},
\end{equation}
where $P_{\text{LiDAR}}$ is the received power density, $P_t$ is the transmitted power, $A_r$ is the effective aperture area of the receiving antenna which is directly related to the physical diameter of the receiving antenna, $\rho(\theta)$ is the target's reflection coefficient dependent on the angle $\theta$, $d$ is the distance between the LiDAR and the target, and $e^{-2\alpha R}$ accounts for atmospheric attenuation with $\alpha$ as the attenuation coefficient. For an ideal diffuse reflection surface (Lambertian body), the reflectivity is modeled as $\rho(\theta) = \rho_0 \cos\theta$, where $\rho_0$ is the maximum reflection coefficient at normal incidence, and $\cos\theta$ captures the angular dependence, representing the attenuation of reflection with the incident angle. 

Noting that all the parameters in \eqref{eq:radar} and \eqref{eq:lidar} can be categorized to two classes: ones related to spatial locations such as $d$, and $\theta$ and the others not,  We are inspired to reformulate \eqref{eq:radar} and \eqref{eq:lidar} as below:
\begin{equation}
    P_\textrm{radar} = K' \frac{A^2 \cos^2\theta}{d^4},
    \label{power_final}
\end{equation}
where the constant scalar $K' = \frac{P_t G_t G_r}{(4\pi)^2 \lambda^2  \eta}$.

\begin{equation}\label{eq:lidar2}
    P_{\text{LiDAR}} = K'' \frac{\rho_0\cos\theta}{d^2} e^{-2\alpha d},
\end{equation}
where the constant scalar constant $K'' = P_t A_r$. 

Then we construct a pseudo radar intensity as below: 
\begin{equation}\label{eq:radar2}
    \tilde{P}_{\text{radar}} = \sqrt{P_{\text{radar}}} \cdot e^{-2\alpha d}=\sqrt{K'}\frac{A\cos\theta}{d^2}e^{-2\alpha d},
\end{equation}
where we employ an empirical atmospheric attenuation coefficient $\alpha = 0.03\,\text{km}^{-1}$, representative of clear-air conditions under which the teaching phase happens.

It is straightforward from \eqref{eq:lidar2}  and \eqref{eq:radar2} that 
\begin{equation}\label{eq:normalize}
    \frac{\tilde{P}_{\text{radar}}}{P_{\text{LiDAR}}}=\frac{\sqrt{K'}A}{K''\rho_0}.
\end{equation}

Given the same working scenarios of LiDAR and radar sensors, the ratio $\frac{\tilde{P}_{\text{radar}}}{P_{\text{LiDAR}}}$ is only related to object materials but independent of the locations of objects, implying similar intensity distribution over space between the pseudo radar and LiDAR, which is experimentally validated in Fig. \ref{fig:preprocess_result}.

\textbf{Cylindrical Projection:}
After LiDAR FOV cropping and pseudo radar generated by intensity alignment, both data are normalized separately and transformed into cylindrical projections to represent their respective geometric and intensity distributions, as illustrated in Fig. \ref{Power-Intensity_alignment}. The cylindrical projection maps 3D point clouds onto a 2D pseudo-image by leveraging the spherical characteristics of LiDAR sensors. Each 3D point with coordinates \((x, y, z)\) is transformed into a 2D pixel position \((u, v)\) using the following equations:
\begin{equation}
\begin{gathered}
u = \arctan2(y, x)/\Delta \theta, \\
v = \arcsin\left(\dfrac{z}{\sqrt{x^2 + y^2 + z^2}}\right)/\Delta \phi,
\end{gathered}
\end{equation}
where \(\Delta \theta\) and \(\Delta \phi\) are the horizontal and vertical resolutions of the LiDAR sensor, respectively. The resulting 2D pseudo-image has dimensions \(H \times W \times 3\). For geometry projection, each pixel is filled with the corresponding raw 3D coordinates \((x, y, z)\), while for intensity projection, each pixel fills a single channel with the normalized intensity.

\begin{figure*}[htbp]
    \centering
    \subfloat[CMR network architecture.\label{fig:CMR_1}]{
        \includegraphics[width=2\columnwidth]{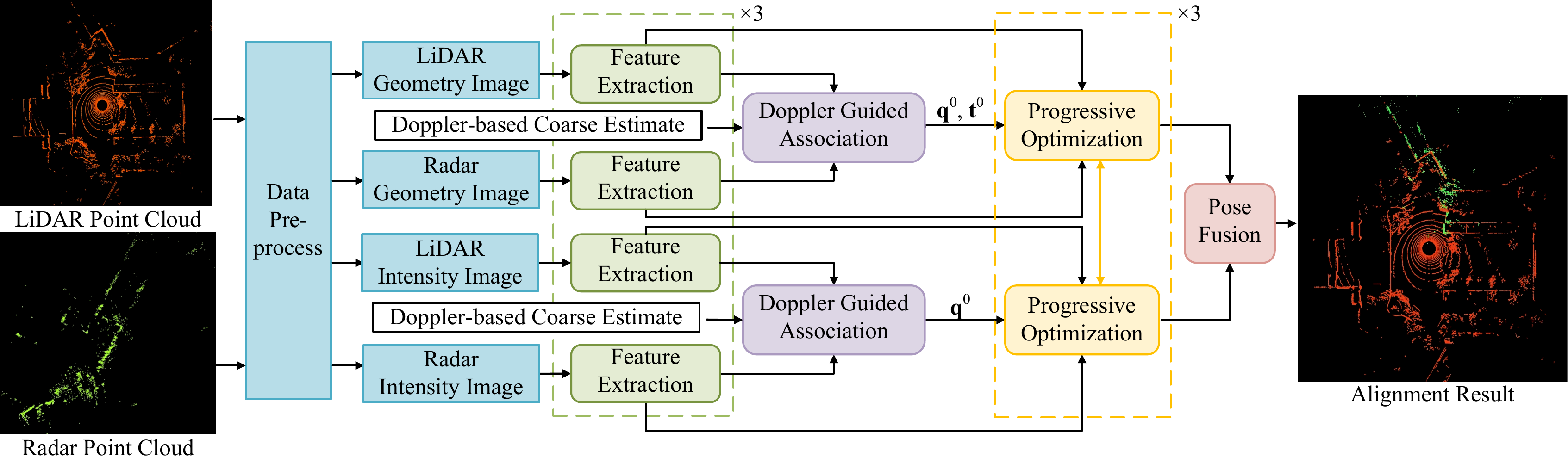}} \\
    \centering
    \subfloat[Doppler guided association block.\label{fig:CMR_2}]{
        \includegraphics[height=100pt]{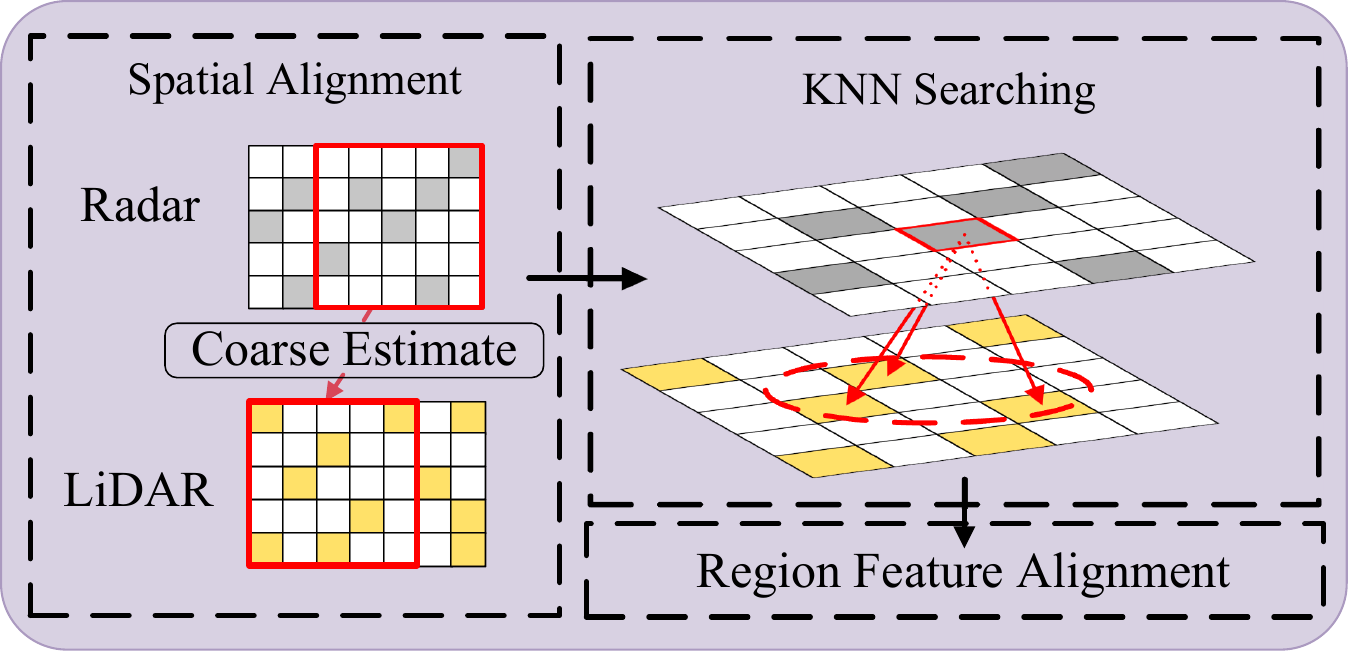}} \vspace{10pt}
    \subfloat[Progressive optimization block.\label{fig:CMR_3}]{
        \includegraphics[height=100pt]{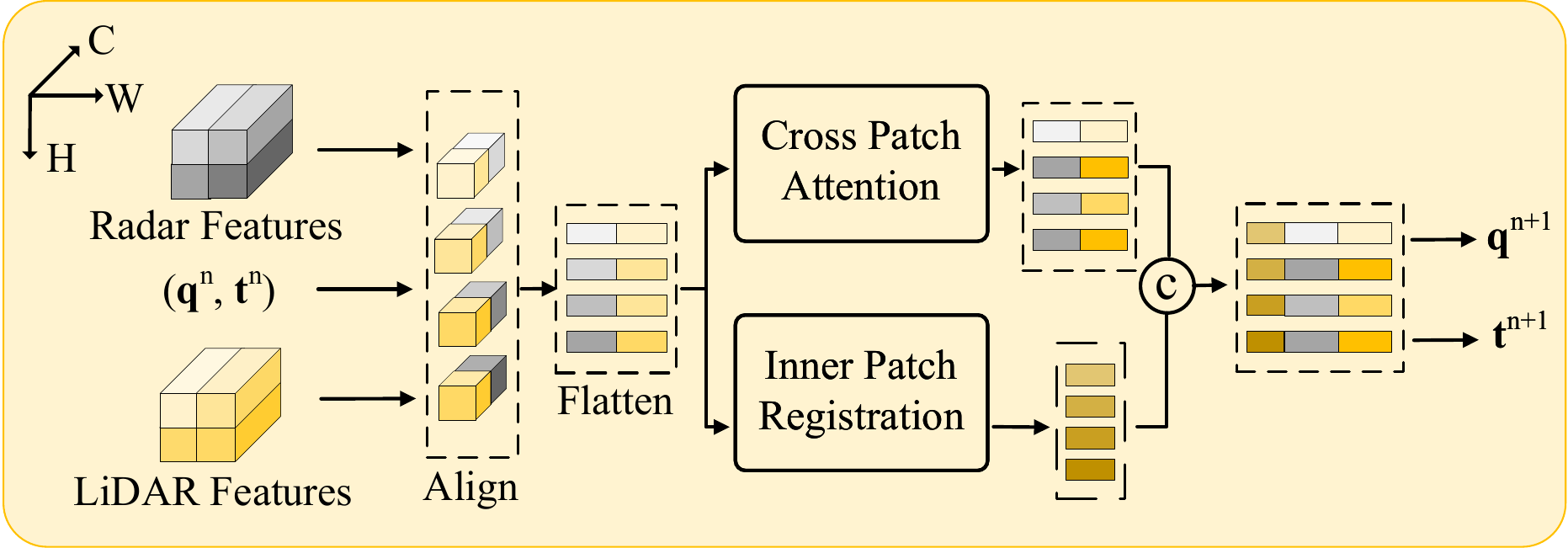}}
    \caption{\textbf{Overall structure of the CMR network.} Preprocessed geometry and intensity images are fed into separate branches,  and features are extracted by a Swin Transformer.  The Doppler-guided association module exploits the coarse estimate to generate cross-modal correspondences. Then the cross-modal attention and progressive optimization refine alignment, and the final registration is achieved through the pose fusion block.}
    \label{fig:backbone}
\end{figure*}

\subsection{LiDAR-Radar Registration Network}
The CMR network aligns the preprocessed LiDAR and radar images through a pipeline of feature extraction, Doppler guided association, and progressive optimization, as shown in Fig. \ref{fig:backbone} (a). 

The geometry and intensity feature extraction are conducted in two parallel branches, each employing Point Swin Transformer\cite{liu2021swin} to capture modality-specific representations. Although both branches share comparable network architectures, the intensity branch focuses exclusively on intensity distributions and excludes geometric cues, thus consistently reinforcing rotational alignment. For clarity, the following discussion details the geometry branch.

\textbf{Geometry-Intensity Swin Transformer:}
We employ a Swin Transformer, which has a complexity that scales linearly with image size, to extract features from geometric and intensity images. In each transformer block, the Window-Based Multi-Head Self-Attention (W-MSA) divides the image into several equally sized windows and computes attention within each window; the Shifted Window-Based Multi-Head Self-Attention (SW-MSA) introduces cross-window connections by shifting the window partitions of W-MSA. The feature extraction block can be represented as:
\begin{equation}
\centering
\begin{aligned}
    \hat{f}^l &= \text{W-MSA} \left( \text{LN} \left( f^{l-1} \right) \right) + f^{l-1}, \\
    f^l &= \text{MLP} \left( \text{LN} \left( \hat{f}^l \right) \right) + \hat{f}^l, \\
    \hat{f}^{l+1} &= \text{SW-MSA} \left( \text{LN} \left( f^l \right) \right) + f^l, \\
    f^{l+1} &= \text{MLP} \left( \text{LN} \left( \hat{f}^{l+1} \right) \right) + \hat{f}^{l+1},
\end{aligned}
\end{equation}
where $f^{l-1}$ and $f^{l+1}$ denote the input and output features of the $l$-th block, LN denotes the Layer Normalization, and MLP denotes the Multi-Layer Perceptron.

The computation process of attention follows the traditional approach with positional encoding:
\begin{equation}
    Attn({Q}, {K}, {V}) = softmax(\frac{{QK}^{\mathrm{T}}}{\sqrt{\mathcal{D}_k}} + B){V},
\end{equation}
where \(Q, K, V\) are derived from the input feature map $f$ through linear mapping, \(D_k\) represents the dimensionality of \(K\), and \(B\) is the relative position bias. We apply a mask \(m\) to identify the blank regions in the images so that they are excluded during the attention computation:
\begin{equation}
    m = \begin{cases}
        0, & \text{where points exist} \\
        -\infty, & \text{where is blank}
    \end{cases}.
\end{equation}

Since the geometric and intensity images are generated by projecting each point $(x_i,y_i,z_i)$ onto a defined pixel coordinate $(u_i,v_i)$, each valid pixel in the feature map corresponds uniquely to a point. We construct the pixel channel of the feature extraction module output $\textbf{F}$ by concatenating the corresponding point coordinate with the feature map:
\begin{equation}
    \mathbf{F}(u_i,v_i) =
    \big[x_i, y_i, z_i \big] \oplus f(u_i, v_i).
\end{equation}
For notational convenience, we define $f_i = f(u_i, v_i)$ to represent the feature belonging to point $i$.

\textbf{Doppler Guided Association:}
As illustrated in Fig.~\ref{fig:backbone} (b), we introduce the Doppler guided association block to quickly converge the registration result. Because the pixel channel retain the original 3D coordinates, the coarse pose estimate in \eqref{eq:regR_m2} can be directly applied to transform these coordinates, yielding an initial alignment between the radar and LiDAR projections.
With this initial alignment, a nearest neighbor search is performed in the pixel coordinate system, which constrains the search space to a fixed region rather than the entire area of the image. This locality constraint greatly accelerates feature association.

Specifically, given the radar point cloud coordinates $\{r_i\}_{i=1}^n$ and the corresponding features $\{f^{radar}_i\}_{i=1}^n$, the LiDAR point cloud coordinates $\{l_j\}_{j=1}^m$ and the corresponding features $\{f^{lidar}_j\}_{j=1}^m$, we calculate the attention of $r_i$ and the $k$ nearest points $\{l_1,l_2,...l_k\}$ within a certain range: 
\begin{equation}
\small
    \hat{f}^{radar}_i = Attn(f^{radar}_iM^Q, \{f^{lidar}\}_kM^K, \{f^{lidar}\}_kM^V),
\end{equation}
where $ \{f^{lidar}\}_k$ are features of $\{l_1,l_2,...l_k\}$, $M^Q, M^K, M^V$ are weight matrices. Afterward, we repeat this process by making the LiDAR point clouds the query vector. The obtained features are concatenated with coordinates to get $\{r_i\oplus \hat{f}^{radar}_i \}_{i=1}^n$, $\{l_j\oplus \hat{f}^{lidar}_j \}_{j=1}^m$. Then, a pose estimate $(\mathbf{q}^{0}, \mathbf{t}^{0})$ of point cloud registration is calculated in the same way as in \cite{liu2023regformer}, where $\mathbf{q}^{0}$ and $\mathbf{t}^{0}$ are the rotation quaternion and the translation vector, respectively.

\textbf{Progressive Optimization:}
We progressively optimize the pose estimate $(\mathbf{q}^{0}, \mathbf{t}^{0})$ with features of different resolutions extracted by the encoder, as illustrated in Fig. \ref{fig:backbone} (c). The feature maps are divided into patches. Next the PWC structure \cite{wu2020pointpwc}  is used to estimate local transformations within each patch, and a cross-patch attention is obtained as well. Subsequently, the local patch registration and global cross-patch attentions are concatenated and processed through MLP and average pooling to obtain residual rotation $\Delta \mathbf{q}$ in quaternion and the translation vector $\Delta \mathbf{t}$  for refinement. These residuals are added to the estimated pose incrementally to yield improved outcomes:
\begin{equation}
\begin{split}
    &\mathbf{q}^l = \Delta \mathbf{q}^l  \mathbf{q}^{l+1},\\
   [0, \mathbf{t}^l] = \Delta \mathbf{q}^l &[0, \mathbf{t}^{l+1}] (\Delta \mathbf{q}^l)^{-1} +  [0, \Delta \mathbf{t}^l].
\end{split}
\end{equation}
where  \( \mathbf{q}^{l}\) and \( \hat{\mathbf{q}}^{l+1} \) represent the input and output rotation vector of the $l$-th block, \( \mathbf{t}^{l} \) and \( \hat{\mathbf{t}}^{l+1} \) represent the input and output translation vector of the $l$-th block.

\textbf{Loss Function:}
Due to the differences in magnitude and units between the quaternion and the translation vector, we decouple the translation and rotation components in the loss function and introduce two learnable parameters $s_q$ and $s_t$. For each layer, the training loss function is expressed as follows: 
\begin{equation}
    \mathcal{L} = || \frac{\hat{\mathbf{q}}}{||\hat{\mathbf{q}}||} - \mathbf{q} || \cdot exp(-s_q) + s_q + ||\hat{\mathbf{t}}-\mathbf{t}|| \cdot exp(-s_t) + s_t,
\end{equation}
where  \( \mathbf{q} \) and \( \hat{\mathbf{q}} \) represent the ground truth and estimated rotation vector,  \( \mathbf{t} \) and \( \hat{\mathbf{t}} \) are the ground truth and estimated translation vector.

Our network provides registration results at four different scales of resolution, and the total loss function is calculated using multi-level supervision:
\begin{equation}
    \mathcal{L}=\Sigma_{l=1}^4\lambda^l\mathcal{L}^l,
\end{equation}
where $\lambda^l$ denotes the weight parameter in the $l$-th scale of resolution.

\begin{figure}
    \centering
    \includegraphics[width=\linewidth]{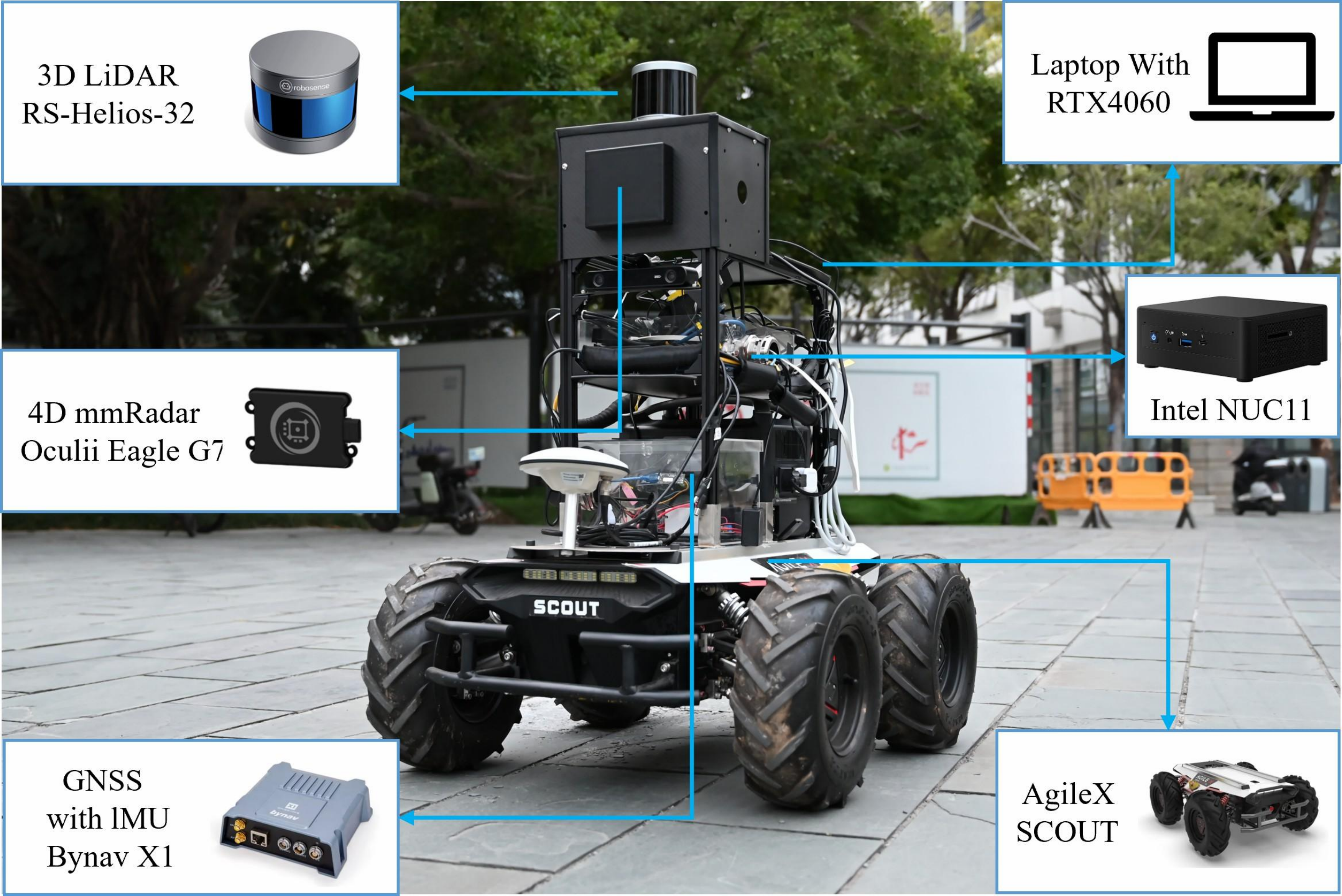}
    \caption{\textbf{The experimental platform.} }
    \label{fig:platform}
\end{figure}

\section{Experiment Analysis}\label{Experiment}
In this section, we comprehensively assess the performance of our proposed CMR network and LTR$^{2}$ system. We begin with the experiment setup, including experiment procedures, implementation platforms and evaluation metrics. Then, we analyze the localization accuracy of the CMR network against various state-of-the-art methods. Next, we evaluate navigation performance in extensive closed-loop real-world scenarios, considering accuracy, success rates and storage requirements. Finally, we assess system robustness when trajectory repeats are subject to static environment changes, smoke, or implemented on different platforms. 

\subsection{Experiment Setup}

\textbf{Experimental Procedure: }To ensure fair performance comparison across different baselines, all repeat paths start from the same initial pose in the teaching phase during the quantitative experiments. In Sec. \ref{reg_dataset}, we conduct open-loop experiments on the open-sourced VoD datasets \cite{apalffy2022vod} to assess the cross-modal registration performance. where the estimate error of the relative pose between each LiDAR–radar pair is assessed independently without executing the full navigation pipeline. In Sec. \ref{localizationperformance}, we conduct extensive closed-loop navigation experiments in the real world to assess the navigation performance, measuring the trajectory error between the teach path and the repeat path during real-world navigation. \ref {trajcompare} compares our iterative node selection approach with the traditional threshold-based ones used in T$\&$R. Sec.\ref{smoketest} explores the robustness of our navigation system in smoke-filled areas. Sect. \ref{timefinetunecompare} investigates the effects of our finetuning strategy against static environment changes. Finally Sect. \ref{crossplatform}  highlights the cross-platform generality of our method using different robot platforms during the teaching and repeating phases.
\begin{table}[t]
\centering
\caption{Parameters of the LTR$^2$ system}
\label{tab:param}
\resizebox{\linewidth}{!}{
\large
\begin{tabular}{@{}m{2.2cm} m{8.5cm} m{2.3cm}@{}}
\toprule
\textbf{Parameter} & \textbf{Description} & \textbf{Value} \\ 
\midrule
$\theta_{\text{mask}}$ & Azimuth Range for LiDAR Mask prior & \SI{135}{\degree} \\
$v_i$ & Expect velocity $\alpha$ weight & \SI{1.2}{\meter\per\second} \\ 
$w_{\text{rot}}$ & Rotation fitting weight & \SI{1.15}{\meter\per\radian} \\ 
$w_{\text{pos}}$ & Position fitting weight & \SI{1}{} \\  
$\alpha$ & Empirical atmospheric attenuation coefficient & \SI{0.03}{\per\kilo\meter} \\
$\delta$ & Threshold for switching nodes & \SI{0.12}{\meter} \\
$\tau$& Negative sample screening threshold & \SI{0.25}{\meter} \\
$\varepsilon_{\text{seg}}$ & Threshold of segmental trajectory & \SI{0.04}{\meter} \\
$\varepsilon$ & Threshold of global trajectory & \SI{0.16}{\meter} \\
$M$ & Minimum sample size for RANSAC & \SI{3}{} \\
$p$ & RANSAC confidence level & \SI{0.99}{} \\ 
$\eta$ & Expected inlier ratio for RANSAC & \SI{0.7}{} \\
$N_{\text{iter}}$ & Maximum RANSAC iterations & \SI{120}{} \\ 
$\gamma_{\text{doppler}}$ & Doppler residual inlier threshold & \SI{0.10}{\meter\per\second} \\
$a$ & Slope of linear classifier in (14) & \SI{0.3}{}\\ 
$b$ & Intercept of linear classifier in (14) & \SI{0.245}{}\\ 
$r_a$ &The ratio between positive and negative samples &\SI{3}{}\\
$\tau_{\mu}$ &The maximum drift &\SI{0.25}{m}\\
$L_{min}$ &The minimum total drift in trajectory &\SI{400}{m}\\

\bottomrule
\end{tabular}
}
\end{table}

\textbf{Platform:} Our LTR$^2$ navigation system is deployed on an AgileX SCOUT 2.0 robot platform (see Fig.~\ref{fig:platform}), which features differential four-wheel steering for agile mobility. It utilizes ROS-based master-slave communication, with an Intel NUC running ROS Melodic (Ubuntu~\SI{18.04}{}) as the master and a laptop equipped with an NVIDIA GeForce RTX 4060 GPU (CUDA~\SI{11.4}{}), an Intel Core i9-13900H processor, and \SI{16}{\giga\byte} of RAM, running ROS Noetic (Ubuntu~\SI{20.04}{}) as the slave.  The robot platform is equipped with a RoboSense RS-Helios 32-channel LiDAR operating at a sampling rate of \SI{10}{\hertz}, an OCULII-EAGLE 4D radar with a sampling rate of \SI{14}{\hertz} and a \ang{113} horizontal $\times$ \ang{45} vertical FOV, and a Bynav X1 GNSS.

We collected paired LiDAR-4D radar data using the robot platform on the university campus, which is used for training the CMR network during the pre-training phase. During teaching, the robot platform is manually controlled to travel a path with both LiDAR and 4D radar collected along the path. During the repeating phase, real-time data is collected from the 4D radar only. 

\textbf{Implementation Details:} The projected pseudo images are generated with a resolution of \ang{64} $\times$ \ang{768} for VoD dataset and \ang{32} $\times$ \ang{675} for our self-collected dataset, matching the LiDAR angular resolutions. The window size and shift size for feature extraction are set to 4 and 2, respectively. All experiments are performed using PyTorch 2.0.1 on four NVIDIA A6000 GPUs, optimized using the Adam algorithm with momentum parameters $\beta_1 = 0.9$ and $\beta_2 = 0.999$. The initial learning rate is set to \SI{0.002}{} and exponentially decays every \SI{350000}{} steps down to a minimum of \SI{0.00001}{}. The batch size is fixed to \SI{16}{}. In the loss function, the layer-wise weighting coefficients $\alpha^l$ are set to \SI{1.6}{}, \SI{0.8}{}, \SI{0.4}{}, and \SI{0.2}{} for the four hierarchical stages. The learnable scaling parameters $k_t$ and $k_r$ are initialized as \SI{0.0}{} and \SI{-2.5}{}, respectively. In the fine-tuning stage, we train 30 epochs on the constructed fine-tuning dataset with the learning rate of \SI{0.0001}{}. Throughout all experiments, all parameters of the LTR$^2$ system except the network weights are kept the same, as detailed in Tab.~\ref{tab:param}.  All experiment results are averaged over three random runs to ensure statistical robustness.


\begin{table*}[t]
\centering
\caption{RTE of Open-loop Performance Comparison in VoD datasets [RTE (\si{\meter})$\downarrow$, RTE STD (\si{\meter})$\downarrow$]}
\label{table:RTE_vod}
\renewcommand{\arraystretch}{1.3}
\resizebox{\textwidth}{!}{
\begin{tabular}{lcccccccccccc}
\toprule
\multirow{2}{*}{\textbf{Methods}} & \multicolumn{2}{c}{\textbf{01}} & \multicolumn{2}{c}{\textbf{02}} & \multicolumn{2}{c}{\textbf{03}} & \multicolumn{2}{c}{\textbf{04}} & \multicolumn{2}{c}{\textbf{14}} & \multicolumn{2}{c}{\textbf{19}} \\
\cmidrule(r){2-3} \cmidrule(r){4-5} \cmidrule(r){6-7} \cmidrule(r){8-9} \cmidrule(r){10-11} \cmidrule(r){12-13}
& \textbf{RTE} & \textbf{RTE STD} & \textbf{RTE} & \textbf{RTE STD} & \textbf{RTE} & \textbf{RTE STD} & \textbf{RTE} & \textbf{RTE STD} & \textbf{RTE} & \textbf{RTE STD} & \textbf{RTE} & \textbf{RTE STD} \\
\midrule
\textbf{GICP}                
&0.470 &0.606
&0.411 &0.365
&0.388 &0.366
&0.428 &0.379
&0.293 &0.242
&0.591 &0.896 \\

\textbf{ICP}     
&0.428 &0.254
&0.451 &0.264
&0.517 &0.260
&0.488 &0.262
&0.520 &0.193
&0.670 &0.330 \\

\textbf{NDT}                
&0.306 &0.102
&0.304 &\underline{0.136}
&0.371 &0.127
&0.333 &0.135
&0.325 &0.108
&0.376 &\underline{0.133} \\

\textbf{VGICP}      
&3.947 &3.605
&4.501 &4.615
&3.859 &3.407
&4.516 &5.255
&4.008 &3.312
&7.741 &11.248\\

\textbf{ADPGICP\cite{zhang20234dradarslam}}                
&6.476 &9.961
&20.681 &12.688
&9.316 &10.132
&9.551 &10.590
&11.839 &11.248
&17.504 &13.517\\ 

\midrule
\textbf{CMflow\cite{cmflow}}                      
& 7.307 & 4.508
& 3.777 & 2.804
& 10.195 & 5.261
& 5.707 & 3.971
& 8.876 & 5.664
& 4.810 & 4.088 \\

\textbf{Zhang \etal\cite{zhang2024adaptive}}                   
& 3.472 & 2.438
& 11.461 & 7.537
& 6.036 & 3.195
& 7.232 & 4.531
& 3.567 & 2.696
& 9.247 & 7.163 \\

\textbf{RegFormer\cite{liu2023regformer}}             
& 0.819 & 0.614
& 0.431 & 0.509
& 0.398 & 0.732
& 0.554 & 0.696
& 0.619 & 0.316
& 0.817 & 1.024 \\

\textbf{HRegNet\cite{hregnet}}             
& 10.321 & 6.057
& 6.394 & 3.658
& 4.225 & 2.910
& 5.117 & 3.078
& 10.396 & 6.300
& 6.267 & 3.808 \\

\midrule
\textbf{Ours (w/o Intensity)}                      
& 0.408 & 0.270
& 0.400 & 0.294
& 0.409 & 0.286
& 0.387 & 0.268
& 0.598 & 0.277
& 0.397 & 0.290 \\

\textbf{Ours}                    
& \textbf{0.092} & \textbf{0.052}
& \textbf{0.106} & \textbf{0.115}
& \textbf{0.100} & \textbf{0.070}
& \textbf{0.066} & \textbf{0.089}
& \textbf{0.116} & \textbf{0.041}
& \textbf{0.121} & \textbf{0.096} \\

\textbf{Ours ($\mathcal{N}(0, \sigma^2_{5^\circ})$)}             
& \underline{0.167} & \underline{0.066}
& \underline{0.152} & 0.160
& \underline{0.146} & \underline{0.085}
& \underline{0.112} & \underline{0.104}
& \underline{0.160} & \underline{0.056}
& \underline{0.186} & 0.141 \\

\textbf{Ours ($\mathcal{N}(0, \sigma^2_{15^\circ})$)}
& 0.172 & 0.079
& 0.162 & 0.164
& 0.153 & 0.110
& 0.142 & 0.107
& 0.180 & 0.068
& 0.202 & 0.176 \\

\textbf{Ours ($\mathcal{N}(0, \sigma^2_{30^\circ})$)}
& 3.819 & 3.962
& 4.601 & 4.477
& 3.919 & 3.321
& 4.243 & 5.339
& 4.083 & 3.291
& 5.954 & 4.608 \\
\bottomrule
\end{tabular}
}

\vspace{0.5em}
{\raggedright\small Best performance in \textbf{bold}, second best \underline{underlined}.  The bottom three rows present our CMR network under yaw perturbations by adding Gaussian
noises of varying magnitudes. \par}
\end{table*}

\begin{table*}[t]
\centering
\caption{RYE of Open-loop Performance Comparison in VoD datasets [RYE (\si{\degree})$\downarrow$, RYE STD (\si{\degree})$\downarrow$]}
\label{table:yae_vod}
\renewcommand{\arraystretch}{1.3}
\resizebox{\textwidth}{!}{
\begin{tabular}{lcccccccccccc}
\toprule
\multirow{2}{*}{\textbf{Methods}} & \multicolumn{2}{c}{\textbf{01}} & \multicolumn{2}{c}{\textbf{02}} & \multicolumn{2}{c}{\textbf{03}} & \multicolumn{2}{c}{\textbf{04}} & \multicolumn{2}{c}{\textbf{14}} & \multicolumn{2}{c}{\textbf{19}} \\
\cmidrule(r){2-3} \cmidrule(r){4-5} \cmidrule(r){6-7} \cmidrule(r){8-9} \cmidrule(r){10-11} \cmidrule(r){12-13}
& \textbf{RYE} & \textbf{RYE STD} & \textbf{RYE} & \textbf{RYE STD} & \textbf{RYE} & \textbf{RYE STD} & \textbf{RYE} & \textbf{RYE STD} & \textbf{RYE} & \textbf{RYE STD} & \textbf{RYE} & \textbf{RYE STD} \\
\midrule
\textbf{GICP} 
& 1.978 & 2.810 & 1.503 & 1.269 & 1.807 & 1.947 & 1.581 & 1.010 & 1.324 & 0.918 & 2.010 & 2.438 \\

\textbf{ICP} 
& 1.837 & 2.012 & 1.633 & 1.555 & 2.204 & 1.987 & 2.230 & 1.541 & 1.990 & 1.238 & 2.591 & 1.846 \\

\textbf{NDT} 
& 0.398 & 0.402 & 0.454 & 0.391 & 0.436 & 0.380 & 0.459 & 0.366 & 0.376 & 0.242 & 0.812 & 0.443 \\

\textbf{VGICP} 
& 3.250 & 4.478 & 3.179 & 3.567 & 4.186 & 8.221 & 3.718 & 4.071 & 3.164 & 3.463 & 7.627 & 11.586 \\

\textbf{ADPGICP\cite{zhang20234dradarslam}}
& 3.523 & 8.636 & 7.293 & 15.165 & 5.175 & 13.212 & 2.529 & 3.897 & 1.260 & 1.830 & 4.978 & 6.061 \\

\midrule
\textbf{CMflow\cite{cmflow}} 
& 1.057 & 0.614 & 0.774 & 0.577 & 1.569 & 0.938 & 0.956 & 0.611 & 1.477 & 0.824 & 0.874 & 0.556 \\

\textbf{Zhang \etal\cite{zhang2024adaptive}} 
& 0.635 & 0.415 & 1.468 & 0.964 & 1.064 & 0.618 & 1.204 & 0.690 & 0.719 & 0.457 & 1.550 & 0.947 \\

\textbf{RegFormer\cite{liu2023regformer}} 
& 1.224 & 0.737 & 0.805 & 0.519 & 0.628 & 0.363 & 1.627 & 1.024 & 1.429 & 0.848 & 0.731 & 0.422 \\

\textbf{HRegNet\cite{hregnet}} 
& 1.540 & 0.938 & 1.192 & 0.717 & 0.753 & 0.487 & 0.904 & 0.584 & 1.596 & 0.956 & 1.128 & 0.679 \\

\midrule
\textbf{Ours (w/o Intensity)} 
& 0.726 & 0.491 & 0.660 & 0.497 & 0.708 & 0.503 & 0.728 & 0.509 & 0.768 & 0.540 & 0.798 & 0.592 \\

\textbf{Ours} 
& \textbf{0.219} & \textbf{0.163} & \textbf{0.221} & \textbf{0.159} & \textbf{0.217} & \textbf{0.162} & \underline{0.232} & \underline{0.180} & \textbf{0.160} & \textbf{0.102} & \textbf{0.191} & \textbf{0.130} \\

\textbf{Ours ($\mathcal{N}(0, \sigma^2_{5^\circ})$)} 
& \underline{0.262} & \underline{0.196} & \underline{0.254} & \underline{0.184} & \underline{0.236} & \underline{0.184} & \textbf{0.213} & \textbf{0.159} & \underline{0.196} & \underline{0.136} & \underline{0.218} & \underline{0.148} \\

\textbf{Ours ($\mathcal{N}(0, \sigma^2_{15^\circ})$)} 
& 0.279 & 0.207 & 0.275 & 0.194 & 0.251 & 0.198 & 0.262 & 0.208 & 0.217 & 0.146 & 0.243 & 0.158 \\

\textbf{Ours ($\mathcal{N}(0, \sigma^2_{30^\circ})$)} 
& 1.477 & 0.747 & 1.726 & 0.738 & 1.850 & 1.194 & 2.287 & 0.901 & 2.684 & 0.440 & 1.471 & 0.429 \\

\bottomrule
\end{tabular}
}

\vspace{0.5em}
{\raggedright\small Best performance in \textbf{bold}, second best \underline{underlined}. The bottom three rows present our CMR network under yaw perturbations by adding Gaussian
noises of varying magnitudes. \par}
\end{table*}

\textbf{Evaluation Metrics: }
We use the Relative Translation Error (RTE), the Relative Yaw Error (RYE), and their standard deviations (STDs) to evaluate single-frame registration performance in open-loop experiments.  For closed-loop navigation, we employ the Absolute Translation Error (ATE) and the Yaw Angle Error (YAE), together with their STDs, to quantify global localization accuracy and assess stability across repeated trials. Both the teach and repeat trajectories for evaluation are estimated using LiDAR odometry LIO-SAM with GNSS factor incorporated \cite{shan2020lio}. The odometries are recorded in the same global coordinate with the aid of GNSS, and then aligned in a time–independent manner for computing the ATE.

\begin{figure*}
    \centering
    \includegraphics[width=\textwidth]{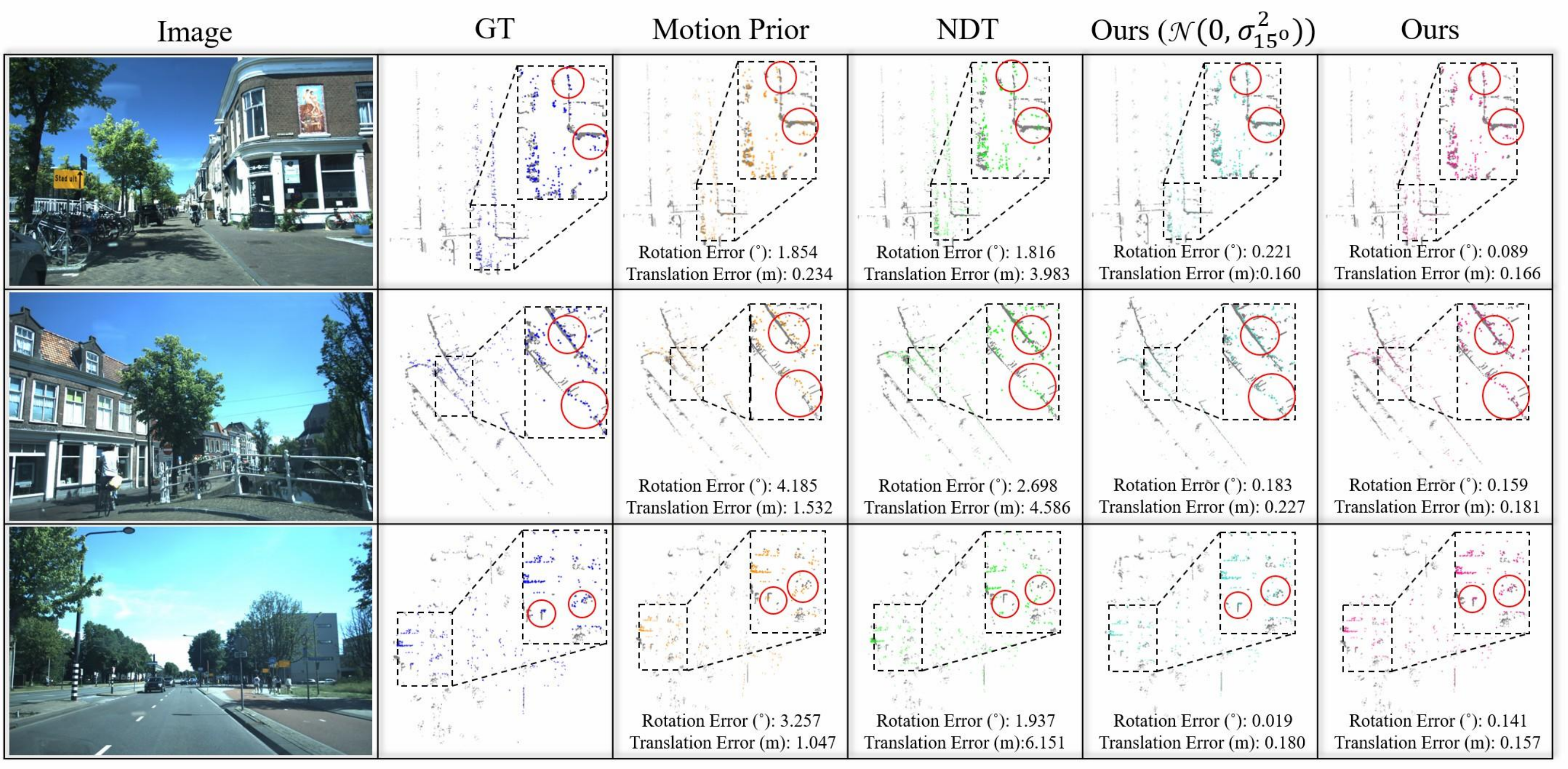}
    \caption{\textbf{The point cloud overlay via registration in the VoD dataset. }Grey and color points are  LiDAR and radar point cloud, respectively. Zoomed color overlays highlight CMR network's superior alignment accuracy over baseline methods and robustness to inaccurate pose priors.}
    \label{fig:vod_visualized}
\end{figure*}

\subsection{Open-loop Experiments}\label{reg_dataset}

\textbf{Dataset: }The VoD dataset comprises 8600 synchronized frames from 64-channel LiDAR, camera, and 4D radar sensors collected in complex urban traffic environments. The ground truth of relative poses between consecutive frames is provided, which was obtained from fused RTK GPS, IMU, and wheel odometry. Sequences 01, 02, 03, 04, 14, and 19 are used for testing, while the other sequences are for training.

\textbf{Benchmarking Methods and Evaluation Protocol: }We evaluate our approach against both classical and learning-based registration methods, which include GICP, NDT, ADPGICP \cite{zhang20234dradarslam}, VGICP, RegFormer \cite{liu2023regformer}, a visual-aided radar odometry \cite{zhang2024adaptive} and CMflow\cite{cmflow}.  To ensure a fair comparison, all methods use the same pose prior for matching. The resulting transformation estimates are evaluated against the ground truth for performance assessment. 

\textbf{Performance Comparison and Analysis: }As shown in Tab.~\ref{table:RTE_vod} and \ref{table:yae_vod}, our CMR network outperforms all baseline methods in RTE and RYE, thanks to its dual-branch architecture that integrates geometric and intensity cues. Other network-based methods have poor cross-modal performance due to training only for LiDAR geometry, and methods based on geometric feature optimization also have difficulty aligning heterogeneous data due to the multi-path noise of 4D radar, as shown in Fig.~\ref{fig:vod_visualized}. Our CMR can alleviate structural ambiguity and improve robustness by fusing complementary geometric and intensity information.

Furthermore, we conduct a detailed analysis of the environmental complexity of different test sequences. In particular, in sequences 04 and 14, the proportion of dynamic obstacles in these two sequences is significantly higher than that in other test sequences, and the frequent appearance of moving objects makes the pure geometric method unstable. For example, in sequence 04, our method reduces RTE by about \SI{80.2}{\%} compared to the second-best NDT method. Similarly, in sequence 14, our method reduces RTE from \SI{0.293}{m} of GICP to \SI{0.116}{m}, with an error reduction of about \SI{60.4}{\%}. This performance gap clearly illustrates the significant advantage of our proposed method of fusion of intensity and geometric information in highly dynamic and structurally variable environments.

We further assess the impact of intensity features by disabling the corresponding branch. Without intensity, RTE and RYE increase significantly (e.g., RTE rises from \SI{0.0918}{m} to \SI{0.4080}{m} in Seq. 01, with an error increase of about \SI{344}{\%}), confirming the importance of power density-intensity alignment between 4D radar and LiDAR for accurate cross-modal registration.

\textbf{Robustness Evaluation Under Pose Perturbations: }Additionally, we evaluate the robustness of our proposed CMR network under perturbed pose priors. We simulate yaw perturbations by adding Gaussian noises of varying magnitudes to the pose prior. Tab. \ref{table:RTE_vod} and Tab. \ref{table:yae_vod} show that RTE remains below \SI{0.15}{m} for perturbations up to \ang{15}, significantly better than baseline methods (e.g., GICP and ICP exceed \SI{0.4}{m}). Only at perturbations beyond \ang{20} does error notably increase, highlighting the boundary conditions of robustness. With large perturbations of \ang{30}, a large estimate error emerges, but our dual-branch architecture still outperforms other comparative methods, highlighting its resilience against inaccuracies in the match prior.

\begin{table}[t]
\centering
\caption{Difficulty Metrics for the Teach Paths}
\label{tab:teach_path}
\resizebox{\linewidth}{!}{
\large
\begin{tabular}{lrrrrr}
\toprule
\textbf{Tag} & \textbf{Range} & \textbf{Elevation} & \textbf{Pitch} & \textbf{Min radius} & \textbf{Challenges} \\
\midrule
A & 1325.8m & 9.6m & 8$^\circ$  & 1.32m & Long range  \\
B & 516.7m & 7.6m & 6$^\circ$  & 1.24m & Parked Cars \\
C & 293.4m & 1.0m & 3$^\circ$  & 1.16m & Pedestrains \\
D & 353.7m & 3.0m & 5$^\circ$  & 1.38m & Riverbank \\
E & 439.2m & 0.4m & 1$^\circ$  & 0.89m & Narrow pathway \\
F & 643.3m & 4.2m & 4$^\circ$  & 1.38m & Long corridor \\
G & 711.8m & 0.9m & 30$^\circ$  & 1.21m & Large  pitch oscillation \\
Finetune1 & 246.6m & 0.2m & 3$^\circ$ & 1.37m & Scene change \\
Finetune2 & 380.7m & 0.4m & 5$^\circ$ & 1.54m & Scene change \\
Smoke & 105.1m & 0.1m & 2$^\circ$ & 1.01m & Smoke \\
Large-scale & 4627m & 26.25m & 16.4$^\circ$ & 1.01m & Largescale \\

\bottomrule
\end{tabular}
}
\end{table}

\begin{figure}[t]
    \centering
    \includegraphics[width=\linewidth]{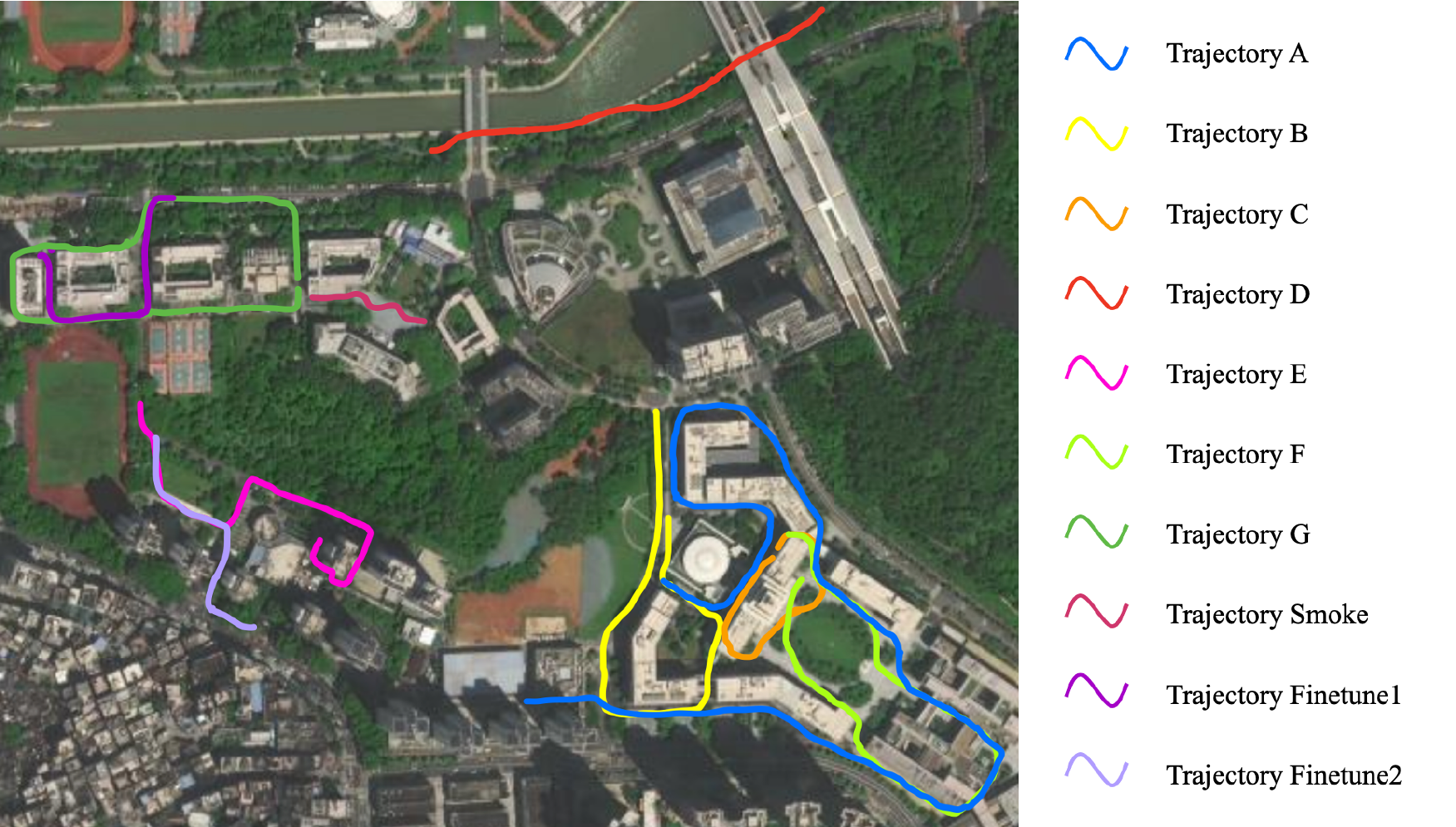}
    \caption{\textbf{Display of the real-world experimental trajectory on satellite map, }which illustrates multiple teach-and-repeat trajectories across diverse environments within a university campus.}
    \label{fig:satellite}
\end{figure}

\subsection{Closed-Loop Navigation in Real-World Environments}\label{localizationperformance}
We evaluate the closed-loop navigation performance of the proposed LTR$^2$ system across seven experimental trajectories (A to G) deployed in a university campus, as shown in Fig.~\ref{fig:satellite}. Each trajectory presents distinct environmental challenges, summarized in Tab.~\ref{tab:teach_path}. It is worth pointing out that the LiDAR odometry is employed only for point cloud visualization and trajectory ground truth.  Both LiDAR and radar data collected from trajectory A to F are used to train the CMR network in the pre-training phase. Once trained, the network weights are kept fixed unless updated during fine-tuning.

Trajectory A spans over \SI{1.327}{
km} and covers a variety of campus scenes and serves as the primary benchmark for comparing different 
navigation solutions. We evaluated four categories of methods:
\begin{itemize}
    \item \textbf{SLAM-based navigation:} The tightly coupled LIO-SAM \cite{shan2020lio} is used to construct a point cloud map and localization during the teaching phase, followed by constant localization on the map and trajectory tracking during the repeating phase; 
    \item \textbf{LiDAR T$\&$R navigation:} Both GICP and learning-based Regformer \cite{liu2023regformer} methods are implemented for point cloud registration; 
    \item \textbf{Radar T$\&$R navigation:} Three different point cloud registration methods including GICP, ADPGICP, and NDT are implemented for 4D radar-only registration; 
    \item \textbf{Cross-modal T$\&$R navigation:} A cross-modal registration baseline using GICP, ADPGICP\cite{zhang20234dradarslam}, and our proposed deep learning-based CMR method has been established. Since the only open-source system VT$\&$R3\cite{tim2010} is developed based on the ROS2 environment, but only a ROS1-based driver is provided for our used 4D radar,  we select CT-ICP \cite{burnett_ral22} from the registration module of VT$\&$R3 and replace it with our localization block for comparative experiments.
\end{itemize}

Furthermore, our CMR network without the radar power LiDAR intensity alignment branch is also implemented for ablation study of the closed-loop system. The methods that fail on trajectory A are excluded from subsequent evaluations on trajectories B-F.

\begin{table*}[ht]
\centering
\caption{Navigation Performance Comparison in Trajectory A [ATE (\si{\meter})$\downarrow$, YAE (\si{\degree})$\downarrow$, ATE STD (m)$\downarrow$, YAE STD (\si{\degree})$\downarrow$]}
\label{localization1}
\resizebox{\textwidth}{!}{%
\scriptsize
\begin{tabular}{@{}l|c|cc|cccccc@{}}
\toprule
\textbf{Category} & \textbf{Methods} & \textbf{Success Rate} & \textbf{Max Run Distance} & \textbf{ATE} & \textbf{YAE} & \textbf{ATE STD} & \textbf{YAE STD} & \textbf{Registration Frequency} \\
\midrule
\multirow{1}{*}{\textbf{SLAM-based}} 
& \textbf{LIO-SAM} & 3 / 3 & 1325m / 1325m & \textbf{0.071} & \textbf{1.180} & 0.056 & \underline{1.305} & 23.16Hz \\
\midrule
\multirow{2}{*}{\textbf{LiDAR T$\&$R}} 
& \textbf{GICP} & 3 / 3 & 1325m / 1325m & 0.083 & 2.158 & \underline{0.052} & 2.340 & 20.15Hz \\
& \textbf{Regformer} & 3 / 3 & 1325m / 1325m & 0.116 & 2.835 & 0.086 & 2.244 & 18.83Hz  \\
\midrule
\multirow{3}{*}{\textbf{Radar T$\&$R}} 
& \textbf{GICP} & 0 / 3 & 380.4m / 1325m & / & / & / & / & / \\
& \textbf{ADPGICP} & 0 / 3 & 349.5m / 1325m & / & / & / & / & / \\
& \textbf{NDT} & 0 / 3 & 378.6m / 1325m & / & / & / & / & / \\
\midrule
\multirow{5}{*}{\textbf{Cross-modal T$\&$R}} 
& \textbf{CT-ICP} & 0 / 3 & 93.2m / 1325m & / & / & / & / & / \\
& \textbf{GICP} & 0 / 3 & 44.8m / 1325m & / & / & / & / & / \\
& \textbf{ADPGICP} & 0 / 3 & 51.6m / 1325m & / & / & / & / & / \\
& \textbf{Ours (w/o Intensity)} & 3 / 3 & 1325m / 1325m & 0.126 & 2.922 & 0.084 & 1.401 & 17.92Hz \\
& \textbf{Ours} & 3 / 3  & 1325m / 1325m & \underline{0.073} & \underline{1.481} & \textbf{0.051} & \textbf{1.147} & 16.11Hz \\
\bottomrule
\end{tabular}
}

\vspace{0.5em}
{\raggedright\small Best performance in \textbf{bold}, second best \underline{underlined}.\par}
\end{table*}

\begin{table*}[t]
\centering
\caption{ATE of Closed-loop Performance Comparison Across Trajectories B-G [ATE (\si{\meter})$\downarrow$, ATE STD (\si{\meter})$\downarrow$]}
\label{table:ate_localization}
\resizebox{\textwidth}{!}{
\begin{tabular}{lcccccccccccc}
\toprule
\multirow{2}{*}{\textbf{Methods}} & \multicolumn{2}{c}{\textbf{B}} & \multicolumn{2}{c}{\textbf{C}} & \multicolumn{2}{c}{\textbf{D}} & \multicolumn{2}{c}{\textbf{E}} & \multicolumn{2}{c}{\textbf{F}} & \multicolumn{2}{c}{\textbf{G}} \\
\cmidrule(r){2-3} \cmidrule(r){4-5} \cmidrule(r){6-7} \cmidrule(r){8-9} \cmidrule(r){10-11} \cmidrule(r){12-13}
& \textbf{ATE} & \textbf{ATE STD} & \textbf{ATE} & \textbf{ATE STD} & \textbf{ATE} & \textbf{ATE STD} & \textbf{ATE} & \textbf{ATE STD} & \textbf{ATE} & \textbf{ATE STD} & \textbf{ATE} & \textbf{ATE STD} \\
\midrule
\textbf{LiDAR-LiDAR(GICP)}                
& 0.073 & 0.049
& 0.089 & 0.060
& 0.068 & 0.038
& 0.083 & \underline{0.043}
& 0.085 & 0.068
& 0.080 & 0.067 \\

\textbf{LiDAR-LiDAR(Regformer)}     
& 0.113 & 0.066
& 0.107 & 0.098
& 0.073 & 0.037
& \underline{0.080} & 0.058
& 0.104 & 0.079
& \underline{0.074} & 0.069 \\

\textbf{LiDAR SLAM}                
& \textbf{0.060} & \textbf{0.040}
& \textbf{0.068} & \textbf{0.054}
& \textbf{0.057} & \textbf{0.031}
& / & /
& \textbf{0.065} & \textbf{0.034}
& \textbf{0.063} & \textbf{0.045} \\

\textbf{Ours (w/o Intensity)}      
& 0.078 & 0.052
& 0.106 & 0.102
& 0.082 & 0.043
& / & /
& 0.105 & 0.090
& / & / \\

\textbf{Ours}                      
& \underline{0.071} & \underline{0.044}
& \underline{0.080} & \underline{0.058}
& \underline{0.067} & \underline{0.033}
& \textbf{0.067} & \textbf{0.042}
& \underline{0.069} & \underline{0.044}
& 0.077 & \underline{0.066} \\

\bottomrule
\end{tabular}
}

\vspace{0.5em}
{\raggedright\small Best performance in \textbf{bold}, second best \underline{underlined}.\par}
\end{table*}

\begin{table*}[t]
\centering
\caption{YAE of Closed-Loop Performance Comparison Across Trajectories B-G [YAE (\si{\degree})$\downarrow$, YAE STD (\si{\degree})$\downarrow$]}
\label{table:yae_localization}
\resizebox{\textwidth}{!}{
\begin{tabular}{lcccccccccccc}
\toprule
\multirow{2}{*}{\textbf{Methods}} & \multicolumn{2}{c}{\textbf{B}} & \multicolumn{2}{c}{\textbf{C}} & \multicolumn{2}{c}{\textbf{D}} & \multicolumn{2}{c}{\textbf{E}} & \multicolumn{2}{c}{\textbf{F}} & \multicolumn{2}{c}{\textbf{G}} \\
\cmidrule(r){2-3} \cmidrule(r){4-5} \cmidrule(r){6-7} \cmidrule(r){8-9} \cmidrule(r){10-11} \cmidrule(r){12-13}
& \textbf{YAE} & \textbf{YAE STD} & \textbf{YAE} & \textbf{YAE STD} & \textbf{YAE} & \textbf{YAE STD} & \textbf{YAE} & \textbf{YAE STD} & \textbf{YAE} & \textbf{YAE STD} & \textbf{YAE} & \textbf{YAE STD} \\
\midrule
\textbf{LiDAR-LiDAR(GICP)}                
& \underline{2.262} & 2.976
& 3.503 & 4.150
& 2.600 & 1.844
& \underline{2.882} & \underline{2.200}
& 3.213 & 3.089
& \underline{2.792} & \underline{2.658} \\

\textbf{LiDAR-LiDAR(Regformer)}     
& 3.506 & 3.404
& 4.747 & 4.530
& 2.454 & 1.897
& 3.021 & 2.790
& 4.665 & 4.427
& 2.825 & 3.177 \\

\textbf{LiDAR SLAM}                
& \textbf{1.971} & \textbf{2.008}
& \textbf{2.166} & \textbf{2.476}
& \textbf{1.695} & \textbf{1.152}
& / & /
& \textbf{1.279} & \textbf{1.645}
& \textbf{2.055} & \textbf{1.874} \\

\textbf{Ours (w/o Intensity)}      
& 3.456 & 3.177
& 4.433 & 4.414
& 3.515 & 2.430
& / & /
& 3.537 & 3.234
& / & / \\

\textbf{Ours}                      
& 2.657 & \underline{2.544}
& \underline{2.468} & \underline{3.146}
& \underline{2.326} & \underline{1.547}
& \textbf{2.151} & \textbf{1.693}
& \underline{1.982} & \underline{2.030}
& 2.949 & 3.139 \\

\bottomrule
\end{tabular}
}

\vspace{0.5em}
{\raggedright\small Best performance in \textbf{bold}, second best \underline{underlined}.\par}
\end{table*}

\begin{figure}[t]
    \centering
    \includegraphics[width=\linewidth]{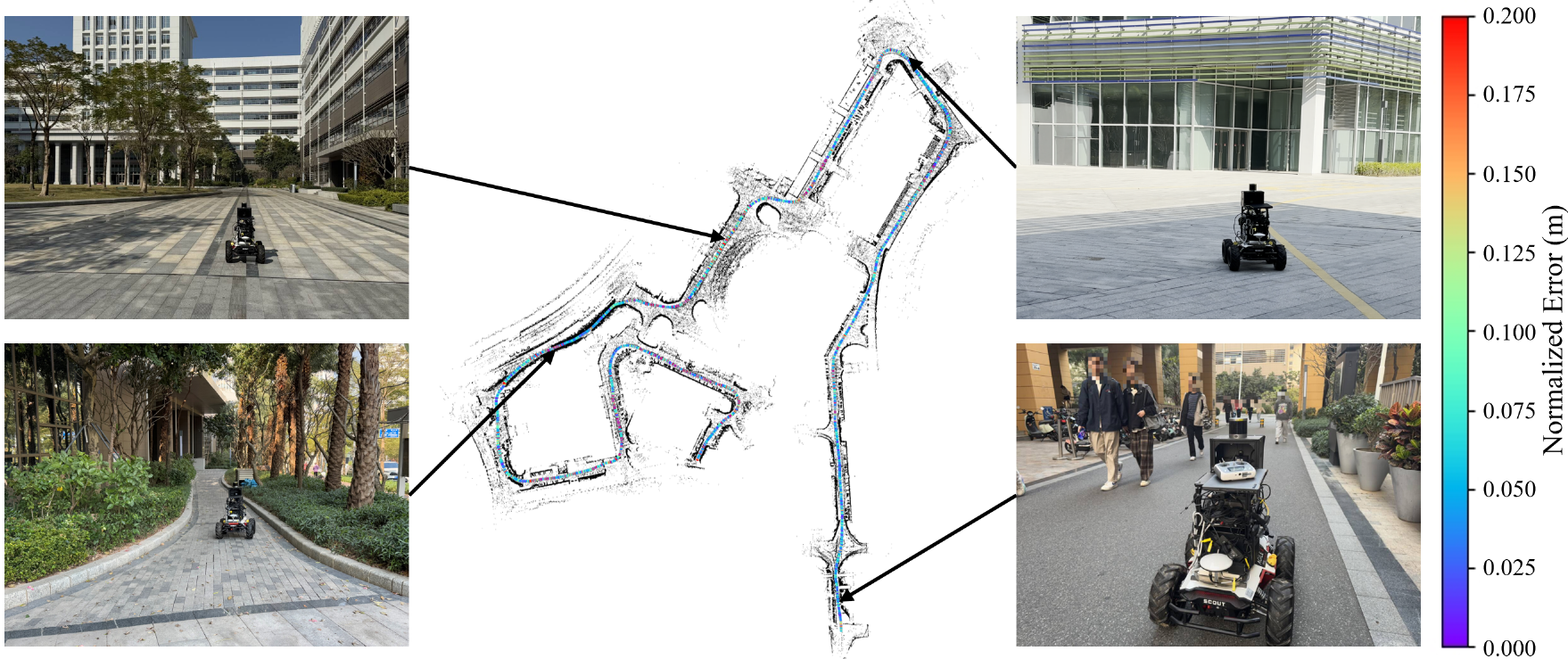}
    \caption{\textbf{Illustration of the repeat error in trajectory A.} The color-coded path shows centimeter-level positioning errors between the teach-and-repeat trajectories.}
    \label{fig:traja}
\end{figure}

\textbf{Performance Comparison:}
Unlike open-loop experiments that evaluates only the cross-modal registration methods, we evaluate three complete T$\&$R pipelines with different modalities, that is LiDAR T$\&$R,  Radar T$\&$R and cross-modal T$\&$R, together with the SLAM-based navigation baseline in terms of navigation success and localization accuracy.

\begin{figure*}[htbp]
    \centering
    \subfloat[\textcolor{green}{Ours}, \textcolor{arroworange}{ADPGICP} in Slope\label{fig:traj_slope}]{
        \includegraphics[width=0.25\textwidth]{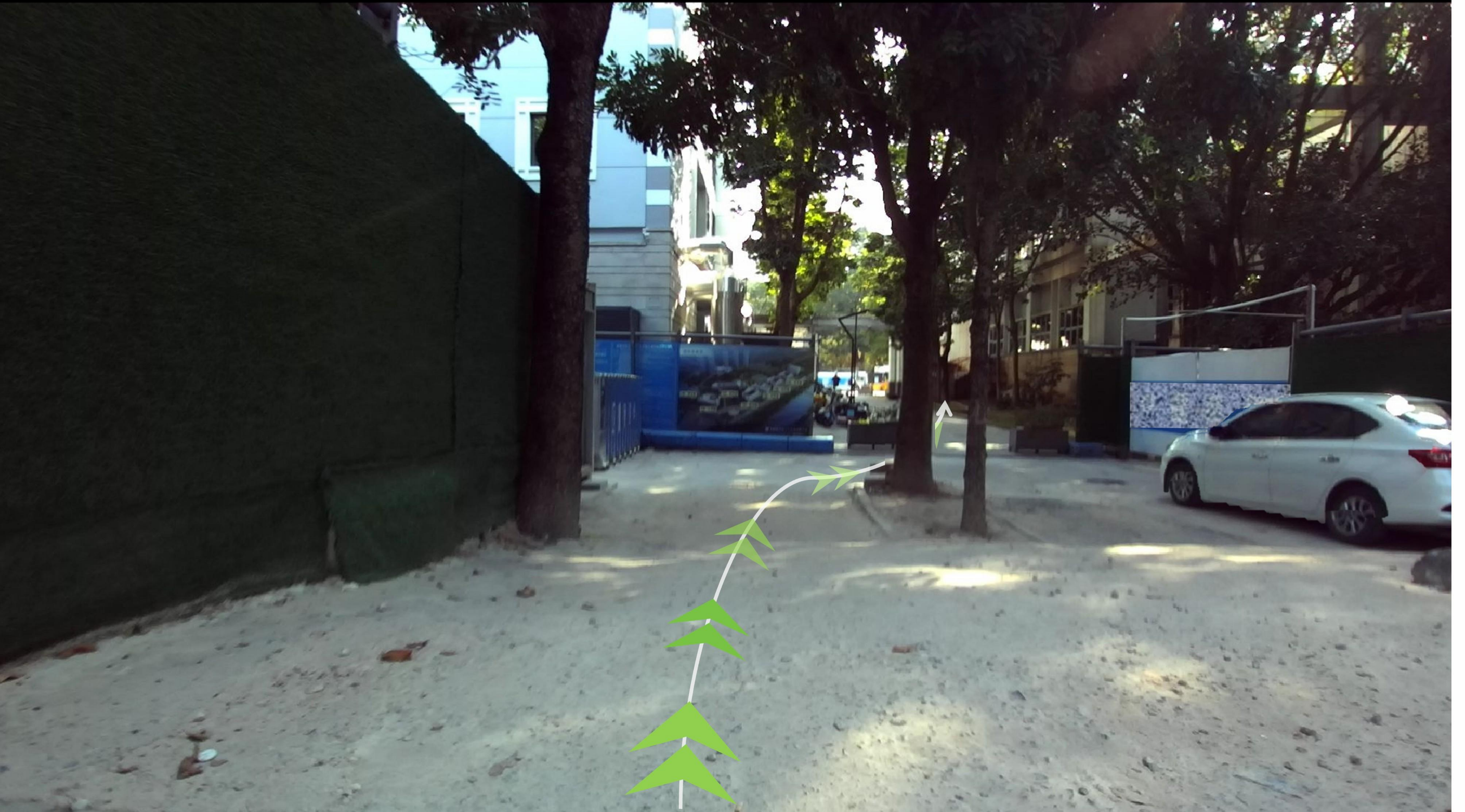}
        \includegraphics[width=0.25\textwidth]{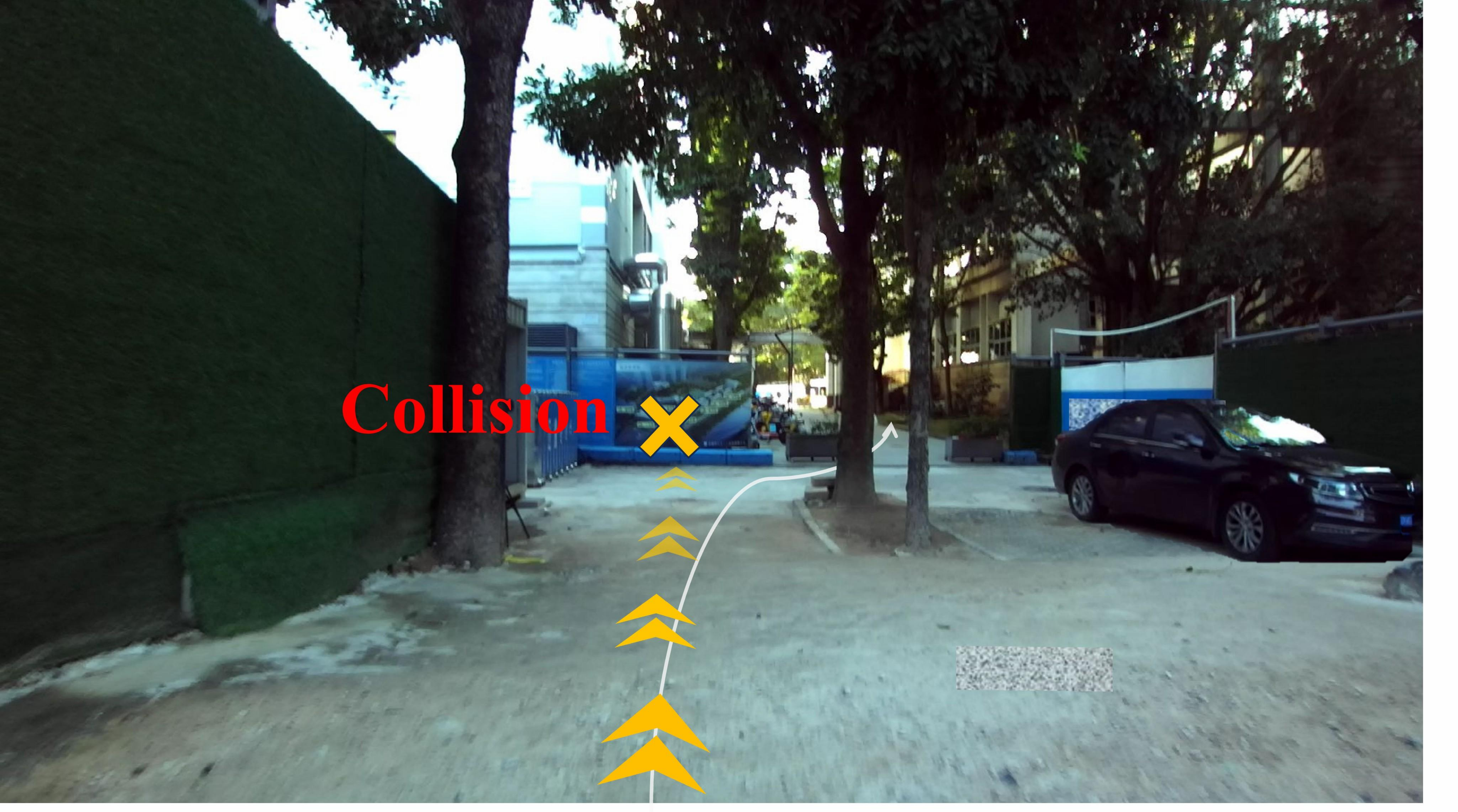}
    }
    \subfloat[\textcolor{green}{Ours}, \textcolor{red}{LIO-SAM} in Narrow Area\label{fig:traj_narrow}]{
        \includegraphics[width=0.25\textwidth]{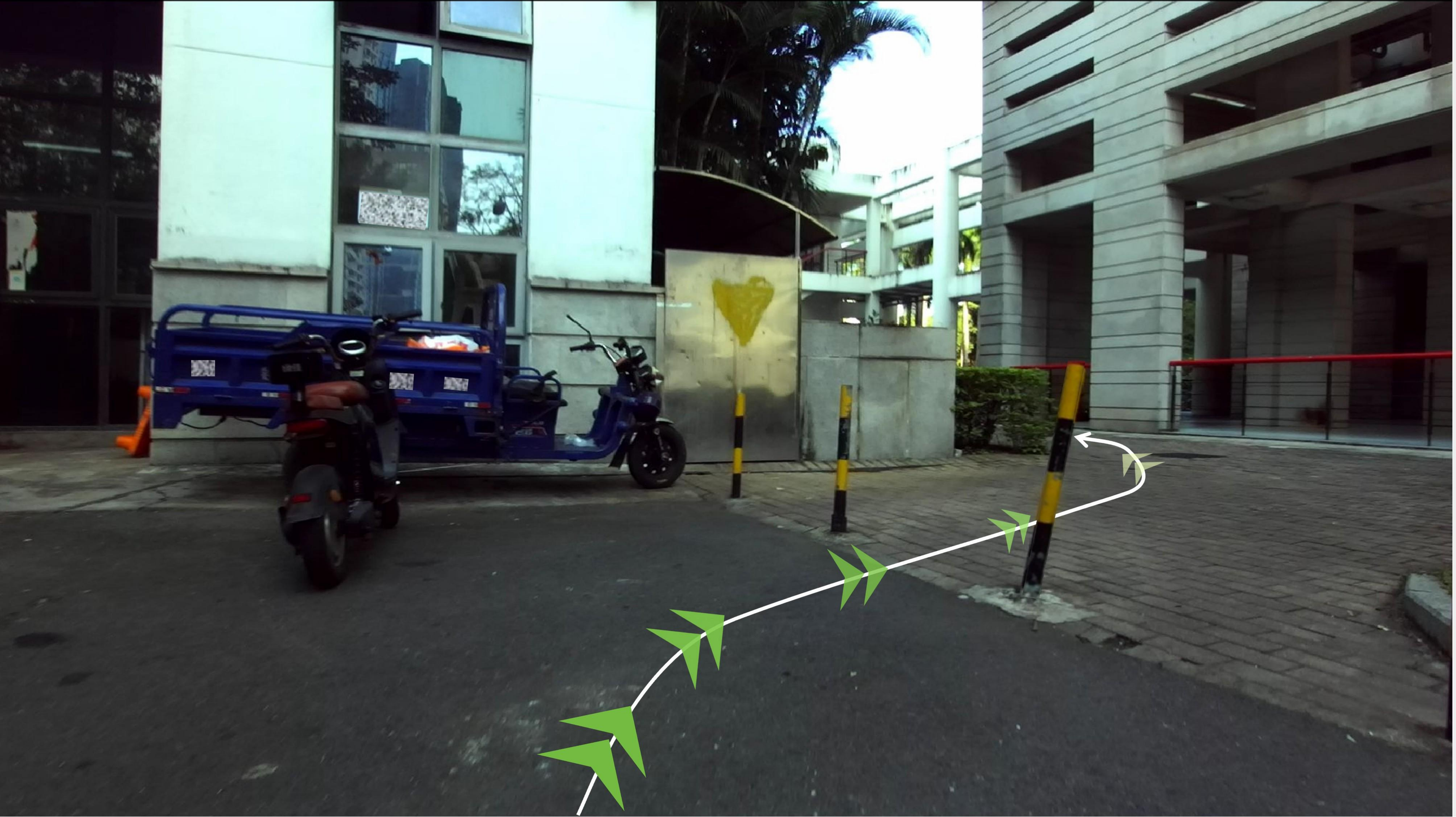}
        \includegraphics[width=0.25\textwidth]{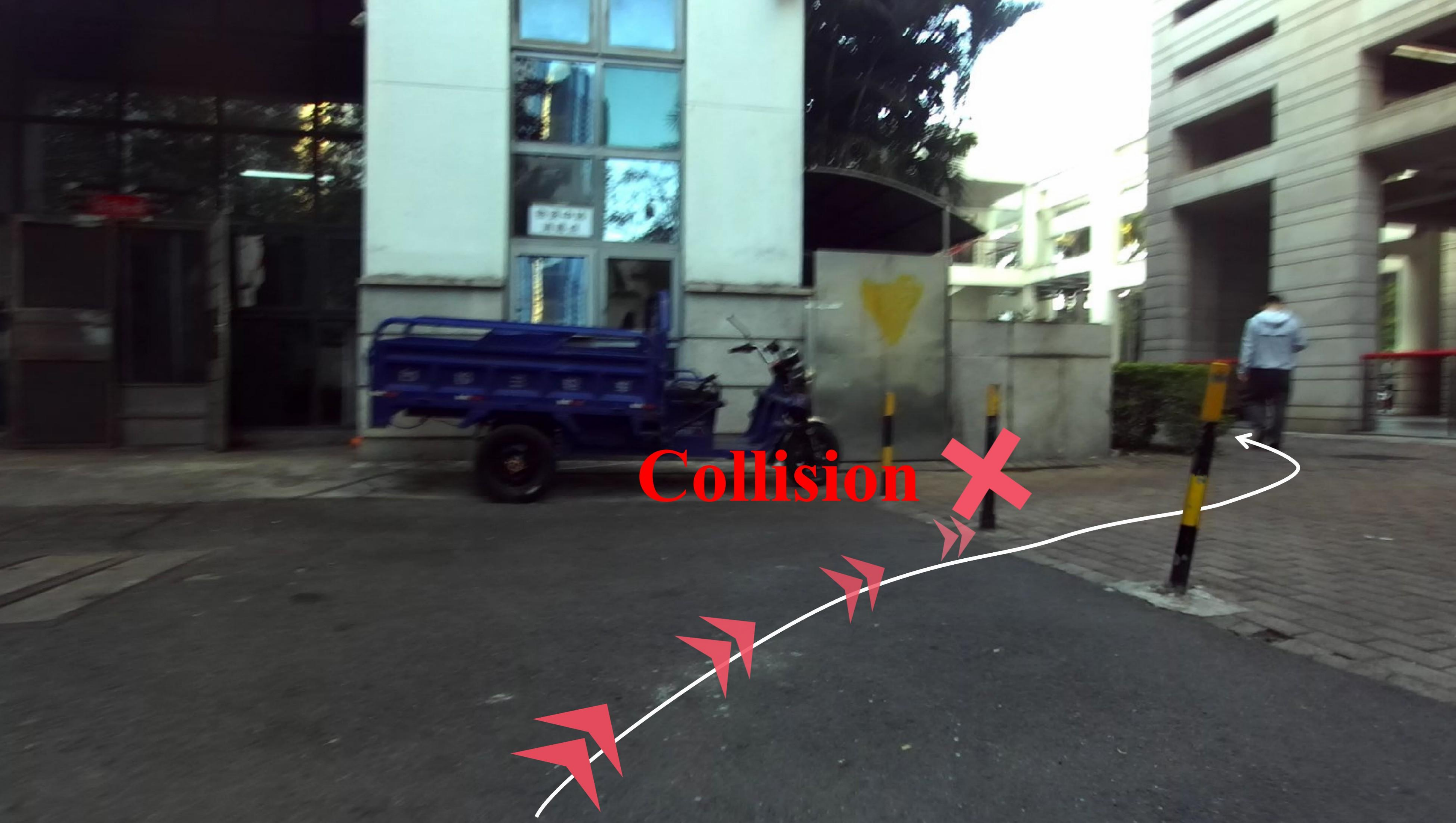}
    }\\
    \subfloat[\textcolor{green}{Ours}, \textcolor{blue}{CT-ICP} with Dynamic Vehicles\label{fig:traj_vehicles}]{
        \includegraphics[width=0.25\textwidth]{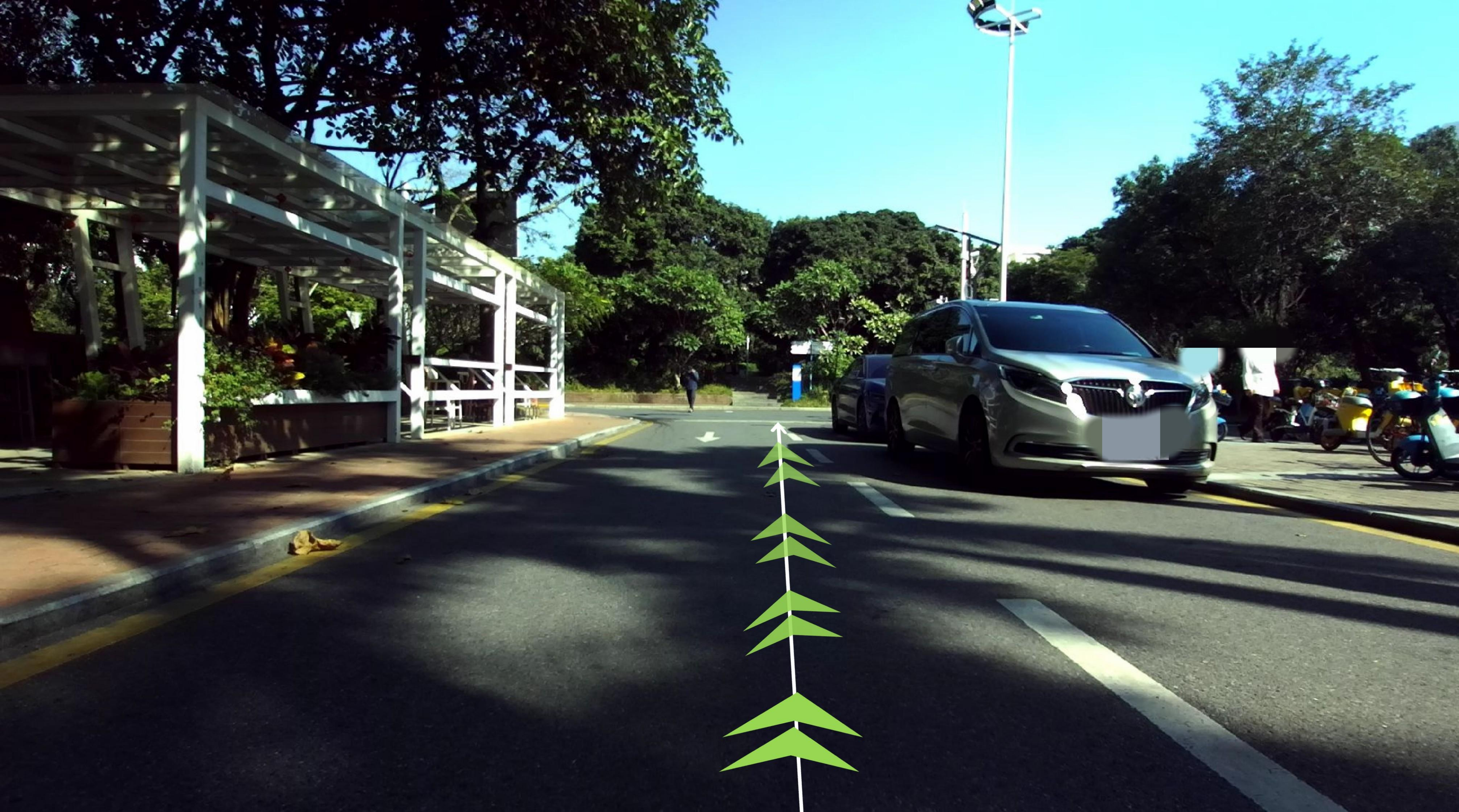}
        \includegraphics[width=0.25\textwidth]{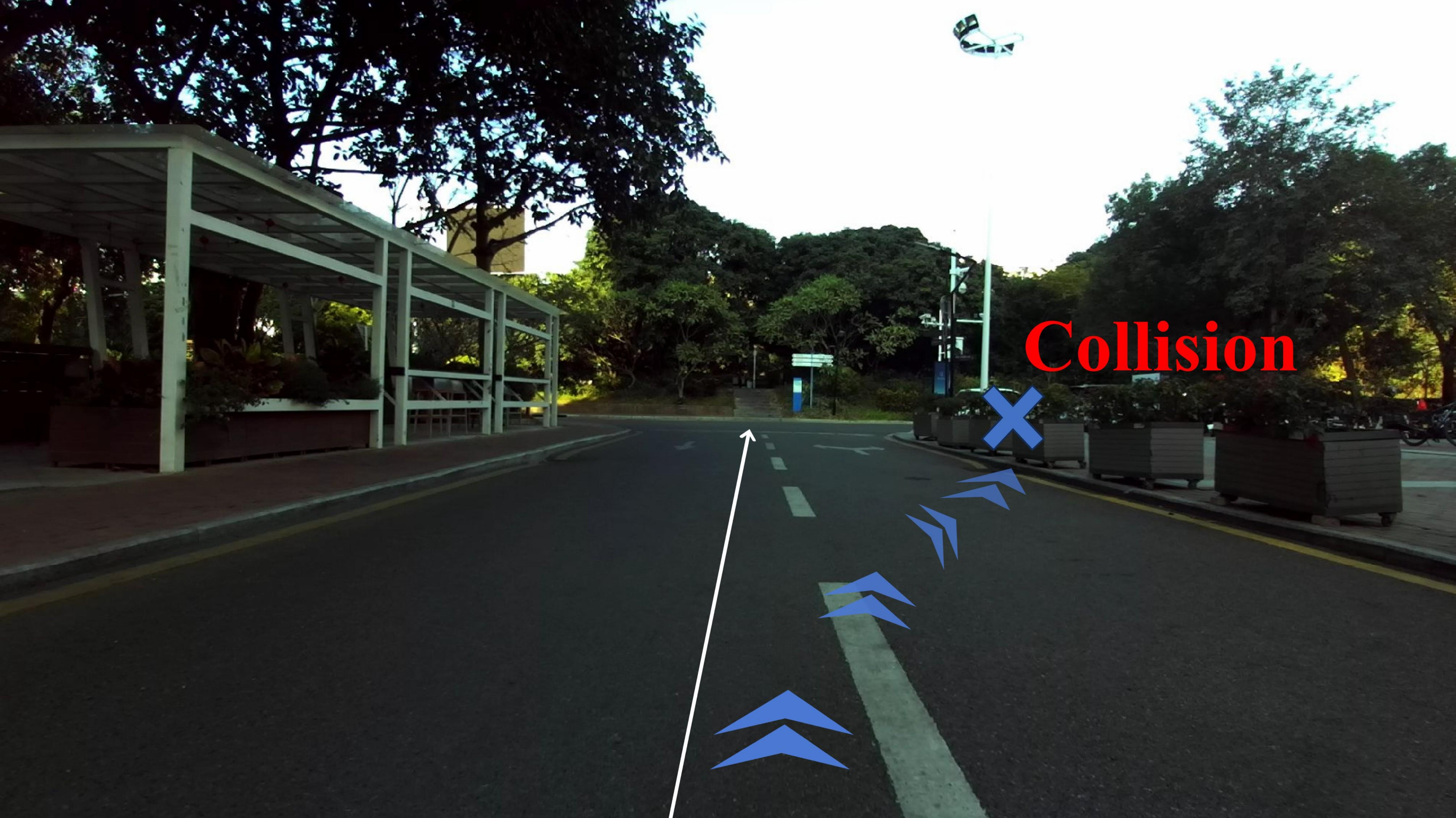}
    }
    \subfloat[\textcolor{green}{Ours}, \textcolor{arrowpurple}{Regformer} with Dynamic Pedestrians\label{fig:traj_pedestrians}]{
        \includegraphics[width=0.25\textwidth]{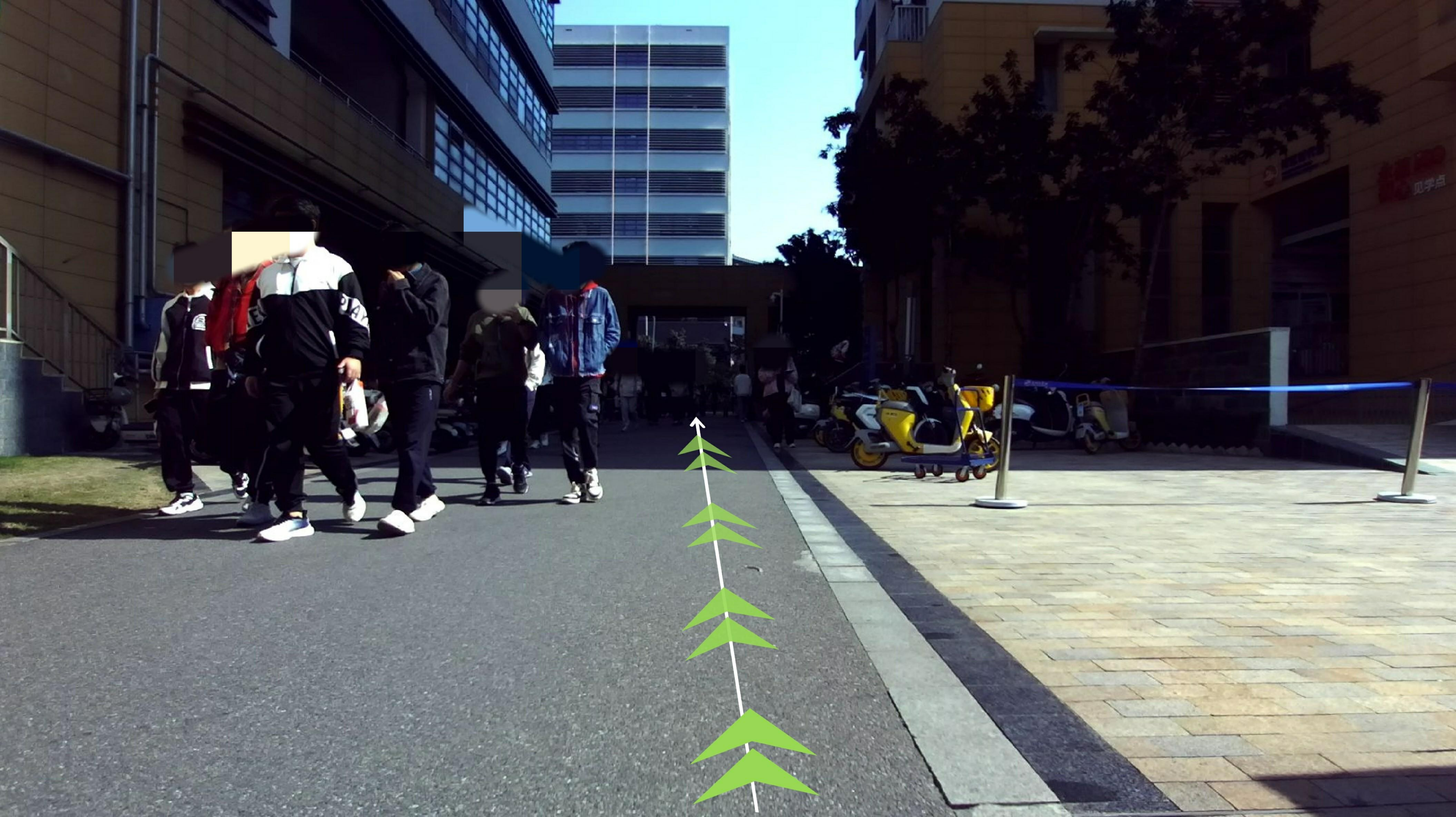}
        \includegraphics[width=0.25\textwidth]{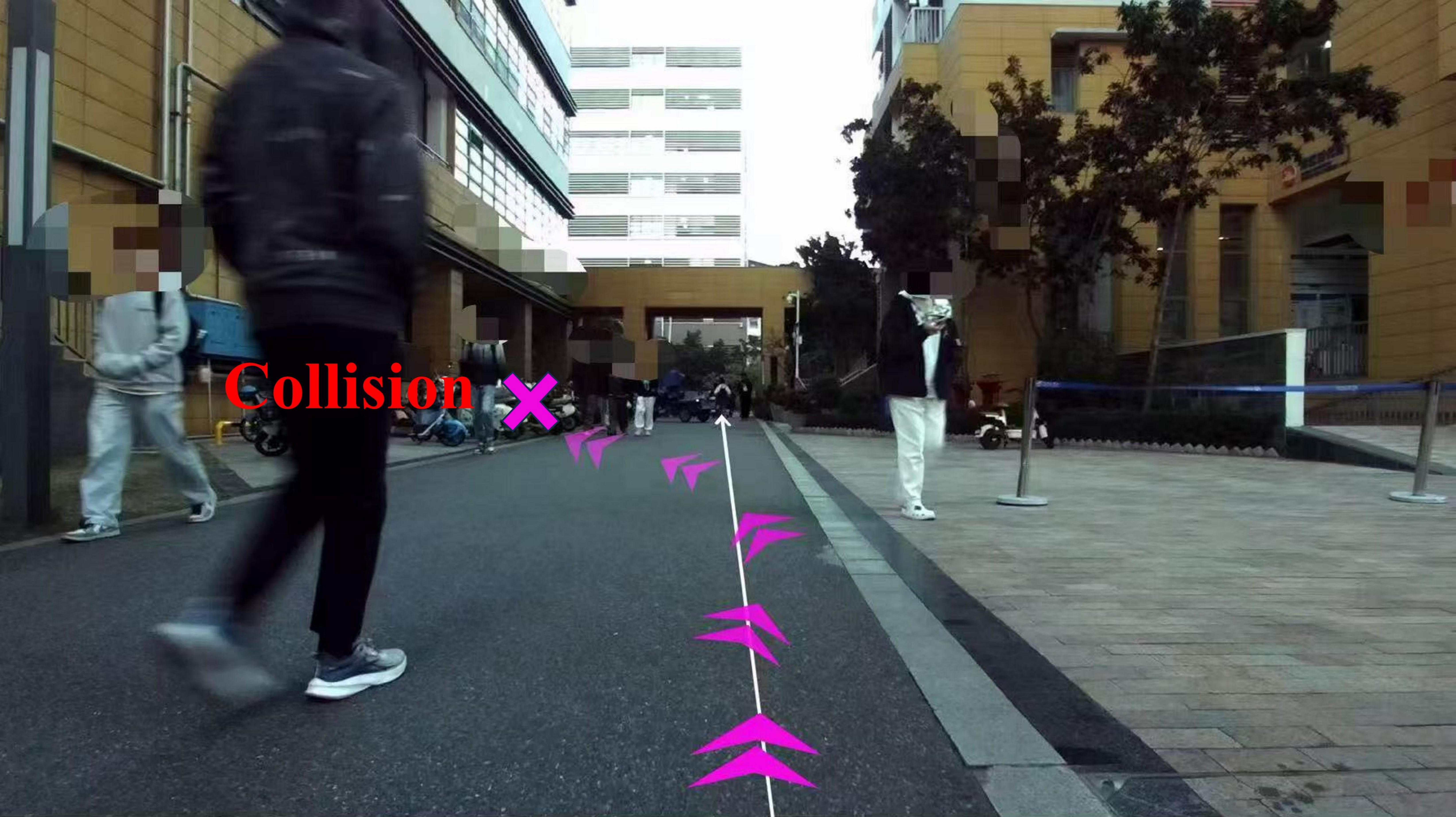}
    }

    \caption{\textbf{Screenshots of different navigation methods in four representative and challenging scenarios.} The white trajectory indicates the teach path, while the colored arrows represent projections of run-time positions onto the current-view image. Compared to representative baselines ADPGICP (Radar T$\&$R), LIO-SAM (SLAM-based), CT-ICP (Cross-modal T$\&$R), Regformer (LiDAR T$\&$R), and our LTR$^2$ system avoids trajectory repeat failures.}  
    \label{fig:trajB2F}
\end{figure*}

As to navigation success rate, only the SLAM-based method, LiDAR T$\&$R, and our proposed CMR method complete all trajectory repeats successfully in trajectory A (see Tab.~\ref{localization1}) and thus are further evaluated on trajectories B to F. 
As shown in Tab.~\ref{localization1},  all single-modal radar T$\&$R methods fail to maintain stable registration during long-range navigation. Radar T$\&$R suffers from the limited  FOV of 4D radar, which is exacerbated during agile movement of the robot, as evidenced in Fig. \ref{fig:trajB2F}  (a), the 4D radar T$\&$R system fails easily on slopes where the yaw angle of AGV changes fast.

The SLAM-based method exhibits high global accuracy and consistency due to map-based localization, but shows spatially non-uniform error distribution, which undermines local accuracy. In contrast, the CMR network maintains uniformly low location errors at discrete repeat nodes, enabling more robust long-range navigation in narrow or complex environments. This is further verified in Fig. \ref{fig:trajB2F}  (b), where the baseline method LIO-SAM collides with the pole, but the LTR$^2$ navigates successfully through the narrow space between poles. 

Similarly, existing cross-modal registration methods struggle to match dense omnidirectional LiDAR with sparse 4D radar scans. As further verified in Fig. \ref{fig:trajB2F} (c), CT-ICP still fails when optimizing its ICP factor, highlighting the limitation of geometry-based features alone for cross-modal registration between LiDAR and 4D radar. By contrast, our CMR network achieves stable and accurate localization comparable to LiDAR T$\&$R approaches, as evidenced by the consistently centimeter-level repeat error in Fig. \ref{fig:traja}. The ablation study of our model with and without the intensity alignment branch is conducted. The one without the intensity alignment branch performs significantly worse and even fails in trajectories E and G (Bottom rows in both Tab.~\ref{table:ate_localization} and Tab.~\ref{table:yae_localization}), highlighting the necessity of aligning radar power density with LiDAR intensity for cross-modal registration. 

In highly dynamic scenarios (e.g., bicycle parking areas, pedestrian zones, or traffic-heavy roads covered in trajectories B and C), the LiDAR-only methods exhibit large localization drift and inconsistent trajectory-repeating navigation. On the contrary, the 4D radar with its ability to perceive beyond-line-of-sight consistently captures static scene features,  maintaining high repeat accuracy in our LTR$^2$ system. This is further verified in Fig. \ref{fig:trajB2F}  (d) where the LiDAR-only baseline Reformer fails due to moving vehicles and pedestrians, but our LTR$^2$ navigates successfully. 

\textbf{Large-Scale Validation:}
To test the generalization of our LTR$^2$ system, we deployed it on a \SI{4.627}{km} trajectory different from all trajectories A-G without any additional training or fine-tuning. As shown in Fig. \ref{fig:intro}(b), many scenes had never been encountered during the CMR pretraining phase. Nonetheless, the LTR$^2$ maintained stable navigation over kilometer-scale trajectories for periods exceeding one hour achieving an average ATE of approximately \SI{0.093}{m}, which shows the robustness and scalability of our CMR network-based navigation system. 

\begin{table}[t]
\centering
\caption{Performance Comparison of Node Selection Methods on Trajectory A [EWA$_{\text{ATE}}$ $\uparrow$, EWA$_{\text{YAE}}$ $\uparrow$]}
\label{tab:traj_Discretization}
\resizebox{\linewidth}{!}{
\begin{tabular}{lcccccc}
\toprule
Threshold & ATE & YAE & Storage & \#Node & EWA$_{\text{ATE}}$ & EWA$_{\text{YAE}}$ \\
\midrule
0.5m/5$^\circ$ & 0.065 & 1.163 & 5068.5 & 2816 & 1.937 & 0.108 \\
1m/10$^\circ$ & 0.076 & 1.410 & 2586.9 & 1436 & 1.810 & 0.098 \\
2m/15$^\circ$ & 0.078 & 1.723 & 1302.8 & 724 & 1.974 & 0.088 \\
3m/30$^\circ$ & 0.102 & 2.833 & 876.1 & 487 & 1.584 & 0.057 \\
5m/45$^\circ$ & 0.158 & 4.261 & 538.1 & 299 & 1.110 & 0.041 \\
8m/60$^\circ$ & / & / & 428.4 & 238 & / & / \\
\textbf{Ours} & 0.073 & 1.481 & 888.3 & 492 & \textbf{2.210} & \textbf{0.109} \\
\bottomrule
\end{tabular}
}
\end{table}

\begin{table*}[t]
\centering
\caption{Quantitative Analysis for Fine-tuning [ATE (\si{\meter})$\downarrow$, YAE (\si{\degree})$\downarrow$, ATE\,STD (\si{\meter})$\downarrow$, YAE\,STD (\si{\degree})$\downarrow$]}
\label{finetune_tab}
\renewcommand{\arraystretch}{1.3}
\resizebox{\linewidth}{!}{
\begin{tabular}{c|c|c|c|c|c|c|c|c|c|c|c}
\toprule
\multicolumn{1}{c|}{} 
& \multirow{2}{*}{\textbf{Method}} 
& \multicolumn{5}{c|}{\textbf{Finetune 01}} 
& \multicolumn{5}{c}{\textbf{Finetune 02}} \\
\cmidrule(r){3-7} \cmidrule(r){8-12}
& & \textbf{TimeStamp} & \textbf{ATE} & \textbf{YAE} & \textbf{ATE STD} & \textbf{YAE STD}
  & \textbf{TimeStamp} & \textbf{ATE} & \textbf{YAE} & \textbf{ATE STD} & \textbf{YAE STD} \\
\midrule

\multirow{3}{*}{\textbf{Current}} 
& LiDAR-LiDAR (GICP)         & 2024-11-19 & 0.099 & 1.861 & 0.068 & 1.831 & 2024-11-04 & 0.128 & 3.160 & 0.073 & 2.113 \\
& LiDAR SLAM (LIO-SAM)               & 2024-11-19 & 0.059 & 1.325 & 0.041 & 1.320 & 2024-11-04 & 0.061 & 2.031 & 0.040 & 1.666 \\
& Ours                      & 2024-11-19 & 0.084 & 1.667 & 0.051 & 1.440 & 2024-11-04 & 0.064 & 2.094 & 0.048 & 1.720 \\
\midrule

\multirow{4}{*}{\textbf{After 2 Months}} 
& LiDAR-LiDAR (GICP)         & / & / & / & / & / & / & / & / & / & / \\
& LiDAR SLAM  (LIO-SAM)               & 2025-01-14 & 0.097 & 1.863 & 0.078 & 1.976 & 2025-01-14 & 0.084 & 2.771 & 0.042 & 2.640 \\
& Ours (Before finetuning)    & 2025-01-14 & 0.134 & 2.281 & 0.231 & 2.304 & 2025-01-14 & 0.194 & 3.008 & 0.324 & 3.710 \\
& Ours (After Finetuning)     & 2025-01-14 & 0.089 & 1.930 & 0.062 & 1.721 & 2025-01-14 & 0.084 & 2.398 & 0.050 & 1.858 \\
\bottomrule
\end{tabular}
}
\end{table*}

\subsection{Node Selection Strategy Comparison}\label{trajcompare}
\begin{figure}[t]
    \centering
        \includegraphics[width=\linewidth]{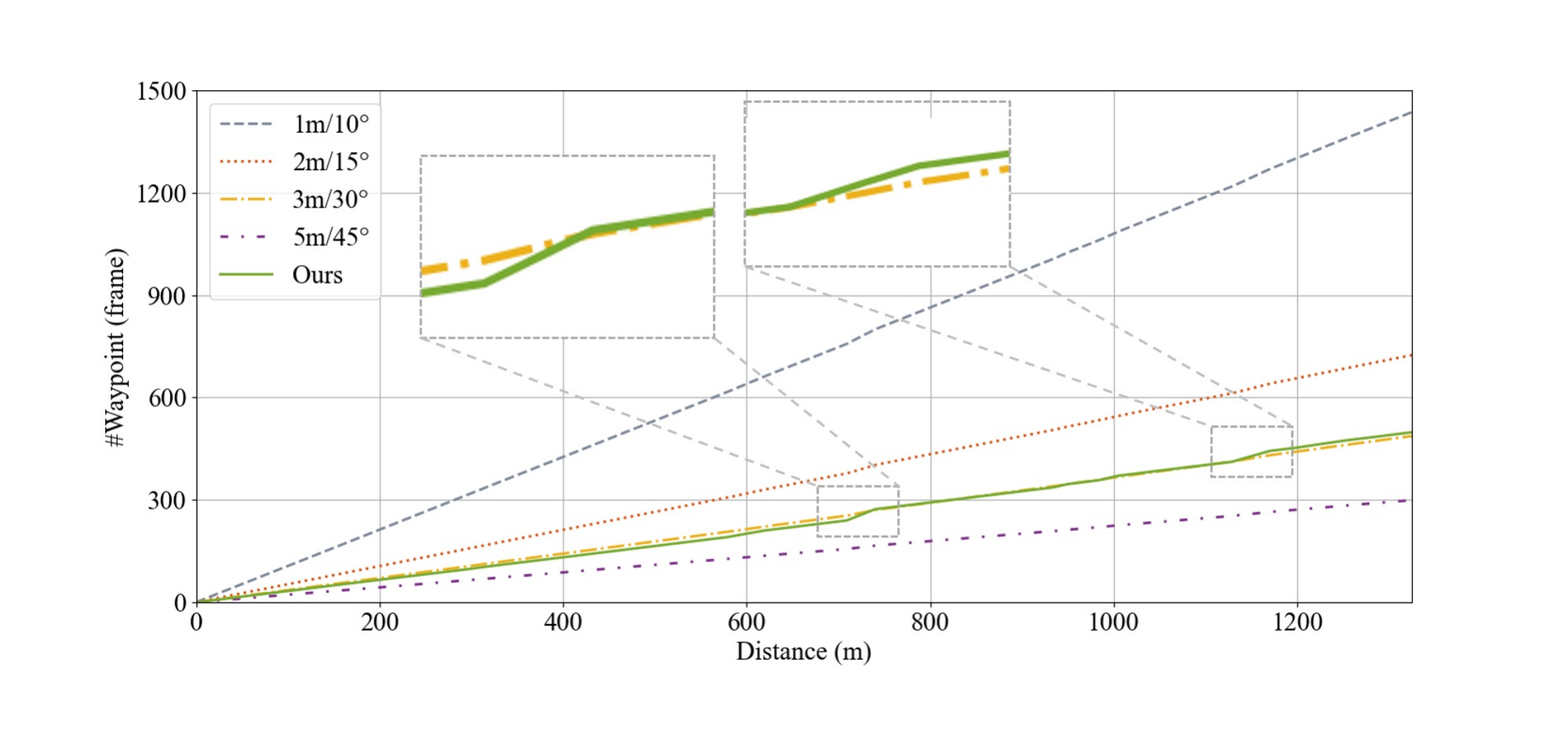}
    \caption{\textbf{Visualization of storage requirements of different node selection strategies on Trajectory A.}}
    \label{fig:threshold}
\end{figure}

We compare our proposed adaptive node selection strategy with the fixed threshold-based ones on trajectory A. Traditional T$\&$R systems often rely on a simple threshold of travelling distance and yaw changes to determine nodes. However, this heuristic approach leads to either excessive storage or navigation failure depending on the manual threshold settings. As shown in Tab. \ref{tab:traj_Discretization} and Fig. \ref{fig:threshold}, a bigger threshold reduces storage burden but increases trajectory repeat errors and even fails in trajectory tracking, while a smaller threshold improves trajectory repeat accuracy but increases storage burden. Moreover, fixed thresholds fail to adapt to varying scene complexity along a trajectory.  

To assess both localization accuracy and storage efficiency, we define a dimensionless metric:
\begin{equation}
\mathrm{EWA}_{\text{error}}
= \frac{1}{\big(\text{error}/e_0\big)\cdot \log N},
\end{equation}
where \text{error} denotes the localization error (ATE or YAE), $N$ is the number of target nodes, 
and $e_0$ is a constant with the same unit as \text{error}. 
We use $e_0^{t}=1\,\mathrm{m}$ for ATE and $e_0^{r}=1\,^\circ$ for YAE. The logarithmic term accounts for diminishing returns with increased node density. The larger $\text{EWA}_{\text{error}}$ means the more accurate localization with less nodes. 

As shown in Tab.~\ref{tab:traj_Discretization}, our adaptive node selection method achieves the best trade-off between navigation accuracy and storage burden. Fig.~\ref{fig:threshold} illustrates that in our node selection method, the node density is automatically adjusted based on scene complexity: sparser in open regions and denser in challenging areas, which removes the need for manual threshold tuning.

\begin{figure}
    \centering
    \includegraphics[width=\linewidth]{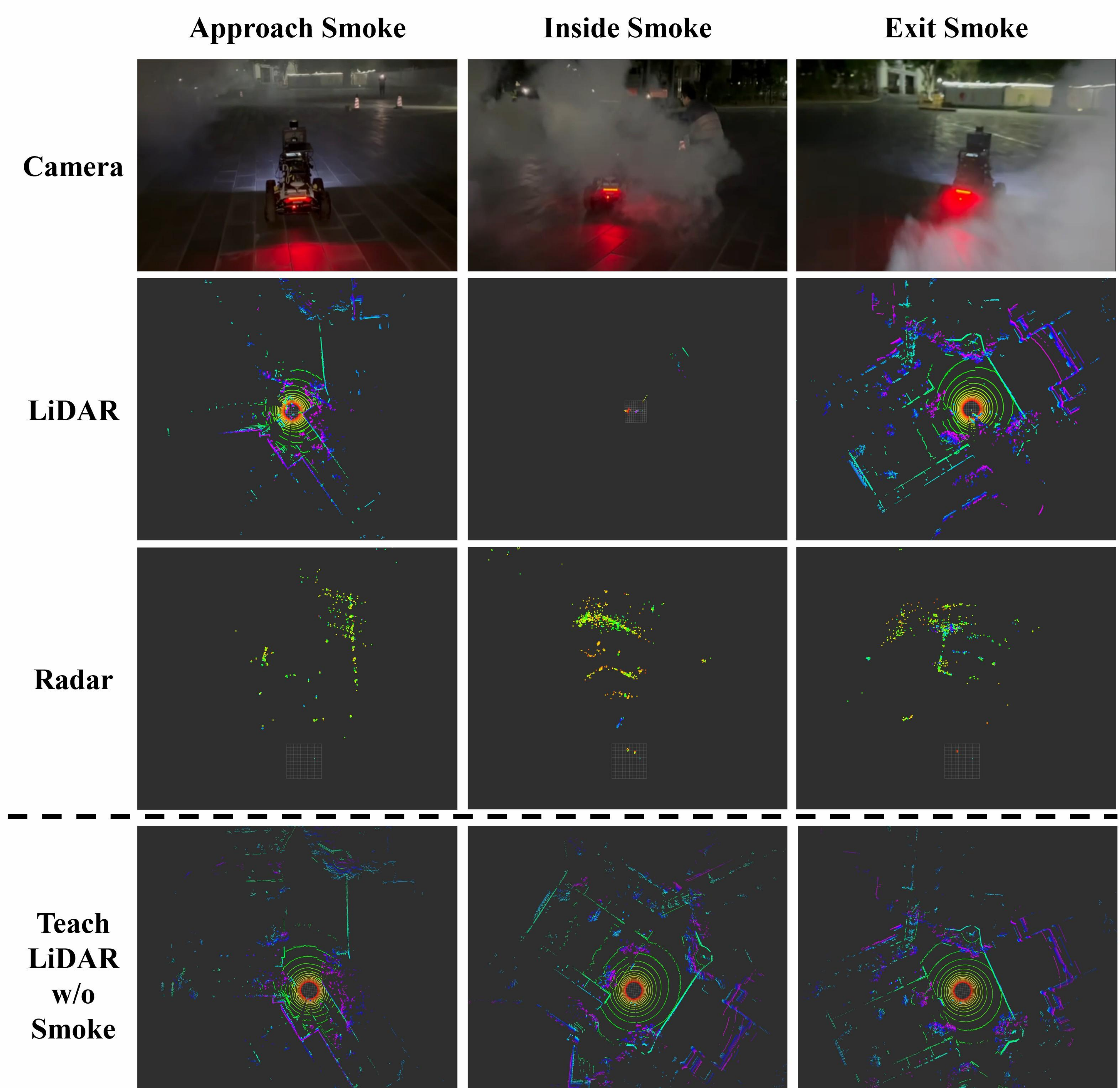}
    \caption{\textbf{Perception comparison of different sensors when repeating in a heavy smoke environment.} LiDAR data experiences severe perception degradation, while 4D Radar maintains robust perception. The visual image is only for visualization,  not used in navigation.}
    \label{sensor_smoke_show}
\end{figure}

\begin{figure}[htbp]
    \centering
    \subfloat[Finetune1\label{fig:Finetune1}]{
        \includegraphics[width=\linewidth]{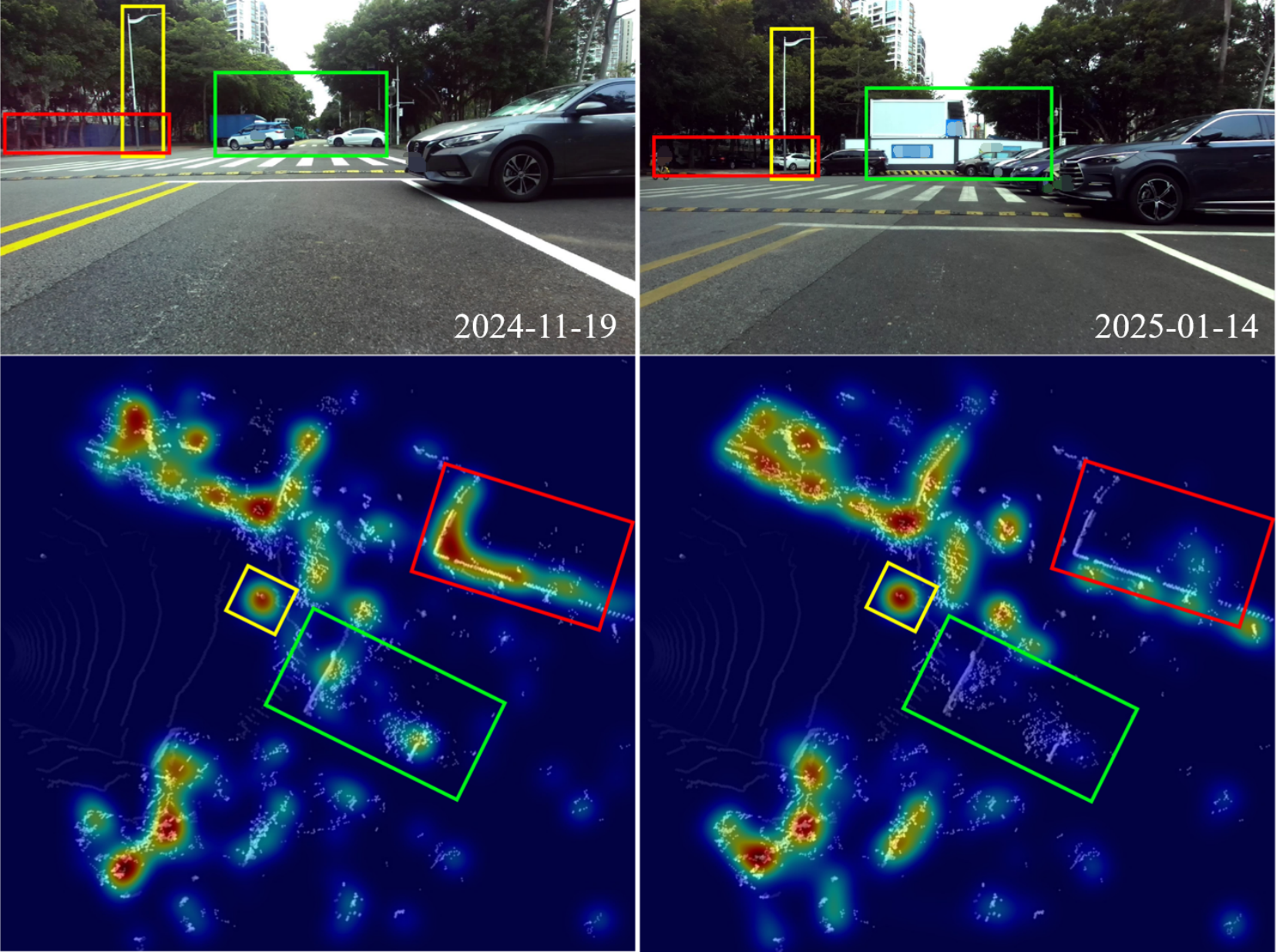}
    }\
    \subfloat[Finetune2\label{fig:Finetune2}]{
        \includegraphics[width=\linewidth]{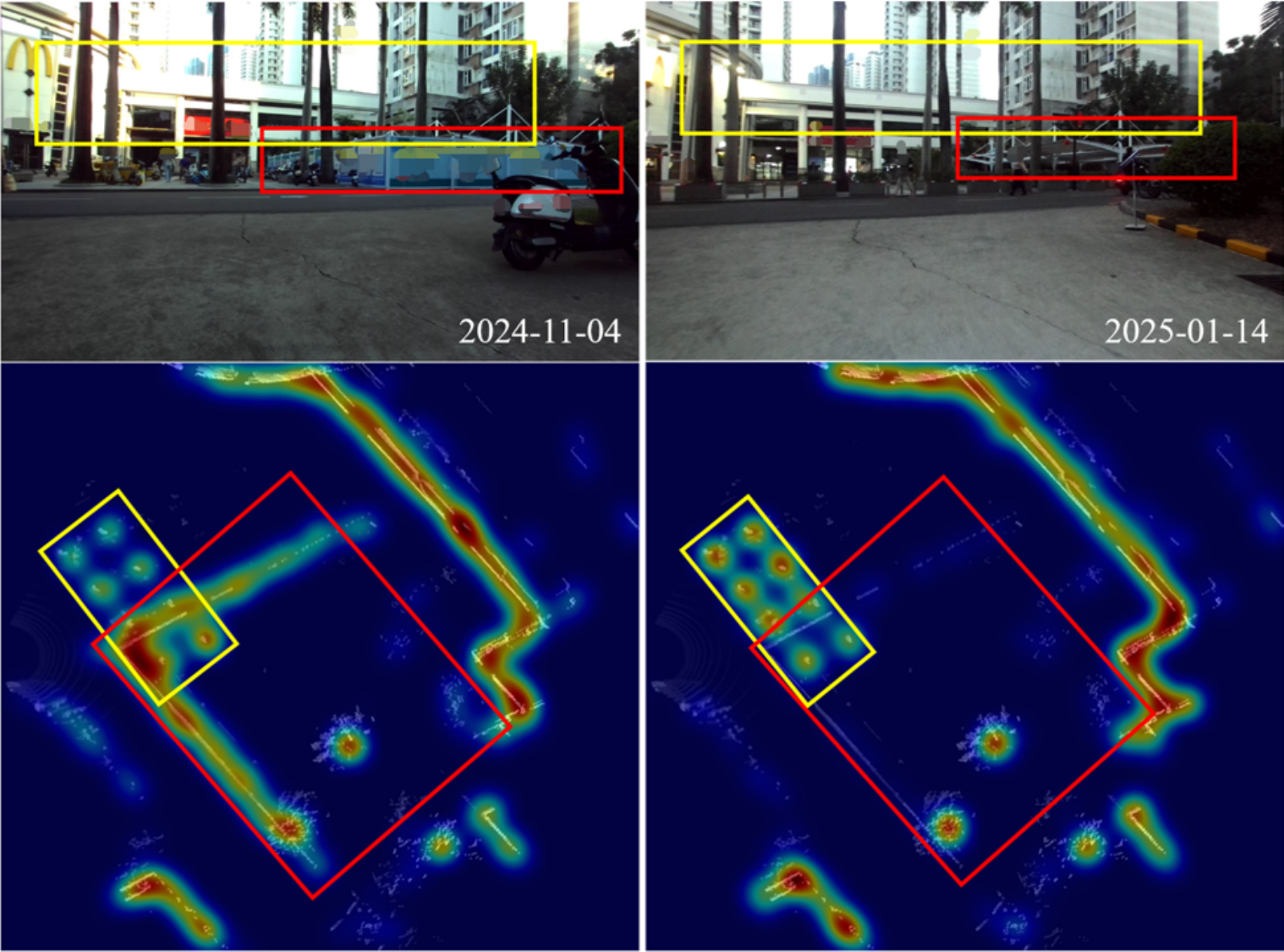}
    }
    \caption{\textbf{Visualization of static environmental changes.} The top rows in (a) and (b) illustrate repeat trajectories before and after fine-tuning. Structures marked by green boxes denote newly constructed elements, red boxes indicate demolished elements, and yellow boxes represent persistent, unchanged structures. Bottom rows display distribution shifts of attention mechanisms corresponding to environmental changes.}
    \label{fig:fintunevisualization}
\end{figure}

\subsection{Robustness in Smoke-Filled Environments}\label{smoketest}
To demonstrate the advantage of radar-based repeat navigation under degraded visibility, we conducted experiments in a smoke-filled environment. A LiDAR-based teaching trajectory was recorded on a clear day. During the repeating phase, dense chemical smoke was filled over a \SI{20}{m} $\times$ \SI{20}{m} area to simulate harsh visual degradation. As shown in  Fig.~\ref{sensor_smoke_show}, 4D radar perception remained unaffected, while LiDAR performance degraded significantly due to occlusion and scattering of smoke particles. SLAM-based ground truth could not be reliably obtained in heavy smoke, so we globally aligned repeat trajectories based on the same start/end nodes and the registered teach and repeat nodes. As shown in Fig. \ref{fig:intro}(a), the two trajectories are very close to each other,  validating the resilience of the CMR network and the reliable navigation of LTR$^2$ system under adverse conditions.

\subsection{CMR Network Finetuning}\label{timefinetunecompare}
In this subsection, we assess the long-term adaptability of LTR$^2$ to static environmental changes. Two trajectories termed Finetune1 and Finetune2 were recorded two months after their respective teaching phases. As illustrated in the upper row of Fig. \ref{fig:fintunevisualization} (a)-(b), the two scenes encountered new construction and demolition in some areas, though many stable structures remained after two months.  

\begin{figure}[t]
    \centering
    \subfloat[Finetune1 qualitative evaluation\label{fig:Finetune1error_vis}]{
        \includegraphics[width=0.9\linewidth]{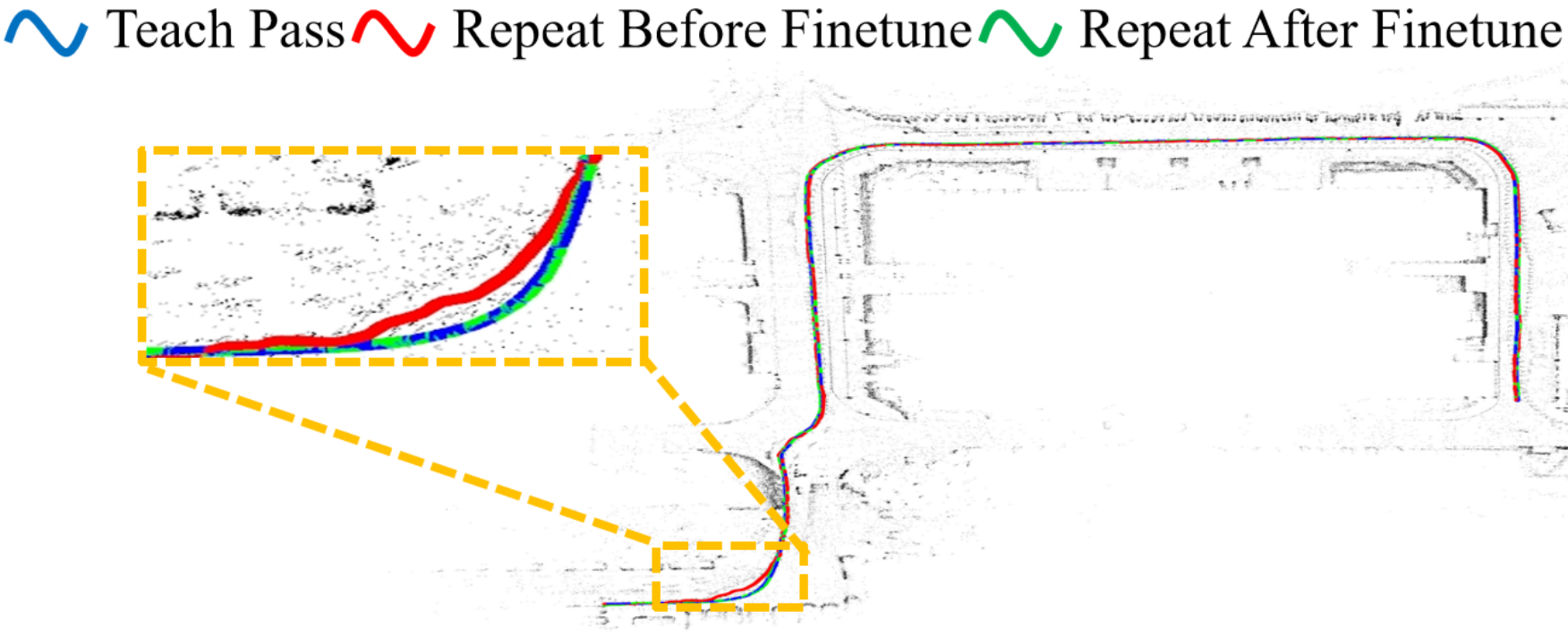}
    }\
    \subfloat[Finetune1 quantitative evaluation\label{fig:Finetune1error}]{
        \includegraphics[width=\linewidth]{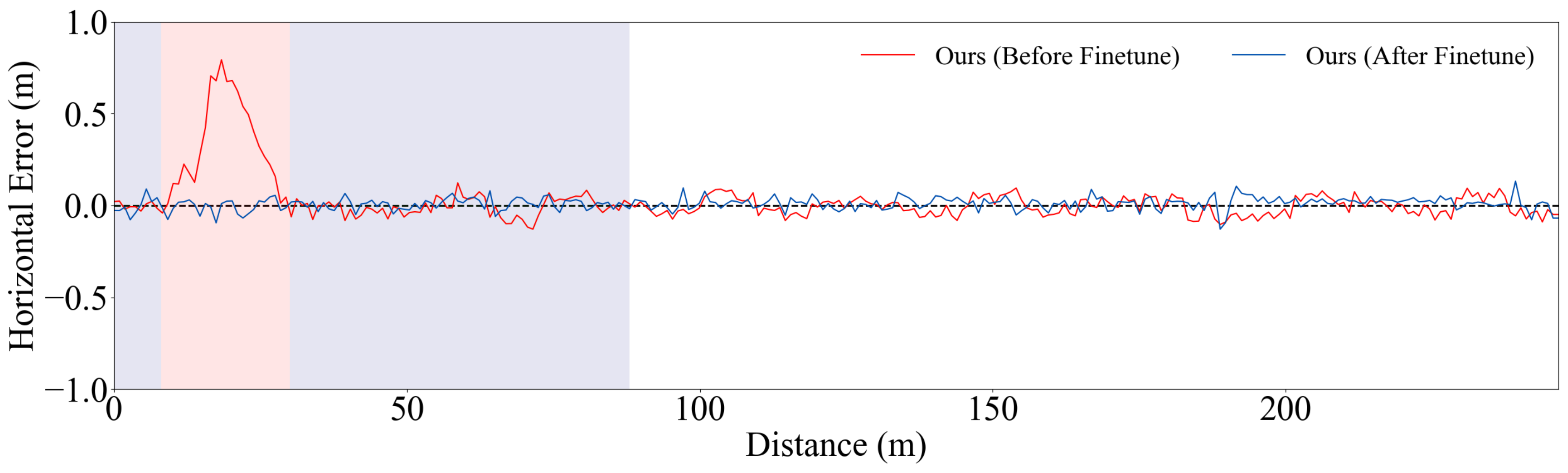}
    }\
    \subfloat[Finetune2 qualitative evaluation\label{fig:Finetune2error_vis}]{
        \includegraphics[width=0.9\linewidth]{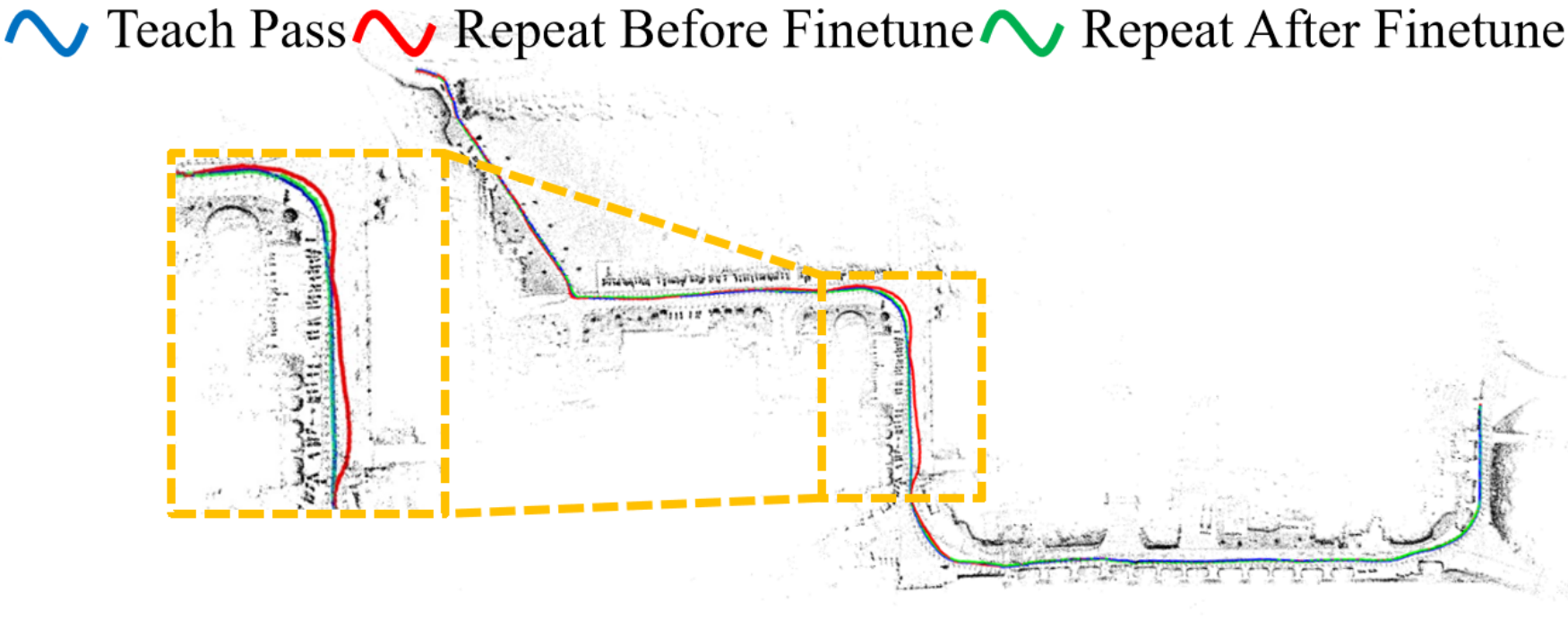}
    }\
    \subfloat[Finetune2 quantitative evaluation\label{fig:Finetune2error}]{
        \includegraphics[width=\linewidth]{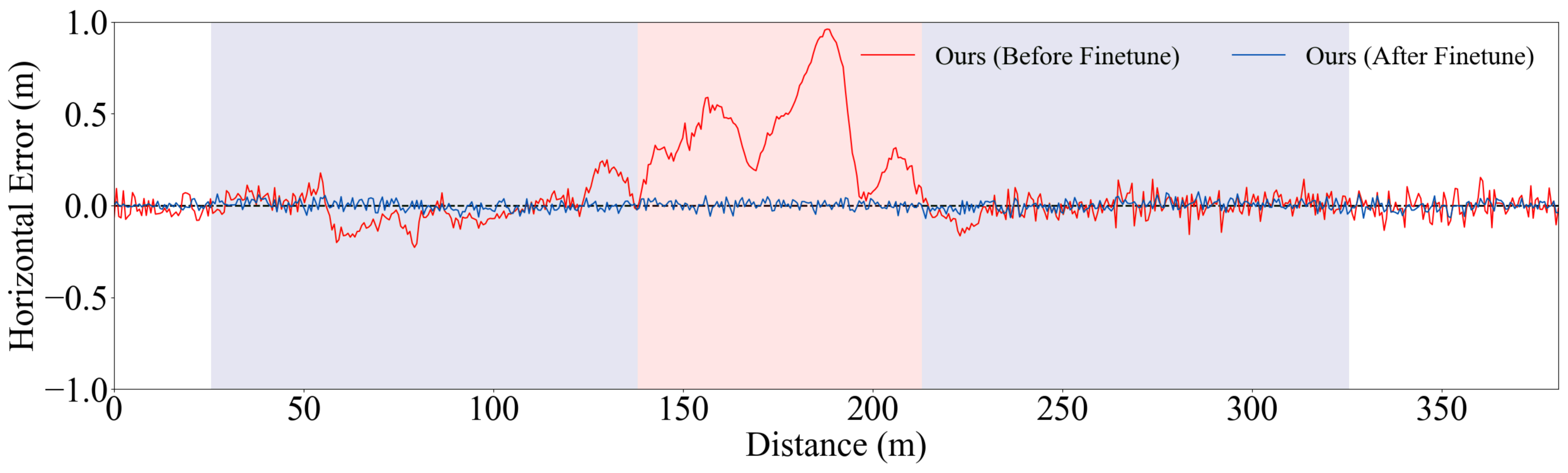}
    }\
    \caption{\textbf{Trajectory evaluation of fine-tuning.}  Larger localization errors occur against static environmental changes, and are mitigated after the fine-tuning. The red and blue shadowed regions denote where negative and positive samples are collected, respectively.}
    \label{fig:fintune_error}
\end{figure}

As described in Subsection \ref{ft}, the evaluation identifies the large deviation of error in trajectory segments, as shown in Fig. \ref{fig:fintune_error}.   Tab. \ref{finetune_tab} indicates that LTR$^2$ after finetuning provides a more accurate trajectory repeat, maintaining the reliability of navigation against static environment changes. A deeper insight into the adaptive finetuning is also provided in the lower rows of Fig. \ref{fig:fintunevisualization} (a) and (b), which visualizes the distribution shift of attention weighting of the CMR network before and after finetuning. After finetuning, the attention mechanism refocuses from altered areas (red and green boxes) to stable areas (yellow boxes).  In comparison,  all three competing methods suffer degraded performance against static environmental changes due to the lack of adaptive update mechanisms.  

\subsection{Cross-Platform Feasibility Demonstration}\label{crossplatform}
\begin{figure}[t]
    \centering
    \subfloat[The robot platforms used for trajectory teaching, including a vehicle and a point-legged bipedal robot.\label{fig:platform2}]{
        \includegraphics[width=\linewidth]{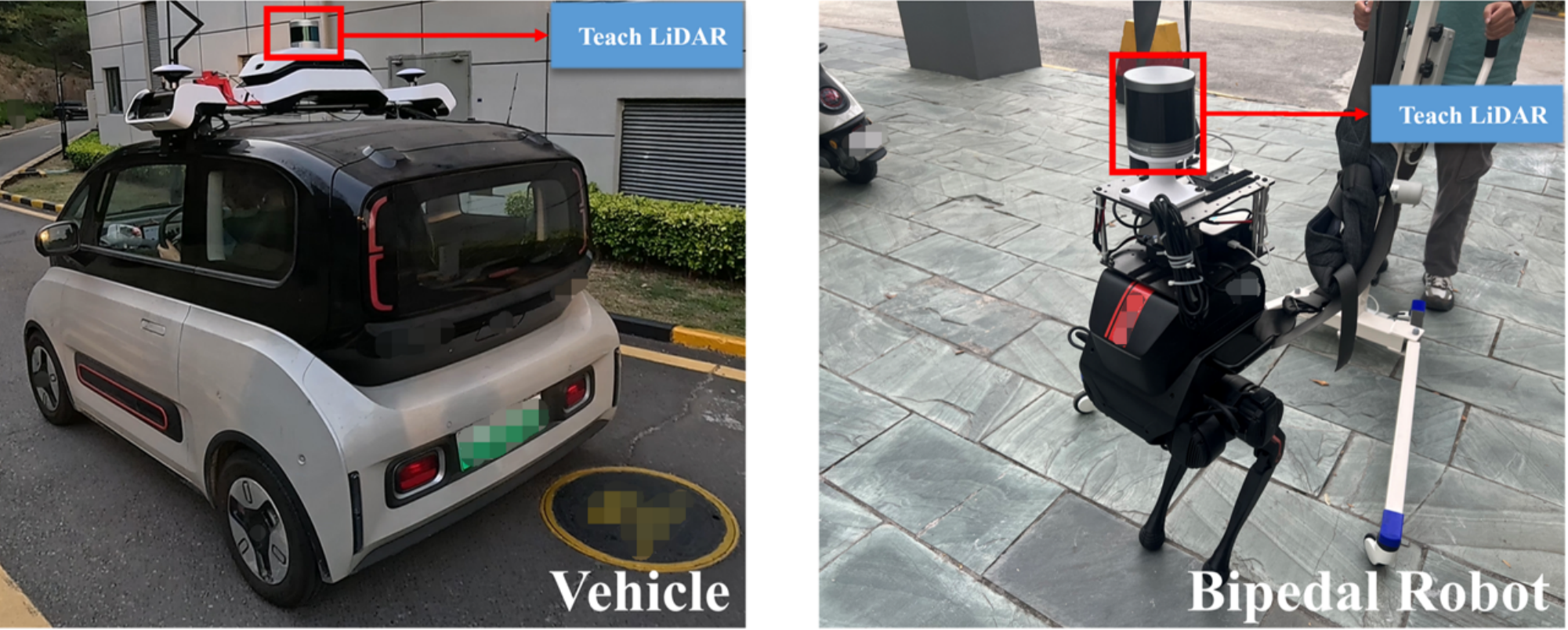}
    }\
    \subfloat[Trajectory repeat error for vehicle-based teach.\label{fig:car-platform}]{
        \includegraphics[width=\linewidth]{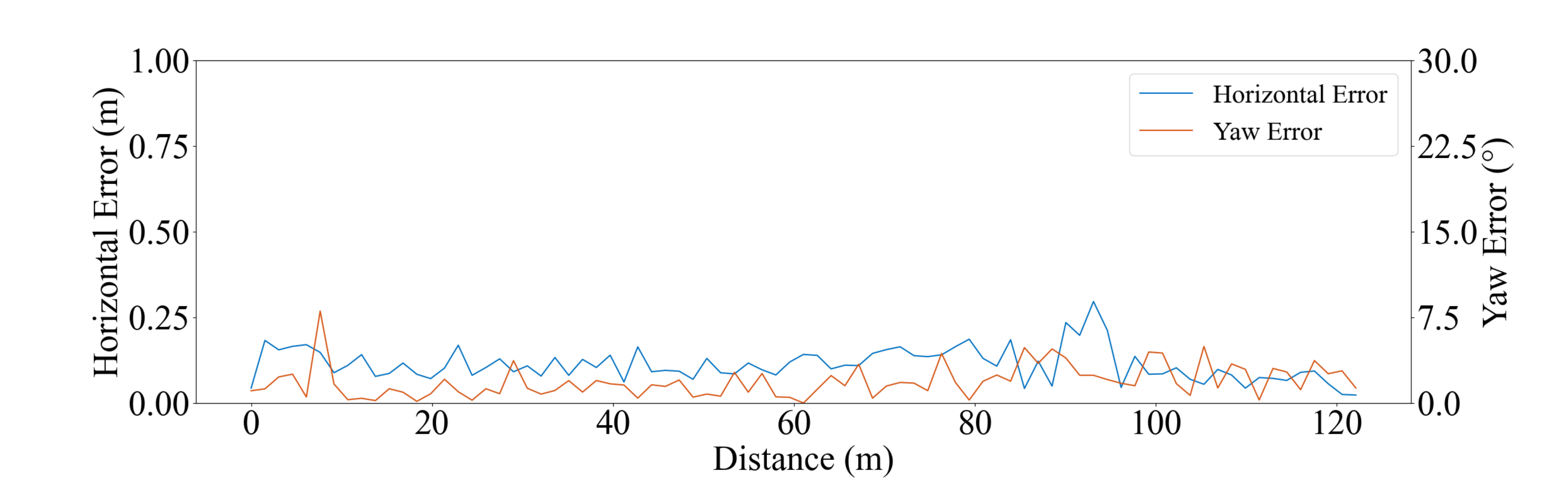}
    }\
    \subfloat[Trajectory repeat error  for bipedal robot-based teach.\label{fig:Bipedal_platform}]{
        \includegraphics[width=\linewidth]{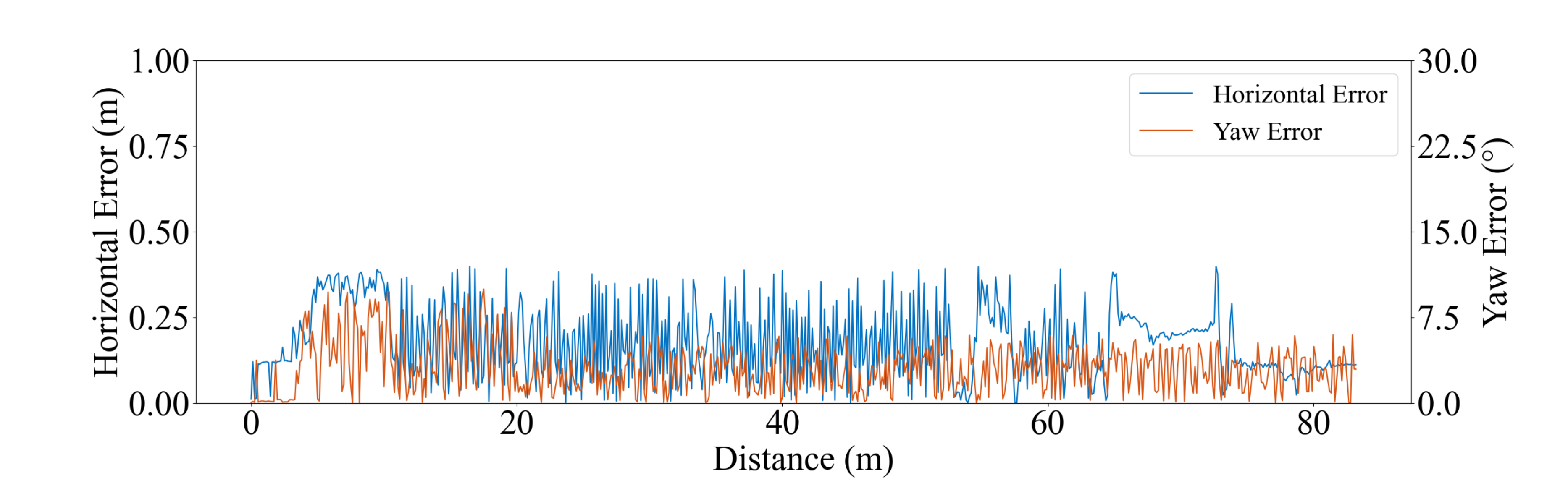}
    }
    \caption{\textbf{Visualization of cross-platform navigation.}}
    \label{fig:cross-platform}
\end{figure}

To further examine the robustness of the proposed CMR network under substantial platform and teach-side sensing differences, we conducted a qualitative cross-platform teach-and-repeat demonstration using different robot platforms for teaching and repeating. Unlike the previous closed-loop evaluations, this part is intended as a feasibility study to test whether the same CMR network can still support route reuse when the teach trajectory is provided by substantially different platforms. Specifically, the teaching phase was performed separately using a data collection vehicle and a point-legged biped robot, as shown in Fig. \ref{fig:cross-platform} (a), while the repeat phase was always carried out by the AGV platform using the same CMR network as in the previous closed-loop experiments. The data collection vehicle was equipped with a 128-channel Ruby plus LiDAR mounted at approximately \SI{1.6}{\meter}, whereas the point-legged biped robot used a 32-channel Helios LiDAR mounted at approximately \SI{0.6}{\meter} when standing upright. In addition to the different LiDAR configurations, the platforms also differ markedly in body morphology, mounting geometry, and motion characteristics. The point-legged biped robot and the data collection vehicle are standard research platforms without integrated 4D radar sensors, so the teach node selection was sampled at a fixed sampling rate of \SI{14}{\hertz}. Then the AGV platform repeated the path using the same CMR network as that employed in the previous closed-loop experiments.

As shown in Fig. \ref{fig:cross-platform} (b) and (c), the AGV was able to complete route repetition in both  cases. These results demonstrate that the proposed CMR network can support cross-platform route reuse at the localization and route-following level, showing robustness to substantial differences in teach-side sensor configuration and platform. The full cross-platform demonstration is also provided in the supplementary video.

\section{Conclusion}\label{Conclusion}
In this paper, we propose a novel omnidirectional LiDAR-forward 4D radar cross-modal registration method, integrated into a LiDAR-teach-and-radar-repeat navigation system. To address the limited geometric resolution of raw radar data, we incorporate a Doppler-based motion prior for FOV alignment and introduce a physics-informed radar power  LiDAR intensity alignment mechanism. An adaptive node selection strategy is adopted to maintain viewpoint consistency and registration success while minimizing storage overhead. A CMR network fine-tuning strategy is proposed to adapt to static environmental changes, ensuring robust cross-modal registration and long-term navigation. Experimental results demonstrate consistently accurate positioning and navigation under degenerate and changing environments. Future work will investigate how to enhance 4D mmWave radar occupancy mapping to support reliable planning and obstacle avoidance, thereby increasing the safety of cross-modal T$\&$R navigation system.


\bibliographystyle{IEEEtran}
\bibliography{IEEEabrv , reference}
\end{document}